\documentclass[a4paper,12pt,reqno]{amsart}
\usepackage{booktabs}
\usepackage{multirow}
\usepackage{etex}
\usepackage{amsmath}
\usepackage{amstext}
\usepackage{amsfonts}
\usepackage{verbatim}
\usepackage{amssymb}
\usepackage{a4wide,amscd}
\usepackage{graphicx}
\usepackage{bm}
\usepackage{color}
\definecolor{vdarkred}{rgb}{0.6,0,0.2}
\definecolor{vdarkblue}{rgb}{0,0.2,0.6}
\usepackage[pdftex,colorlinks]{hyperref}
\usepackage{hyperref}

\usepackage{tikz}
\usepackage{pgfplots,filecontents}
\usetikzlibrary{spy}
\usetikzlibrary{shapes}
\usetikzlibrary{plotmarks}

\tikzset{new spy style/.style={spy scope={%
	magnification=5,
	size=1.25cm,
	connect spies,
	every spy on node/.style={
		rectangle,
		draw,
	},
	every spy in node/.style={
		draw,
		rectangle,
		fill=gray!40,
	}
}}}

\usepackage{optional}

\newcommand{\RR}{{\mathbb{R}}}
\newcommand{\NN}{{\mathbb{N}}}
\newcommand{\ZZ}{{\mathbb{Z}}}
\newcommand{\CC}{{\mathbb{C}}}

\newcommand{\trans}{{\sf T}}

\newcommand{\EE}{{\rm E}}

\DeclareMathOperator{\tr}{tr}

\DeclareMathOperator{\diag}{\rm diag}

\newcounter{ctheorem}
\newtheorem{theorem}[ctheorem]{Theorem}
\newcounter{cassumption}
\newtheorem{assumption}[cassumption]{Assumption}

\newcounter{cproposition}
\newtheorem{proposition}[cproposition]{Proposition}
\newcounter{ccorollary}
\newtheorem{corollary}[ccorollary]{Corollary}
\newcounter{clemma}
\newtheorem{lemma}[clemma]{Lemma}

\theoremstyle{definition}
\newcounter{cremark}
\newtheorem{remark}[cremark]{Remark}

\definecolor{darkgreen}{rgb}{0.125,0.5,0.169}

\begin{document}


\title{The Asymptotic Performance of Linear Echo State Neural Networks}


\author{Romain Couillet}
\address[R.C.]{CentraleSup\'elec -- LSS -- Universit\'e ParisSud (Gif-sur-Yvette, France).}
\email{romain.couillet@centralesupelec.fr}
\author{Gilles Wainrib}
\address[G.W.]{D\'epartement Informatique, team DATA, Ecole Normale Sup\'erieure (Paris, France).}
\email{gilles.wainrib@ens.fr}
\author{Harry Sevi}
\address[H.S.]{Laboratoire de Physique, Ecole Normale Sup\'erieure de Lyon (Lyon, France).}
\email{}
\author{Hafiz Tiomoko Ali}
\address[H.T.A.]{CentraleSup\'elec -- LSS -- Universit\'e ParisSud (Gif-sur-Yvette, France).}
\email{hafiz.tiomokoali@centralesupelec.fr}



\thanks{Romain Couillet and Hafiz Tiomoko Ali are with the LSS Signals and Stats lab of CentraleSup\'elec, Gif sur Yvette, France; Gilles Wainrib is with the Computer Science department of ENS Paris, Paris, France; Harry Sevi is with the Physics lab of ENS Lyon, Lyon, France. \\ \indent Couillet's work is supported by the ANR RMT4GRAPH (ANR-14-CE28-0006).}




\begin{abstract}
	In this article, a study of the mean-square error (MSE) performance of linear echo-state neural networks is performed, both for training and testing tasks. Considering the realistic setting of noise present at the network nodes, we derive deterministic equivalents for the aforementioned MSE in the limit where the number of input data $T$ and network size $n$ both grow large. Specializing then the network connectivity matrix to specific random settings, we further obtain simple formulas that provide new insights on the performance of such networks.
\end{abstract}	

\maketitle


\section{Introduction}
\label{sec:intro}

Echo State Networks (ESN) are a class of recurrent neural networks (RNN) designed for performing supervised learning tasks, such as time-series prediction \cite{JAE01b,JAE04} or more generally any supervised learning task involving sequences. The ESN architecture is a special case of the general framework of reservoir computing \cite{LUK09}. The ESN reservoir is a fixed (generally randomly designed) recurrent neural network, driven by a (usually time dependent) input. Since the internal connectivity matrix is not modified during learning, the number of parameters to learn is much smaller than in a classical RNN setting and the system is as such less prone to overfitting. However, the prediction performance of ESN often depends significantly on several hyper-parameters controlling the law of the internal connectivity matrix.

It has in particular been understood that the spectral radius and spectral norm of the connectivity matrix play a key role on the stability of the network \cite{JAE01b} and that the structure of the connectivity matrix may be adapted to trade memory capacities versus task complexity \cite{VER10,ROD11,STRA12,OZT07}. Nonetheless, to date, and to the best of the authors' knowledge, the understanding of echo-state networks has progressed mostly through extensive empirical studies and lacks solid theoretical foundations.

In the present article, we consider linear ESN's with a general connectivity matrix and internal network noise. By leveraging tools from the field of random matrix theory, we shall attempt to provide a first {\it theoretical} study of the performance of ESN's. Beyond the obvious advantage of exploiting theoretical formulas to select the optimal hyper-parameters, this mathematical study reveals key quantities that intimately relate the internal network memory to the input-target relationship, therefore contributing to a better understanding of short-term memory properties of RNNs. 

\medskip

More specifically, we shall consider an $n$-node ESN trained with an input of size $T$ and shall show that, assuming the internal noise variance $\eta^2$ remains large compared to $1/\sqrt{n}$, the training and testing performances of the network can be well approximated by a deterministic quantity which is a function of the training and test data as well as the connectivity matrix. Under the further assumption that the connectivity matrix is random, we shall then obtain closed-form formulas of the aforementioned performances for a large class of connectivity structures. The reach of our study so far only addresses ESN's with linear activation functions, a limitation which we anticipate to work around with more elaborate methods in the future, as discussed in Section~\ref{sec:conclusion}.

\medskip

At this point, we wish to highlight the specificity of our approach regarding (i) the introduction of noise perturbing the internal dynamics of the reservoir and (ii) the restriction to linear networks. 

The introduction of additive noise in the reservoir is inspired on the one hand by it being a natural assumption in modelling biological neural networks \cite{GAN08,TOY11} and on the other hand by the observation in \cite{JAE01b} that ESN's are very sensitive to low variance noise and thus likely unstable in this regime, a problem successfully cured by internal noise addition \cite{JAE05}.\footnote{According to Jaeger in \cite{JAE05} (specifically in the context of reservoirs with output feedback): ``When you are training an ESN with output feedback from accurate (mathematical, noise-free) training data, stability of the trained network is often difficult to achieve. A method that works wonders is to inject noise into the dynamical reservoir update during sampling [\dots]. It is not clearly understood why this works.''} From the neuro-computational perspective, we shall observe tight connections between the ESN performance and the reservoir information processing capacities discussed in \cite{GAN08}. As for the artificial neural network viewpoint, it shall be noticed that the internal noise regularizes the network in a way sharing interesting similarities with the well-known connection between noise at the network output and Tikhonov regularization \cite{BIS95}. It is, as a matter of fact, already mentioned in \cite[Section~8.2]{LUK09} that internal noise behaves as a natural regularization option (similar to what input or output noise would) although this aspect was not deeply investigated. More importantly, while in-network noise necessarily leads to random outputs (a not necessarily desirable feature on the onset), we shall show that all these outputs (almost surely) asymptotically have the same performance, thus inducing random but equally useful innovation; this we believe is a more desirable feature than deterministic biases as innovation noise, beyond additionally bringing closer neurophysiological and artificial neural network considerations. 

As for the choice of studying linear activation functions, rarely considered in the practical side of RNNs, it obviously follows first from a mathematical tractability of the problem under study. Nonetheless, while being clearly a strong limitation of our study (recall that the non-linearity is the main driver for the network to perform complex tasks), we believe it brings sufficient insights and exploitable results when it comes to parametrizing non-linear network counterparts. This belief is mainly motivated by the fact that, as the network size grows, if the reservoir connections are drawn at random, the reservoir mixing of the internal states should keep most of these states small in amplitude, thus being for most of them in the linear part of the activation function support. This should be all the more adequate that the network inputs are not too erratic.

\medskip

Among other results, the main findings of our study are as follows:
\begin{enumerate}
	\item we retrieve a deterministic implicit expression for the mean-square error (MSE) performance of training and testing for any fixed connectivity matrix $W\in\RR^{n\times n}$ which, for every given internal noise variance $\eta^2>0$, is all the more accurate that the network size $n$ is large
	\item the aforementioned expression reveals fundamental quantities which generalize several known qualitative notions of ESN's, such as the {\it memory capacity} and the {\it Fisher memory curve} \cite{JAE01b,GAN08};
	\item we obtain more tractable closed-form expressions for the same quantities for simple classes of random normal and non-normal matrices; these two classes exhibit a strikingly different asymptotic performance behavior;
	\item from the previous analysis, we shall also introduce a novel multi-modal connectivity matrix that adapts to a wider scope of memory ranges and that is reminiscent to the long short-term memory ESNs designed in \cite{XUE07};
	\item an important interplay between memory and internal noise will be shed light on, by which the questions of noise-induced stability are better understood.
\end{enumerate}

\medskip

The remainder of the article is organized as follows. In Section~\ref{sec:results}, we introduce the ESN model and the associated supervised learning problem and we give our main theoretical results in Theorems~\ref{th:deteq1} and \ref{th:deteq2} (technical proofs are deferred to the Appendix). Then, in Section~\ref{sec:applications}, we apply our theoretical results for various choices of specific connectivity matrices and discuss their consequences in terms of prediction performance. Finally, in Section~\ref{sec:conclusion}, we discuss our findings and their limitations.

{\it Notations:} In the remainder of the article, uppercase characters will stand for matrices, lowercase for scalars or vectors. The transpose operation will be denoted $(\cdot)^\trans$. The multivariate Gaussian distribution of mean $\mu$ and covariance $C$ will be denoted $\mathcal N(\mu,C)$. The notation $V=\{V_{ij}\}_{i=1,j=1}^{n,T}$ denotes the matrix with $(i,j)$-entry $V_{ij}$ (scalar or matrix), $1\leq i\leq n$, $1\leq j\leq T$, while $\{V_i\}_{i=1}^n$ is the row-wise concatenation of the $V_i$'s and $\{V_j\}_{j=1}^T$ the column-wise concatenation of the $V_j$'s. We further introduce the notation $(x)^+\equiv \max(x,0)$. For random or deterministic matrices $X_n$ and $Y_n\in\RR^{n\times n}$, the notation $X_n\leftrightarrow Y_n$ stands for $\frac1n\tr A_n (X_n-Y_n)\to 0$ and $a_n^\trans (X_n-Y_n)b_n\to 0$, almost surely, for every deterministic matrix $A_n$ and vectors $a_n$, $b_n$ having bounded norm (spectral norm for matrices and Euclidean norm for vectors); for $X_n,Y_n\in\RR$ scalar, the notation will simply mean that $X_n-Y_n\to 0$ almost surely. The notation $\rho(X)$ will denote the spectral radius of matrix $X$, while $\|X\|$ will denote its operator norm (and for vectors, $\|x\|$ is the Euclidean norm).

\section{Main Results}
\label{sec:results}

We consider here an echo-state neural network constituted of $n$ nodes, with state $x_t\in\RR^n$ at time instant $t$, connectivity matrix $W\neq 0$, and input source sequence $\ldots,u_{-1},u_0,u_1,\ldots \in\RR$. The state evolution is given by the linear recurrent equation
\begin{align*}
	x_{t+1} &= Wx_t + m u_{t+1} + \eta \varepsilon_{t+1}
\end{align*}
for all $t\in\ZZ$, in which $\eta>0$ and $\varepsilon_t\sim \mathcal N(0,I_n)$, while $m\in\RR^n$ is the input-to-network connectivity.

Our first objective is to understand the training performance of such a network. To this end, we shall focus on a (training) time window $\{0,\ldots,T-1\}$ and will denote $X=\{x_j\}_{j=0}^{T-1}\in\RR^{n\times T}$ as well as $A=MU$, $M\in\RR^{n\times T}$, $U\in\RR^{T\times T}$, where 
\begin{align*}
	M &\equiv \left\{ W^j m \right\}_{j=0}^{T-1}\\
	U &\equiv \frac1{\sqrt{T}} \left\{ u_{j-i} \right\}_{i,j=1}^T.
\end{align*}
With these notations, we especially have $X=\sqrt{T}(A+Z)$, where $Z=\frac{\eta}{\sqrt{T}} \{ \sum_{k=0}^\infty W^k \varepsilon_{j-k} \}_{j=0}^{T-1}$.

For $X$ to be properly defined (at least almost surely so), we shall impose the following hypothesis.
\begin{assumption}[Spectral Norm]
	\label{ass:spectral_radius}
	The spectral norm $\|W\|$ of $W$ satisfies $\|W\|<1$.
\end{assumption}

Note that this constraint is in general quite strong and it is believed (following the insights of previous works \cite{JAE01b}) that for many model choices of $W$, it can be lighten to merely requiring that the spectral {\it radius} $\rho(W)$ be smaller than one. Nonetheless, in the course of the article, we shall often take $W$ to be such that both its spectral norm and spectral radius coincide.

\subsection{Training Performance}
\label{sec:training}

The training step consists in teaching the network to obtain a specific output sequence $r=\{r_j\}_{j=0}^{T-1}$ out of the network, when fed by a corresponding input vector $u=\{u_j\}_{j=0}^{T-1}$ over the time window. To this end, unlike conventional neural networks, where $W$ is adapted to $u$ and $r$, ESN's adopt the strategy to solely enforce an output link from the network to a sink (or readout). Letting $\omega=\{\omega_i\}_{i=1}^n$ be the network-to-sink connectivity vector, we shall consider here that $\omega$ is obtained as the (least-square) minimizer of $\|X^\trans \omega - r\|^2$. When $T>n$, we have
\begin{align}
	\label{eq:omega_T>n}
	\omega &\equiv \left(XX^\trans\right)^{-1}Xr
\end{align}
which is almost surely well-defined (since $\eta>0$) or, when $T\leq n$,
\begin{align}
	\label{eq:omega_T<n}
	\omega &\equiv X\left(X^\trans X\right)^{-1}r.
\end{align}
The per-input mean-square error in training associated with the couple $(u,r)$ for the ESN under study is then defined as
\begin{align}
	\label{eq:E}
	E_\eta(u,r) &\equiv \frac1T\left\| r - X^\trans \omega \right\|^2
\end{align}
which is identically zero when $T\leq n$.

Our first objective is to study precisely the random quantity $E_\eta(u,r)$ for every given $W$ and noise variance $\eta^2$ in the limit where $n\to\infty$. Our scaling hypotheses are as follows.
\begin{assumption}[Random Matrix Regime]
	\label{ass:rmt}
	The following conditions hold:
	\begin{enumerate}
		\item $\limsup_n n/T < \infty$
		\item $\limsup_n\|AA^\trans\|<\infty$.
	\end{enumerate}
\end{assumption}
That is, according to Item~1, we allow $n$ to grow with $T$. Also, from Item~2, we essentially allow $u_t$ to be of order $O(1)$ (unless $u$ is sparse and then $u_t$ may be as large as $O(\sqrt{T})$) when $m$ remains of bounded Euclidean norm. Under this setting, and along with Assumption~\ref{ass:spectral_radius}, we shall thus essentially require all neural connections to be of order $O(n^{-\frac12})$ while all input and output data (constituents of $u$ and $r$) shall be in general of order $O(1)$.

For every square symmetric matrix $B\in\RR^{n\times n}$, a central quantity in random matrix theory is the resolvent $(B-zI_n)^{-1}$ defined for every $z\in\CC\setminus \mathcal S_B$, with $\mathcal S_B\subset \RR$ the support of the eigenvalues of $B$. Here, letting $z=-\gamma$ for some $\gamma>0$, it is particularly convenient to make the following observation.
\begin{lemma}[Training MSE and resolvent]
	\label{lem:MSE-train}
	For $\gamma>0$, let $\tilde{Q}_\gamma \equiv (\frac1T X^\trans X + \gamma I_T)^{-1}$. Then we have, for $E_\eta(u,r)$ defined as in \eqref{eq:E},
	\begin{align*}
		E_\eta(u,r) &= \lim_{\gamma \downarrow 0} \gamma \frac1T r^\trans \tilde{Q}_\gamma r.
	\end{align*}
\end{lemma}

Our first technical result provides an asymptotically tight approximation for $\tilde{Q}_\gamma$ for every $\gamma>0$. Recall that, for $X_n,Y_n\in\RR^{n\times n}$, the notation $X_n\leftrightarrow Y_n$ means that, for every deterministic and bounded norm matrix $A_n$ or vector $a_n$, $b_n$, $\frac1n\tr A_n(X_n-Y_n)\to 0$ and $a_n^\trans(X_n-Y_n)b_n\to 0$, almost surely.
\begin{theorem}[Deterministic Equivalent]
	\label{th:deteq1}
	Let Assumptions~\ref{ass:spectral_radius}--\ref{ass:rmt} hold. For $\gamma>0$, let also $Q_\gamma\equiv (\frac1TXX^\trans + \gamma I_n)^{-1}$ and $\tilde{Q}_\gamma\equiv (\frac1TX^\trans X+ \gamma I_T)^{-1}$. Then, as $n\to\infty$, the following approximations hold:
	\begin{align*}
		Q_\gamma &\leftrightarrow \bar{Q}_\gamma \equiv \frac1\gamma \left( I_n + \eta^2 \tilde{R}_\gamma + \frac1\gamma A \left( I_T + \eta^2 R_\gamma\right)^{-1} A^\trans \right)^{-1} \\
		\tilde{Q}_\gamma &\leftrightarrow \bar{\tilde Q}_\gamma \equiv \frac1\gamma \left( I_T + \eta^2 R_\gamma + \frac1\gamma A^\trans \left( I_n + \eta^2 \tilde{R}_\gamma\right)^{-1} A \right)^{-1}
	\end{align*}
	where $R_\gamma\in\RR^{T\times T}$ and $\tilde{R}_\gamma\in\RR^{n\times n}$ are solutions to the system of equations
	\begin{align*}
		R_\gamma &= \left\{ \frac1T\tr \left(S_{i-j}\bar{Q}_\gamma\right) \right\}_{i,j=1}^T \\
		\tilde{R}_\gamma &= \sum_{q=-\infty}^\infty \frac1T\tr \left( J^q \bar{\tilde Q}_\gamma\right) S_q
	\end{align*}
	with $[J^q]_{ij}\equiv {\bm\delta}_{i+q,j}$ and $S_q\equiv \sum_{k\geq 0} W^{k+(-q)^+}(W^{k+q^+})^\trans$.\footnote{Note that $\tr (J^q B)$ is merely $\tr (J^q B)=\sum_{i=1+q^+}^{T-q^+}[B_{i,i+q}]$.}
\end{theorem}

\begin{remark}[On Theorem~\ref{th:deteq1}]
	\label{rem:R,tR}
	Theorem~\ref{th:deteq1} is in fact valid under more general assumptions than in the present setting. In particular, $A$ may be any deterministic matrix satisfying Assumption~\ref{ass:rmt}. However, when $A=MU$, an important phenomenon arises, which is that $A$ behaves similar to a low-rank matrix, since, by Assumption~\ref{ass:spectral_radius}, only $o(n)$ columns of $M$ have non-vanishing norm. As such, by a low-rank perturbation argument, it can be shown that the term $\bar{Q}_\gamma$ in the expression of $R_\gamma$ and the term $\bar{\tilde{Q}}_\gamma$ in the expression of $\tilde{R}_\gamma$ can be replaced by $\gamma^{-1}(I_n+\eta^2\tilde{R}_\gamma)^{-1}$ and $\gamma^{-1}(I_T+\eta^2 R_\gamma)^{-1}$, respectively. As such, $R_\gamma$ and $\tilde{R}_\gamma$ only depend on the matrix $W$ and the parameter $\eta^2$, and are thus asymptotically independent of the input data matrix $U$. Note also in passing that, while $\tilde{R}_\gamma$ is defined with a sum over $q=-\infty$ to $\infty$, this summation is empty for all $|q|\geq T$.
\end{remark}

In order to evaluate the training mean-square error $E_\eta(u,r)$ from Lemma~\ref{lem:MSE-train}, one must extend Theorem~\ref{th:deteq1} uniformly over $\gamma$ approaching zero. This can be guaranteed under the following additional assumption.
\begin{assumption}[Network size versus training time]
	\label{ass:c}
	As $n\to\infty$, $n/T\to c\in [0,1)\cup (1,\infty)$.
\end{assumption}

Under Assumption~\ref{ass:c}, two scenarios must be considered. Either $c<1$ or $c>1$. In the former case, we can show that, as $\gamma\downarrow 0$, $R_\gamma$ and $\gamma \tilde{R}_\gamma$ have well defined limits. Besides, it appears that the limit of $\eta^2R_\gamma$ does not depend on $\eta^2$, so that we shall denote $\mathcal R$ and $\tilde{\mathcal R}$ the limits of $\eta^2R_\gamma$ and $\gamma \tilde{R}_\gamma$, as $\gamma\downarrow 0$, respectively. Similarly, $\eta^2\bar{Q}_\gamma$ and $\gamma\bar{\tilde{Q}}_\gamma$ converge to well defined limits, denoted respectively $\mathcal Q$ and $\tilde{\mathcal Q}$. Symmetrically, for $c>1$, as $\gamma\downarrow 0$, $\gamma R_\gamma$ and $\eta^2 \tilde{R}_\gamma$ have well-behaved limits which we shall also refer to as $\mathcal R$ and $\tilde{\mathcal R}$; similarly, $\gamma\bar{Q}_\gamma$ and $\eta^2\bar{\tilde{Q}}_\gamma$ converge to non trivial limits again denoted $\mathcal Q$ and $\tilde{\mathcal Q}$. These results are gathered in the following proposition.

\begin{proposition}[Small $\gamma$ limit of Theorem~\ref{th:deteq1}]
	\label{prop:fundeq}
	Let Assumptions~\ref{ass:spectral_radius}--\ref{ass:c} hold. For all large $n$, define $\mathcal R$ and $\tilde{\mathcal R}$ a pair of solutions of the system
	\begin{align*}
		\mathcal R &= c \left\{ \frac1n\tr \left( S_{i-j} \left( {\bm\delta}_{c>1} I_n+ \tilde{\mathcal R}\right)^{-1} \right) \right\}_{i,j=1}^T \\
		\tilde{\mathcal R} &= \sum_{q=-\infty}^\infty \frac1T\tr \left( J^q ( {\bm\delta}_{c<1} I_T+\mathcal R)^{-1} \right) S_q.
	\end{align*}
	Subsequently define
	\begin{align*}
		\tilde{\mathcal Q} &\equiv \left( {\bm\delta}_{c<1} I_T + \mathcal R + \frac1{\eta^2}A^\trans \left({\bm\delta}_{c>1} I_n+\tilde{\mathcal R}\right)^{-1}A \right)^{-1} \\
		\mathcal Q &\equiv \left( {\bm\delta}_{c>1} I_n+\tilde{\mathcal R} + \frac1{\eta^2}A\left( {\bm\delta}_{c<1} I_T+\mathcal R \right)^{-1}A^\trans  \right)^{-1}.
	\end{align*}
	Then, with the definitions of Theorem~\ref{th:deteq1}, we have the following results.
	\begin{enumerate}
		\item If $c<1$, then in the limit $\gamma\downarrow 0$, $\eta^2 R_\gamma \to \mathcal R$, $\gamma \tilde{R}_\gamma \to \tilde{\mathcal R}$, $\eta^2\bar{Q}_\gamma\to \mathcal Q$, and $\gamma\bar{\tilde{Q}}_\gamma\to \tilde{\mathcal Q}$.
		\item If $c>1$, then in the limit $\gamma\downarrow 0$, $\gamma R_\gamma \to \mathcal R$, $\eta^2 \tilde{R}_\gamma \to \tilde{\mathcal R}$, $\gamma\bar{Q}_\gamma\to \mathcal Q$, and $\eta^2\bar{\tilde{Q}}_\gamma\to \tilde{\mathcal Q}$.
	\end{enumerate}
\end{proposition}

With these notations, we now have the following result.
\begin{corollary}[Training MSE for $n<T$]
	\label{cor:MSEtrain}
	Let Assumptions~\ref{ass:spectral_radius}--\ref{ass:c} hold and let $r\in\RR^T$ be of $O(\sqrt{T})$ Euclidean norm. Then, with $E_\eta(u,r)$ defined in \eqref{eq:E}, as $n\to\infty$,
	\begin{align*}
		E_\eta(u,r) &\leftrightarrow\left\{ \begin{array}{ll} (1/T) r^\trans \tilde{\mathcal Q} r &,~c<1 \\ 0 &,~c>1. \end{array}\right.
	\end{align*}
\end{corollary}

\medskip

It is interesting at this point to discuss the a priori involved expression of Proposition~\ref{prop:fundeq} and Corollary~\ref{cor:MSEtrain}. Let us concentrate on the interesting $c<1$ case. To start with, observe that $\mathcal R$ and $\tilde{\mathcal R}$ are deterministic matrices which only depend on $W$ through the $S_q$ matrices so that the only dependence of $E_\eta(u,r)$ on the noise variance $\eta^2$ lies explicitly in the expression of $\tilde{\mathcal Q}$. Now, making $A^\trans \tilde{\mathcal R}^{-1}A$ explicit, we have the following telling limiting expression for $E_\eta(u,r)$
\begin{align}
	\label{eq:Etrain_approx}
	E_\eta(u,r) \leftrightarrow \frac1T r^\trans \left( I_T + \mathcal R + \frac1{\eta^2}U^\trans \left\{ m^\trans (W^i)^\trans \tilde{\mathcal R}^{-1} W^j m \right\}_{i,j=0}^{T-1} U \right)^{-1}r.
\end{align}
Recalling that $\tilde{\mathcal R}$ is a linear combination of the matrices $S_q=W^{(-q)^+}S_0W^{(q^+)}$, with $S_0=\sum_{k\geq 0}W^k(W^k)^\trans$, the expression $\frac1{\eta^2}m^\trans (W^i)^\trans \tilde{\mathcal R}^{-1} W^j m$ is strongly reminiscent of the Fisher memory curve $f:\NN\to\RR$ of the ESN, introduced in \cite{GAN08} and defined by $f(k)=\frac1{\eta^2} m^\trans (W^k)^\trans S_0^{-1}W^k m$. The Fisher memory curve $f(k)$ qualifies the ability of a $k$-step behind past input to influence the ESN at present time. Correspondingly, it appears here that the ability of the ESN to retrieve the desired expression of $r$ from input $u$ is importantly related to the matrix $\{\frac1{\eta^2} m^\trans (W^i)^\trans \tilde{\mathcal R}^{-1} W^j m \}_{i,j=0}^{T-1}$. As a matter of fact, for $c=0$ (thus for a long training period), note that $\mathcal R=0$ while $\tilde{\mathcal R}=S_0$ and we then find in particular
\begin{align*}
	E_\eta(u,r) &\leftrightarrow \frac1Tr^\trans \left( I_T + \frac1{\eta^2} U^\trans \left\{ m^\trans (W^i)^\trans S_0^{-1}W^j m\right\}_{i,j=0}^{T-1}U\right)^{-1}r.
\end{align*}

\medskip

Pushing further our discussion on $\mathcal R$ and $\tilde{\mathcal R}$, it is interesting to intuit their respective structures. Observe in particular that $\mathcal R_{ij}$ depends only on $i-j$ and thus $\mathcal R$ is a Toeplitz matrix. Besides, since $\tr B=\tr B^\trans$ for square matrices $B$, from $S_{i-j}^\trans=S_{j-i}$ it comes that $\mathcal R_{ij}=\mathcal R_{ji}$. Also note that, since $\rho(W^q)=\rho(W)^q$ decays exponentially as $q\to\infty$, it is expected that $\mathcal R_{i,i+q}$ decays exponentially fast for large $q$. As a consequence, $\mathcal R$ is merely defined by $o(n)$ first entries of its first row. 

From the results of \cite{GRA06} on Toeplitz versus circulant matrices, it then appears that, for every deterministic matrix $B$, $\frac1T\tr B\mathcal R^{-1}$ is well approximated by $\frac1T\tr B\mathcal R_{c}^{-1}$ for $\mathcal R_{c}$ a circulant matrix approximation of $\mathcal R$. Since circulant matrices are diagonalizable in a Fourier basis, so are their inverses and then, {\it as far as normalized traces are concerned}, $(I_T+\mathcal R)^{-1}$ can be seen as approximately Toeplitz with again decaying behavior away from the main diagonals. Although slightly ambiguous, this approximation still makes it that the trace $\frac1T\tr J^q(I_T+\mathcal R)^{-1}$ appearing in the expression of $\tilde{\mathcal R}$, is well approximated by any value $[(I_T+\mathcal R)^{-1}]_{i,i+q}$ for $i$ sufficiently far from $1$ and $T$, and decays to zero as $q$ grows large. This, and the fact that $S_q$ also decays exponentially fast in norm allows us to conclude that $\tilde{\mathcal R}$ can be seen as a decaying weighted sum of $o(n)$ matrices $S_q$.


\medskip

As shall be shown in Section~\ref{sec:applications}, for $W$ taken random with sufficient invariance properties, fundamental differences appear in the structure of $\mathcal R$ and $\tilde{\mathcal R}$ depending on whether $W$ is taken normal or not. In particular, for $W$ non-normal with left and right independent isotropic eigenvectors and $m$ deterministic or random independent of $W$, $\mathcal R$ is well approximated by a scaled identity matrix and $\{m^\trans (W^i)^\trans\tilde{\mathcal R}^{-1}W^jm\}_{i,j=0}^{T-1}$ well approximated by a diagonal matrix with exponential decay along the diagonal.

\medskip

Having a clearer understanding of Corollary~\ref{cor:MSEtrain}, a few key remarks are in order.

\begin{remark}[On the ESN stability to low noise levels]
	\label{rem:low_noise}
	It is easily seen by differentiation along $\eta^2$ that $r^\trans \tilde{\mathcal Q}r$ is an increasing function of $\eta^2$, thus having a minimum as $\eta^2\downarrow 0$. It is thus tempting to suppose that $E_\eta(u,r)$ converges to this limit in the noiseless case (i.e., for $\eta^2=0$). Such a reasoning is however hazardous and incorrect in most cases. Indeed, Corollary~\ref{cor:MSEtrain} only ensures an appropriate approximation of $E_\eta(u,r)$ for given $\eta>0$ in the limit where $n\to\infty$. Classical random matrix considerations allow one to assert slightly stronger results. In particular, for the approximation of $E_\eta(u,r)$ to hold, one may allow $\eta^2$ to depend on $n$ in such a way that $\eta^2\gg n^{-\frac12}$. This indicates that $n$ must be quite large for the ESN behavior at moderate noise levels to be understood through the random matrix method. What seems like a defect of the tool on the onset in fact sheds some light on a deeper feature of ESN's. When $\eta^2$ is of the same order of magnitude or smaller than $n^{-\frac12}$, Corollary~\ref{cor:MSEtrain} may become invalid due to the \emph{resurgence of randomness} from $\cdots,\varepsilon_{-1},\varepsilon_0,\varepsilon_1,\ldots$ Precisely, when $\eta^2$ gets small and thus the training MSE variance should decay, an opposite effect makes the MSE more random and thus possibly no longer tractable; this means in particular that, for any two independent runs of the ESN (with different noise realizations), all other parameters being fixed, the resulting MSE's might be strikingly different, making the network quite unstable. In practice, the opposition of the reduced noise variance $\eta^2$ and the resurgence of noise effects lead to various behaviors depending on the task and input data under considerations, ranging from largely increased MSE fluctuations at low $\eta^2$ to reduced fluctuations, through stabilisation of the fluctuations. In some specific cases discussed later, it might nonetheless be accepted to let $\eta^2\to 0$ irrespective of $n$ while keeping the random matrix approximation valid.
\end{remark}

\begin{remark}[Memory capacity revisited]
	\label{rem:MC}
	For $c<1$, letting $u_k=\sqrt{T}{\bm\delta}_k$ and $r_k=\sqrt{T}{\bm\delta}_{k-\tau}$ (that is, all input-output energy is gathered in a single entry), for $\tau\in\NN$, makes the ESN fill a pure delay task of $\tau$ time-steps. In this case, we find that
	\begin{align*}
		E_\eta(u,r) &\leftrightarrow \left[\left( I_T + \mathcal R + \frac1{\eta^2} \left\{ m^\trans (W^i)^\trans \tilde{\mathcal R}^{-1}W^jm \right\}_{i,j=0}^{T-1} \right)^{-1}\right]_{\tau+1,\tau+1}.
	\end{align*}
	In the particular case where, for all $i\neq j$, $\mathcal R_{ij}=o(1)$ and $m^\trans (W^i)^\trans \tilde{\mathcal R}^{-1}W^jm=o(1)$ (see Section~\ref{sec:non-Hermitian} for a practical application with random non-normal $W$), by a uniform control argument due to the fast decaying far off-diagonal elements of $\mathcal R$ and $\{m^\trans (W^i)^\trans \tilde{\mathcal R}^{-1}W^jm\}$, the training MSE is further (almost surely) well approximated as
	\begin{align*}
		E_\eta(u,r) &\leftrightarrow \frac{\eta^2}{\eta^2 (1+\mathcal R_{11})+m^\trans (W^\tau)^\trans \tilde{\mathcal R}^{-1}W^\tau m}.
	\end{align*}
	If the quantity $m^\trans (W^\tau)^\trans \tilde{\mathcal R}^{-1}W^\tau m$ remains away from zero as $n\to\infty$, then it is allowed here to say (as opposed to the general case discussed in Remark~\ref{rem:low_noise}) that $E_\eta(u,r)\to 0$ as $\eta\to 0$ and that $\eta^2/E_\eta(u,r)\sim m^\trans (W^\tau)^\trans \tilde{\mathcal R}^{-1}W^\tau m$, where we recover again a generalized form of the Fisher information curve at delay $\tau$. From this discussion and Remark~\ref{rem:low_noise}, we propose to define a novel network memory capacity metric ${\rm MC}(\tau)$, representing the inverse slope of decay of $E_\eta(\sqrt{T}{\bm\delta}_k,\sqrt{T}{\bm\delta}_{k-\tau})$ for small $\eta^2$:
	\begin{align*}
		{\rm MC}(\tau) \equiv \lim_{\eta\downarrow 0}\liminf_n \left[ \left( \eta^2(I_T + \mathcal R) + \left\{ m^\trans (W^i)^\trans \tilde{\mathcal R}^{-1}W^jm \right\}_{i,j=0}^{T-1} \right)^{-1}_{\tau+1,\tau+1}\right]^{-1}. 
	\end{align*}
\end{remark}

Practical applications of Corollary~\ref{cor:MSEtrain} to specific matrix models for $W$ shall be derived in Section~\ref{sec:applications}. Beforehand, we will study the more involved question of the test MSE performance.

\subsection{Test Performance}
\label{sec:test}

In this section, we assume $\omega\equiv \omega(X;u,r)$ has been obtained as per \eqref{eq:omega_T>n} or \eqref{eq:omega_T<n}, depending on whether $c<1$ or $c>1$. We now consider the test performance of the ESN that corresponds to its ability to map an input vector $\hat{u}\in\RR^{\hat T}$ to an expected output vector $\hat{r}\in\RR^{\hat T}$ of duration $\hat{T}$ in such a way to fulfill the same task that links $u$ to $r$. For notational convenience, all test data will be denoted with a hat mark on top.

As opposed to the training mean square error, the testing MSE, defined as
\begin{align}
	\label{eq:hatE}
	\hat{E}_\eta(u,r;\hat{u},\hat{r}) &\equiv \frac1{\hat T} \left\| \hat{r} - \hat{X}^\trans \omega \right\|^2
\end{align}
where $\hat{X}=\{\hat{x}_j\}_{j=0}^{\hat{T}-1}\in\RR^{n\times \hat{T}}$ is defined by the recurrent equation $\hat{x}_{t+1}=W\hat{x}_t+m\hat{u}_{t+1}+\eta \hat{\varepsilon}_{t+1}$, with $\hat{\varepsilon}_t\sim \mathcal N(0,I_n)$ independent of the $\varepsilon_t$'s, does not assume a similar simple form as the training MSE. Precisely, we merely have the following result.
\begin{lemma}[Testing MSE]
	For $\gamma>0$, $Q_\gamma=(\frac1TXX^\trans + \gamma I_n)^{-1}$, and $\tilde{Q}_\gamma=(\frac1TX^\trans X + \gamma I_T)^{-1}$, we have
	\begin{align*}
		\hat{E}_\eta(u,r;\hat{u},\hat{r}) &= \lim_{\gamma\downarrow 0} \frac1{\hat{T}}\|\hat r\|^2 + \frac1{T^2\hat{T}}r^\trans X^\trans Q_\gamma \hat{X}\hat{X}^\trans Q_\gamma Xr - \frac2{T\hat{T}}\hat{r}^\trans \hat{X}^\trans Q_\gamma Xr \\
		&= \lim_{\gamma\downarrow 0} \frac1{\hat{T}}\|\hat r\|^2 + \frac1{T^2\hat{T}} r^\trans \tilde{Q}_\gamma X^\trans \hat{X}\hat{X}^\trans X \tilde{Q}_\gamma r - \frac2{T\hat{T}}\hat{r}^\trans \hat{X}^\trans X\tilde{Q}_\gamma r
	\end{align*}
	with $\hat{E}_\eta(u,r;\hat{u},\hat{r})$ defined in \eqref{eq:hatE}.
\end{lemma}
If $n<T$, $Q_\gamma$ is well-defined in the limit $\gamma\downarrow 0$, while if instead $n\geq T$, then one may observe that $X^\trans Q_\gamma=\tilde{Q}_\gamma X^\trans$ with $\tilde{Q}_\gamma$ having well defined limit as $\gamma\downarrow 0$.

Technically, estimating $\hat{E}$ requires to retrieve, in a similar fashion as for Theorem~\ref{th:deteq1}, a deterministic approximation of quantities of the type $Q_\gamma X$ and $X^\trans Q_\gamma B Q_\gamma X=\tilde{Q}_\gamma X^\trans B X \tilde{Q}_\gamma$ for $B$ a matrix independent of $X$. We precisely obtain the following result.
\begin{theorem}[Second order deterministic equivalent]
	\label{th:deteq2}
	Let Assumptions~\ref{ass:spectral_radius}--\ref{ass:rmt} hold and let $B\in\RR^{n\times n}$ be a deterministic symmetric matrix of bounded spectral norm. Then, recalling the notations of Theorem~\ref{th:deteq1}, for every $\gamma>0$,
	\begin{align*}
		Q_\gamma \frac1{\sqrt{T}} X &\leftrightarrow \bar{Q}_\gamma A(I_n+\eta^2R_\gamma)^{-1} \\ 
		\frac1TX^\trans Q_\gamma B Q_\gamma X &\leftrightarrow \eta^2 \gamma^2 \bar{\tilde{Q}}_\gamma G_\gamma^{[B]} \bar{\tilde{Q}}_\gamma + (I_n+\eta^2R_\gamma)^{-1}A^\trans \bar{Q}_\gamma \left[ B + \tilde{G}_\gamma^{[B]} \right]\bar{Q}_\gamma A(I_n+\eta^2R_\gamma)^{-1}
	\end{align*}
	where $G_\gamma^{[B]}\in\RR^{T\times T}$ and $\tilde{G}^{[B]}_\gamma\in\RR^{n\times n}$ are solutions to the system of equations
	\begin{align*}
		G_\gamma^{[B]} &= \left\{ \frac1T\tr \left( S_{i-j} \bar{Q}_\gamma\left[ B + \tilde{G}^{[B]}_\gamma \right] \bar{Q}_\gamma\right) \right\}_{i,j=1}^T \\
		\tilde{G}^{[B]}_\gamma &= \sum_{q=-\infty}^\infty \eta^4 \gamma^2 \frac1T\tr \left( J^q \bar{\tilde{Q}}_\gamma G^{[B]}_\gamma \bar{\tilde{Q}}_\gamma \right) S_q .
	\end{align*}
\end{theorem}

With these results at hand, we may then determine limiting approximations of the test mean-square error under both $n<T$ and $n>T$ regimes. As in Section~\ref{sec:training}, one may observe here that, under Assumption~\ref{ass:c} with, say $c<1$, $\eta^4 G_\gamma^{[B]}$ and $\tilde{G}_\gamma^{[B]}$ both have well defined limits as $\gamma\downarrow 0$ which we shall subsequently refer to as $\mathcal G^{[B]}$ and $\tilde{\mathcal G}^{[B]}$, respectively, and the symmetrical result holds for $c>1$. Precisely, we have the following result.
\begin{proposition}[Small $\gamma$ limit of Theorem~\ref{th:deteq2}]
	\label{prop:fundeq2}
	Let Assumptions~\ref{ass:spectral_radius}--\ref{ass:c} hold and let $B\in\RR^{n\times n}$ be a deterministic symmetric matrix of bounded spectral norm. For all large $n$, define $\mathcal G^{[B]}$ and $\tilde{\mathcal G}^{[B]}$ a pair of solutions of the system
	\begin{align*}
		\mathcal G^{[B]} &= c\left\{ \frac1n\tr \left( S_{i-j} \left( {\bm\delta}_{c>1}I_n+\tilde{\mathcal R}\right)^{-1} \left[ B + \tilde{\mathcal G}^{[B]} \right] \left( {\bm\delta}_{c>1}I_n+\tilde{\mathcal R}\right)^{-1}\right) \right\}_{i,j=1}^T \\
		\tilde{\mathcal G}^{[B]} &= \sum_{q=-\infty}^\infty \frac1T\tr \left( J^q ({\bm\delta}_{c<1}I_T+\mathcal R)^{-1} \mathcal G^{[B]} ({\bm\delta}_{c<1}I_T+\mathcal R)^{-1} \right) S_q.
	\end{align*}
	Then, with the definitions of Theorem~\ref{th:deteq2}, we have the following results. 
	\begin{enumerate}
		\item If $c<1$, then in the limit $\gamma\downarrow 0$, $\eta^4 G^{[B]}_\gamma \to \mathcal G^{[B]}$ and $\tilde{G}^{[B]}_\gamma \to \tilde{\mathcal G}^{[B]}$.
		\item If $c>1$, then in the limit $\gamma\downarrow 0$, $\gamma^2 G^{[B]}_\gamma \to \mathcal G^{[B]}$ and $\tilde{G}^{[B]}_\gamma \to \tilde{\mathcal G}^{[B]}$.
	\end{enumerate}
\end{proposition}
	
Proposition~\ref{prop:fundeq2} will be exploited on the deterministic matrix $\frac1{\hat{T}}\EE[\hat{X}\hat{X}^\trans]=\eta^2S_0+\hat{A}\hat{A}^\trans$. Rather than taking $B=\eta^2S_0+\hat{A}\hat{A}^\trans$, which would induce an implicit dependence of $\mathcal G^{[B]}$ and $\tilde{\mathcal G}^{[B]}$ on $\eta^2$, we shall instead split $\eta^2S_0+\hat{A}\hat{A}^\trans$ into $\eta^2$ times $S_0$ and $\hat{A}\hat{A}^\trans$. Noticing then that $\mathcal G^{[\hat{A}\hat{A}^\trans]}$ is asymptotically the same as $\mathcal G^{[0]}$, with $0$ the all zero matrix, we may then obtain an approximation for the test mean square error. Prior to this, we need the following growth control assumptions.
\begin{assumption}[Random Matrix Regime for Test Data]
	\label{ass:rmthat}
	The following conditions hold:
	\begin{enumerate}
		\item $\limsup_n n/\hat{T}<\infty$
		\item $\limsup_n \|\hat{A}\hat{A}^\trans\|<\infty$.
	\end{enumerate}
\end{assumption}
Note in passing here that the $\min(T,\hat{T})$ first columns of $\hat{M}\in\RR^{n\times \hat{T}}$ in the definition of $\hat{A}$ and $M\in\RR^{n\times T}$ in the definition of $A$ are identical. As such, only $\hat{U}$ actually particularizes the data matrix $\hat{A}$.

With this condition, we have the following corollary of Theorem~\ref{th:deteq2}.
\begin{corollary}[Test MSE]
	\label{cor:MSEtest}
	Let Assumptions~\ref{ass:spectral_radius}--\ref{ass:rmthat} hold and let $\hat{r}\in\RR^{\hat T}$ be a vector of Euclidean norm $O(\sqrt{\hat{T}})$. Then, as $n\to\infty$, both for $c<1$ and $c>1$, we have, with the notations of Propositions~\ref{prop:fundeq}--\ref{prop:fundeq2},
	\begin{align}
		\label{eq:MSEtest}
		\hat{E}_\eta(u,r;\hat{u},\hat{r}) &\leftrightarrow \left\|\frac1{\eta^2\sqrt{T}} \hat{A}^\trans \mathcal Q A ( {\bm\delta}_{c<1}I_T+\mathcal R)^{-1}r -\frac1{\sqrt{\hat T}}\hat{r} \right\|^2 + \frac1Tr^\trans \tilde{\mathcal Q}\mathcal G\tilde{\mathcal Q}r \nonumber \\
		&+ \frac1{\eta^2T} r^\trans ({\bm\delta}_{c<1}I_T+\mathcal R)^{-1}A^\trans\mathcal Q\left[ S_0+\tilde{\mathcal G}\right]\mathcal Q A({\bm\delta}_{c<1} I_T+\mathcal R)^{-1}r
	\end {align}
	where $\mathcal G\equiv \mathcal G^{[S_0]}$ and $\tilde{\mathcal G}\equiv \tilde{\mathcal G}^{[S_0]}$.
\end{corollary}
The form of Corollary~\ref{cor:MSEtest} is more involved than that of Corollary~\ref{cor:MSEtrain} but is nonetheless quite interpretable. To start with, observe that $\mathcal G$ and $\tilde{\mathcal G}$ are again only function of $W$ and therefore quantify the network connectivity only. Then, note that only the first right-hand side term of the approximation of $\hat{E}_\eta(u,r;\hat{u},\hat{r})$ depends on $\hat{u}$ and $\hat{r}$. As such, the quality of the learned task relies mostly on this term.

If $c=0$, for all $B$, $\mathcal G^{[B]}=0$ and $\tilde{\mathcal G}^{[B]}=0$, so we have here the simplified expression
\begin{align*}
	\hat{E}_\eta(u,r;\hat{u},\hat{r}) &\leftrightarrow \left\|\frac1{\sqrt{T}} \hat{A}^\trans \left( \eta^2S_0+AA^\trans \right)^{-1} A r -\frac1{\sqrt{\hat T}}\hat{r} \right\|^2 + \frac1Tr^\trans A^\trans \left( \eta^2S_0+AA^\trans \right)^{-2} Ar.
\end{align*}

Some remarks are in order to appreciate these results.

\begin{remark}[Noiseless case]
	\label{rem:noiseless}
	As a follow-up on Remark~\ref{rem:low_noise}, note that some alternative approaches to ESN normalization assume instead that $\eta=0$ but that $\omega$ is taken to be the \emph{regularized} least-square (or ridge-regression) estimator $\omega=X(X^\trans X+\gamma I_T)^{-1}r$ with $\gamma>0$. In this case, it is easily seen that the corresponding mean-square error performance in training is given by $E^{\gamma}(u,r)\equiv \gamma^2 \frac1T r^\trans \tilde{Q}_\gamma^2 r$, which is precisely
	\begin{align*}
		E^{\gamma}(u,r) &= \frac1T r^\trans \left( I_T + \frac1{\gamma} U^\trans \left\{ m^\trans (W^i)^\trans W^j m \right\}_{i,j=0}^{T-1} U \right)^{-2}r. 
	\end{align*}
	It is interesting to parallel this (exact) expression to the approximation \eqref{eq:Etrain_approx} in which the noise variance $\eta^2$ plays the role of the regularization $\gamma$, but (i) where the two additional quantities $\mathcal R$ and $\tilde{\mathcal R}$ are present, and (ii) where the power factor of the matrix inverse is $1$ in place of $2$. As for the testing performance, we are here comparing Corollary~\ref{cor:MSEtest} to the noiseless regularized MSE 
	\begin{align*}
		E^{\gamma}(u,r;\hat{u},\hat{r})=\left\|\frac1{\sqrt{T}} \hat{A}^\trans \left( \gamma I_n+AA^\trans \right)^{-1} A r -\frac1{\sqrt{\hat T}}\hat{r}\right\|^2.
	\end{align*}
	This is again easily paralleled with the first right-hand side term in \eqref{eq:MSEtest} which, for say $c<1$, reads
	\begin{align*}
		\left\| \frac1{\sqrt{T}} \hat{A}^\trans \left( \eta^2 \tilde{\mathcal R} + A(I_T+\mathcal R)^{-1}A^\trans \right)^{-1}A (I_T+\mathcal R)^{-1}r -\frac1{\sqrt{\hat T}}\hat{r}  \right\|^2.
	\end{align*}
	Again, it is clear that $\eta^2$ plays a similar role as that of $\gamma$, and that the matrices $\mathcal R$ and $\tilde{\mathcal R}$ capture the behavior of the in-network noise.
\end{remark}

\section{Applications}
\label{sec:applications}

In this section, we shall further estimate the results of Corollary~\ref{cor:MSEtrain} and Corollary~\ref{cor:MSEtest} in specific settings for the network connectivity matrix $W$ and the input weights $m$. By leveraging specific properties of certain stochastic models for $W$ (such as invariance by orthogonal matrix product or by normality), the results of Section~\ref{sec:results} will be greatly simplified, by then providing further insights on the network performance.

\subsection{Bi-orthogonally invariant $W$}
\label{sec:non-Hermitian}

We first consider the scenario where $W$ is random with distribution invariant to left- and right-multiplication by orthogonal matrices, which we refer to as {\it bi-orthogonal invariance}. Precisely, in singular-value decomposition form, we shall write $W=U\Omega V^\trans$, where $U$, $V$, and $\Omega$ are independent and $U$, $V$ are real Haar distributed (that is, orthogonal with bi-orthogonally invariant distribution) and shall impose that the eigenvalues of $W$ remain bounded by $\sigma<1$ for all large $n$. Two classical examples of such a scenario are (i) $W$ is itself a scaled Haar matrix, in which case $\Omega=\sqrt{\sigma}I_n$ and the eigenvalues of $W$ all have modulus $\sigma$, or (ii) $W$ has independent $\mathcal N(0,\sigma^2)$ entries, in which case, according to standard random matrix results, for any $\varepsilon>0$, the eigenvalues of $W$ have modulus less than $\sigma+\varepsilon$ for all large $n$ almost surely and $W$ is clearly orthogonally invariant by orthogonal invariance of the real multivariate Gaussian distribution.

In this scenario, one can exploit the fact (arising for instance from free probability considerations \cite{BIA03}) that, for all $i\neq j$ fixed, the moments $\frac1n\tr W^i(W^j)^\trans$ vanish as $n\to\infty$. In our setting, $i$ and $j$ may however be growing with $n$, but then the fact that $\rho(W^i)\leq \sigma^i$ shall easily ensure an exponential decay of these moments. All in all, in the large $n$ setting, only the first few moments $\frac1n\tr W^i(W^i)^\trans$, $i=1,2,\ldots$, do not vanish. Although the implication is not immediate, this remark leads naturally to the intuition that the Toeplitz matrix $\mathcal R$ defined in Proposition~\ref{prop:fundeq} should be diagonal and thus proportional to the identity matrix.

At this point, we need to differentiate the cases where $c<1$ and $c>1$.

\subsubsection{Case $c<1$}

Based on the remarks above, we may explicitly solve for $\mathcal R$ and $\tilde{\mathcal R}$ to find that, in the large $n$ limit
\begin{align*}
	\mathcal R &\leftrightarrow \frac{c}{1-c}I_T \\
	\tilde{\mathcal R} &\leftrightarrow (1-c)S_0 \\
	\mathcal G^{[B]} &\leftrightarrow \frac{c}{(1-c)^3} \frac1n\tr (S_0^{-1}B) I_T \\
	\tilde{\mathcal G}^{[B]} &\leftrightarrow \frac{c}{1-c} \frac1n\tr (S_0^{-1}B) S_0. 
\end{align*}
Replacing in the expressions of both Corollaries~\ref{cor:MSEtrain}--\ref{cor:MSEtest}, we obtain the further corollary
\begin{corollary}[Orthogonally invariant case, $c<1$]
	\label{cor:unitarily_invariant}
	Let $W$ be random and left and right independently orthogonally invariant. Then, under Assumptions~\ref{ass:spectral_radius}--\ref{ass:rmthat} and with $c<1$, the following hold
	\begin{align*}
		E_{\eta}(u,r) &\leftrightarrow (1-c) \frac1Tr^\trans \left( I_T + \frac1{\eta^2} U^\trans D U \right)^{-1}r \\
		\hat{E}_{\eta}(u,r;\hat{u},\hat{r}) &\leftrightarrow \left\| \frac1{\eta^2\sqrt{T}} \hat{U}^\trans \hat{D}U \left( I_T + \frac1{\eta^2}U^\trans DU \right)^{-1}r - \frac1{\sqrt{\hat T}} \hat{r} \right\|^2 \nonumber \\ 
		&+ \frac1{1-c} \frac1Tr^\trans \left( I_T + \frac1{\eta^2} U^\trans D U \right)^{-1}r - \frac1Tr^\trans \left( I_T + \frac1{\eta^2} U^\trans D U \right)^{-2}r
	\end{align*}
	where we defined $D \equiv \left\{ m^\trans (W^i)^\trans S_0^{-1} W^j m \right\}_{i,j=0}^{T-1}$ and $\hat{D} \equiv \left\{ m^\trans (W^i)^\trans S_0^{-1} W^j m \right\}_{i,j=0}^{\hat{T}-1,T-1}$.
\end{corollary}
We see here that the matrix $D$ plays a crucial role in the ESN performance. First, from its Gram structure and the positive definiteness of $S_0$, $D$ is symmetric and nonnegative definite. This matrix has an exponential decaying profile down its rows and columns. As such, the dominating coefficients of the matrix $U^\trans D U$ lie in its upper-left corner. Recalling that the $j$-th column of $\sqrt{T}U^\trans$ is $\{u_{i-j}\}_{i=1}^T$, $U^\trans D U$ is essentially a linear combination of the outer products $\{u_{i-j}\}_{i=1}^T (\{u_{i-j'}\}_{i=1}^T)^\trans$ for small $j,j'$, that is of combinations of (outer-products of) short-time delayed versions of the input vector $u$.

\medskip

Now, it is interesting to particularize the vector $m$ and study its impact on $D$. It may be thought that taking $m$ to be one of the dominant eigenvectors of $W$ could drive the inputs towards interesting memory-capacity levels of $W$; this aspect is discussed in \cite{GAN08} where it is found that such an $m$ maximizes the integrated Fisher-memory curve. If such a real eigenvector having eigenvalue close to $\sigma$ exists, then we would find that $D_{ij}\simeq \sigma^{i+j} m^\trans S_0^{-1}m$ and thus $D$ would essentially be a rank-one matrix. As we shall discuss below, this would lead to extremely bad MSE performance in general.

If instead $m$ is chosen deterministic or random independent of $W$ with say $\|m\|=1$ (or tending to one) for simplicity, then by the trace lemma \cite[Lemma~B.26]{SIL06}, one can show that $m^\trans (W^i)^\trans S_0^{-1} W^jm \leftrightarrow \frac1n\tr W^j(W^i)^\trans S_0^{-1}$. According to our earlier discussion, this quantity vanishes for all $i\neq j$ as $n\to\infty$, and thus $D$ would now essentially be diagonal. Besides, it is clear that $\tr D\leftrightarrow 1$ and thus $D$ here plays the role of affecting a short-term memorization ability, that can be seen as a total load $1$, to the successive delayed versions of $u$. In particular, from our definition in Remark~\ref{rem:MC}, we have precisely here
\begin{align*}
	{\rm MC}(\tau) &= \frac1{1-c} \liminf_n  \frac1n\tr \left(W^\tau (W^\tau)^\trans S_0^{-1}\right)
\end{align*}
which, for the chosen $m$, is precisely the Fisher memory curve \cite{GAN08}, up to the factor $1-c$.

\begin{remark}[Haar $W$ and independent $m$]
	\label{rem:Haar}
	For $W=\sigma Z$ with $Z$ Haar distributed (orthogonal and orthogonally invariant) and $m$ independent of $Z$ and of unit norm, $D$ is asymptotically diagonal and we find precisely
	\begin{align*}
		D_{ii} &\leftrightarrow (1-\sigma^2) \sigma^{2(i-1)}
	\end{align*}
	and in particular
	\begin{align*}
		{\rm MC}(\tau) &= \frac{1-\sigma^2}{1-c}\sigma^{2\tau}.
	\end{align*}
\end{remark}

\begin{figure}[h!]
  \centering
  \begin{tikzpicture}[font=\footnotesize]
    \renewcommand{\axisdefaulttryminticks}{4} 
    \tikzstyle{every major grid}+=[style=densely dashed]       
    \tikzstyle{every axis y label}+=[yshift=-10pt] 
    \tikzstyle{every axis x label}+=[yshift=5pt]
    \tikzstyle{every axis legend}+=[cells={anchor=west},fill=white,
        at={(0.98,0.98)}, anchor=north east, font=\scriptsize ]
    \begin{semilogyaxis}[
      xmin=1,
      ymin=1e-4,
      xmax=30,
      ymax=1,
      grid=major,
      ymajorgrids=false,
      scaled ticks=true,
      xlabel={$\tau$},
      mark repeat = 2,
      mark options= {solid},
      ]
      \addplot[red,smooth,line width=0.5pt] plot coordinates{
	      (1,0.272552)(2,0.113823)(3,0.066520)(4,0.048500)(5,0.038956)(6,0.032470)(7,0.027506)(8,0.023539)(9,0.020320)(10,0.017692)(11,0.015541)(12,0.013774)(13,0.012320)(14,0.011118)(15,0.010123)(16,0.009294)(17,0.008601)(18,0.008018)(19,0.007525)(20,0.007105)(21,0.006745)(22,0.006434)(23,0.006162)(24,0.005924)(25,0.005712)(26,0.005522)(27,0.005350)(28,0.005193)(29,0.005050)(30,0.004916)(31,0.004792)(32,0.004675)(33,0.004564)(34,0.004459)(35,0.004359)(36,0.004263)(37,0.004171)(38,0.004082)(39,0.003996)(40,0.003912)(41,0.003831)(42,0.003752)(43,0.003675)(44,0.003601)(45,0.003528)(46,0.003456)(47,0.003387)(48,0.003318)(49,0.003252)(50,0.003187)(51,0.003123)(52,0.003060)(53,0.002999)(54,0.002939)(55,0.002881)(56,0.002823)(57,0.002767)(58,0.002712)(59,0.002658)(60,0.002605)(61,0.002553)(62,0.002502)(63,0.002452)(64,0.002403)(65,0.002355)(66,0.002309)(67,0.002263)(68,0.002218)(69,0.002173)(70,0.002130)(71,0.002088)(72,0.002046)(73,0.002006)(74,0.001966)(75,0.001927)(76,0.001888)(77,0.001851)(78,0.001814)(79,0.001778)(80,0.001742)(81,0.001708)(82,0.001674)(83,0.001640)(84,0.001608)(85,0.001576)(86,0.001544)(87,0.001514)(88,0.001483)(89,0.001454)(90,0.001425)(91,0.001397)(92,0.001369)(93,0.001342)(94,0.001315)(95,0.001289)(96,0.001263)(97,0.001238)(98,0.001213)(99,0.001189)(100,0.001166)
      };
      \addplot[black,densely dashed,smooth,line width=0.5pt] plot coordinates{
	      (1,0.022979)(2,0.022521)(3,0.022073)(4,0.021634)(5,0.021203)(6,0.020782)(7,0.020368)(8,0.019963)(9,0.019565)(10,0.019176)(11,0.018794)(12,0.018420)(13,0.018054)(14,0.017695)(15,0.017342)(16,0.016997)(17,0.016659)(18,0.016328)(19,0.016003)(20,0.015684)(21,0.015372)(22,0.015066)(23,0.014766)(24,0.014473)(25,0.014185)(26,0.013902)(27,0.013626)(28,0.013354)(29,0.013089)(30,0.012828)(31,0.012573)(32,0.012323)(33,0.012078)(34,0.011837)(35,0.011602)(36,0.011371)(37,0.011144)(38,0.010923)(39,0.010705)(40,0.010492)(41,0.010283)(42,0.010079)(43,0.009878)(44,0.009682)(45,0.009489)(46,0.009300)(47,0.009115)(48,0.008934)(49,0.008756)(50,0.008582)(51,0.008411)(52,0.008244)(53,0.008080)(54,0.007919)(55,0.007761)(56,0.007607)(57,0.007455)(58,0.007307)(59,0.007162)(60,0.007019)(61,0.006879)(62,0.006742)(63,0.006608)(64,0.006477)(65,0.006348)(66,0.006222)(67,0.006098)(68,0.005976)(69,0.005857)(70,0.005741)(71,0.005627)(72,0.005515)(73,0.005405)(74,0.005297)(75,0.005192)(76,0.005089)(77,0.004987)(78,0.004888)(79,0.004791)(80,0.004696)(81,0.004602)(82,0.004511)(83,0.004421)(84,0.004333)(85,0.004247)(86,0.004162)(87,0.004079)(88,0.003998)(89,0.003918)(90,0.003841)(91,0.003764)(92,0.003689)(93,0.003616)(94,0.003544)(95,0.003473)(96,0.003404)(97,0.003336)(98,0.003270)(99,0.003205)(100,0.003141)
      };
      \addplot[black,densely dashed,smooth,mark=o,smooth,line width=0.5pt] plot coordinates{
	      (1,0.190000)(2,0.153900)(3,0.124659)(4,0.100974)(5,0.081789)(6,0.066249)(7,0.053662)(8,0.043466)(9,0.035207)(10,0.028518)(11,0.023100)(12,0.018711)(13,0.015156)(14,0.012276)(15,0.009944)(16,0.008054)(17,0.006524)(18,0.005284)(19,0.004280)(20,0.003467)(21,0.002808)(22,0.002275)(23,0.001843)(24,0.001492)(25,0.001209)(26,0.000979)(27,0.000793)(28,0.000642)(29,0.000520)(30,0.000422)(31,0.000341)(32,0.000277)(33,0.000224)(34,0.000181)(35,0.000147)(36,0.000119)(37,0.000096)(38,0.000078)(39,0.000063)(40,0.000051)(41,0.000042)(42,0.000034)(43,0.000027)(44,0.000022)(45,0.000018)(46,0.000014)(47,0.000012)(48,0.000009)(49,0.000008)(50,0.000006)(51,0.000005)(52,0.000004)(53,0.000003)(54,0.000003)(55,0.000002)(56,0.000002)(57,0.000001)(58,0.000001)(59,0.000001)
      };
      \addplot[black,densely dashed, mark=square,smooth,line width=0.5pt] plot coordinates{
	      (1,0.750000)(2,0.187500)(3,0.046875)(4,0.011719)(5,0.002930)(6,0.000732)(7,0.000183)(8,0.000046)(9,0.000011)(10,0.000003)(11,0.000001)(12,0.000000)(13,0.000000)(14,0.000000)(15,0.000000)(16,0.000000)(17,0.000000)(18,0.000000)(19,0.000000)(20,0.000000)
      };
      \legend{ ${\rm MC}(\tau;W)$, ${\rm MC}(\tau;W_1^+)$, ${\rm MC}(\tau;W_2^+)$, ${\rm MC}(\tau;W_3^+)$}
    \end{semilogyaxis}
  \end{tikzpicture}
  \caption{Memory curve for $W=\diag(W_1,W_2,W_3)$, $W_j=\sigma_jZ_j$, $Z_j\in\RR^{n_j\times n_j}$ Haar distributed, $\sigma_1=.99$, $n_1/n=.01$, $\sigma_2=.9$, $n_2/n=.1$, and $\sigma_3=.5$, $n_3/n=.89$. The matrices $W_i^+$ are defined by $W_i^+=\sigma_i Z_i^+$, with $Z_i^+\in\RR^{n\times n}$ Haar distributed.}
  \label{fig:multimemory}
\end{figure}
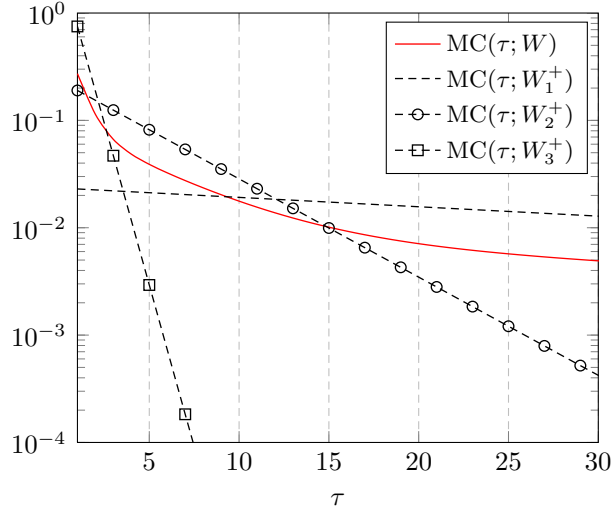

Remark~\ref{rem:Haar} can be extended to design an interesting multiple memory-mode network as follows.
\begin{remark}[Multiple memory modes]
	\label{rem:multimemory}
	Take $W$ to be the block diagonal matrix $W=\diag(W_1,\ldots,W_k)$ where, for $j=1,\ldots,k$, $W_j=\sigma_j Z_j$, $\sigma_j>0$, and $Z_j\in\RR^{n_j\times n_j}$ is Haar distributed, independent across $j$. Take then $m$ independent of $W$ with unit norm. Also assume that $n_j/n\to c_j>0$ as $n\to\infty$ and $\sum_j n_j=n$. Then we find that
	\begin{align*}
		D_{ii} &\leftrightarrow \frac{\sum_{j=1}^k c_j \sigma_j^{2(i-1)}}{\sum_{j=1}^k c_j (1-\sigma_j^2)^{-1}}
	\end{align*}
	and in particular, with ${\rm MC}(\tau)$ defined in Remark~\ref{rem:MC},
	\begin{align*}
		{\rm MC}(\tau) &= \frac1{1-c} \frac{\sum_{j=1}^k c_j \sigma_j^{2\tau}}{\sum_{j=1}^k c_j (1-\sigma_j^2)^{-1}}.
	\end{align*}
	A graph of ${\rm MC}(\tau)$ for $k=3$ is depicted in Figure~\ref{fig:multimemory}, where it clearly appears that the memory curve follows successively each one of the three modes, giving in particular more weight to short-term past inputs at first, and then smoothly providing increasingly more importance to longer term past inputs. This is reminiscent of the {\it long short-term memory} framework devised in \cite{XUE07}.
\end{remark}

\medskip

It is next interesting to study Corollary~\ref{cor:unitarily_invariant} more deeply. Let us first assume that the task to be performed, both in training and testing, consists in retrieving a mere linear combination of latest past inputs $u_{t},u_{t-1},\ldots,u_{t-(k-1)}$ for $k$ fixed. Then we may write $r=\sqrt{T}U^\trans b$ for some vector $b\in\RR^T$ with $b_j=0$ for all $j\geq k$. We then have 
\begin{align*}
	E_\eta(u,r) &\leftrightarrow (1-c) b^\trans U \left( I_T+\frac1{\eta^2}U^\trans DU \right)^{-1}U^\trans b.
\end{align*}
For $D$ positive diagonal with exponential decaying profile, $D^{-1}$ is extremely ill-conditioned and may only be used with extreme care. However, for $k$ fixed, $D^{-\frac12}b$ is well behaved as its norm is bounded by $\|b\| D_{k-1,k-1}^{-\frac12}$. We may thus write $b=D^{\frac12}(D^{-\frac12}b)$ to obtain, after basic algebraic manipulations
\begin{align*}
	E_\eta(u,r) &\leftrightarrow \eta^2(1-c) (D^{-\frac12}b)^\trans  \frac1{\eta^2} D^{\frac12}UU^\trans D^{\frac12} \left( I_T+\frac1{\eta^2}D^{\frac12}UU^\trans D^{\frac12} \right)^{-1} (D^{-\frac12}b).
\end{align*}
Since $\|A(I+A)^{-1}\|\leq 1$ for any symmetric nonnegative definite matrix $A$, we thus conclude that, for every $\eta,\varepsilon>0$, $E_\eta(u,r)\leq (1-c)\eta^2 b^\trans D^{-1}b + \varepsilon$ for all large $n$ almost surely. Thus, for sufficiently large $n$, $E_\eta(u,r)$ can be made arbitrarily small in the limit where $\eta\to 0$ and thus the task can be performed accurately. As for $\hat{E}_\eta$, note that, since $\hat{D}$ and $D$ are essentially zero away from the upper left corner and otherwise equal, if $\hat{r}=\sqrt{\hat T}\hat{U}\hat{b}$, for $\hat{b}\in\RR^{\hat T}$ having the same first $k$ entries as $b$ and zeroes next, then we find
\begin{align}
	\label{eq:hatE_r=Ub}
	\hat{E}_\eta(u,r;\hat{u},\hat{r}) &\leftrightarrow \frac{\eta^2}{1-c} (D^{-\frac12}b)^\trans  \frac1{\eta^2} D^{\frac12}UU^\trans D^{\frac12} \left( I_T+\frac1{\eta^2} D^{\frac12}UU^\trans D^{\frac12} \right)^{-1} (D^{-\frac12}b) \nonumber \\
	&+ (D^{-\frac12}b)^\trans \left( I_T+\frac1{\eta^2} D^{\frac12}UU^\trans D^{\frac12} \right)^{-1} D^{\frac12}\Delta D^{\frac12} \left( I_T+\frac1{\eta^2} D^{\frac12}UU^\trans D^{\frac12} \right)^{-1} (D^{-\frac12}b)
\end{align}
where $\Delta \equiv [\hat{U}\hat{U}^\trans]_{T\times T} - UU^\trans$, with the operator $[X]_{T\times T}$ extending (or reducing) $X$ to a $T\times T$ matrix by filling it with zeroes (or discarding last rows and columns). Note here that, for $U=\hat{U}$, $\Delta=0$ and we find that 
\begin{align}
	\label{eq:hatE_vs_E}
	\hat{E}_\eta(u,r;u,r) &\leftrightarrow \frac1{(1-c)^2} E_\eta(u,r).
\end{align}
When $\Delta\neq 0$, observe first that $\|( I_T+\eta^{-2} D^{\frac12}UU^\trans D^{\frac12})^{-1}\|\leq 1$ and thus $\hat{E}_\eta(u,r;\hat{u},\hat{r})$ remains bounded. Now, with a more subtle analysis, note that, since the product $BD^{-\frac12}b$ for any matrix $B$ only concerns the first $k$ columns of $B$, the behavior of $\hat{E}_\eta$ as $\eta\to 0$ merely depends on the behavior of the top-left $k\times k$ submatrix of $(I_T+\eta^{-2} D^{\frac12}UU^\trans D^{\frac12})^{-1}$. A block matrix inverse then reveals that the second right-hand side term of \eqref{eq:hatE_r=Ub} goes to zero as $\eta\to 0$ provided that the $k$-th largest eigenvalue of $D^{\frac12}UU^\trans D^{\frac12}$ remains away from zero as $T\to\infty$. From the structure of $U$, we thus conclude that, for $\hat{E}_\eta$ to vanish as $\eta\to 0$, it is sufficient for the vector $u$ to be sufficiently ``diverse'' in its constituents (that is, so that the first columns of $U$ remain linearly independent). An obvious counter-example is when the sequence $\ldots,u_{-1},u_0,u_1,\ldots\in\RR$ is periodic of period less than $k$. Note that the specific choice of $\hat{u}$ does not alter this behavior.

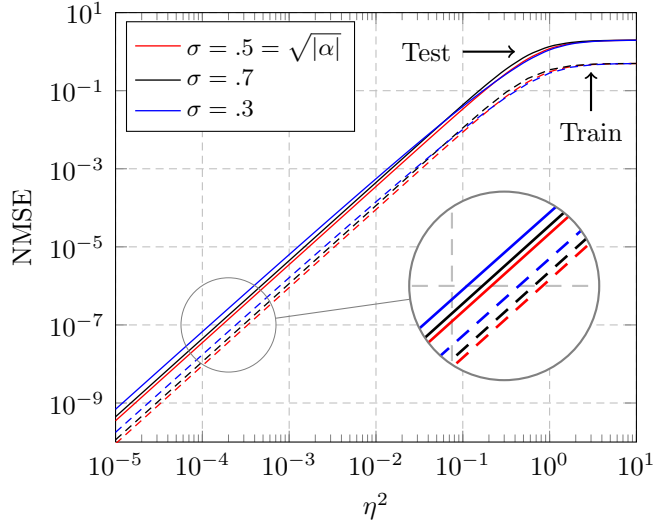
\begin{figure}[h!]
  \centering
  \begin{tikzpicture}[font=\footnotesize,spy using outlines={circle, magnification=2, connect spies}]
    \renewcommand{\axisdefaulttryminticks}{4} 
    \tikzstyle{every major grid}+=[style=densely dashed]       
    \tikzstyle{every axis y label}+=[yshift=-10pt] 
    \tikzstyle{every axis x label}+=[yshift=5pt]
    \tikzstyle{every axis legend}+=[cells={anchor=west},fill=white,
        at={(0.02,0.98)}, anchor=north west, font=\scriptsize ]
    \begin{loglogaxis}[
      xmin=1e-5,
      ymin=1e-10,
      xmax=10,
      ymax=10,
      grid=major,
      scaled ticks=true,
      xlabel={$\eta^2$},
      mark options= {solid},
      ylabel={NMSE},
      ]
      \addplot[blue,error bars/.cd,y dir=both,y explicit, error bar style={mark size=2.5pt}] plot coordinates{
      };
      \addplot[red,smooth,line width=0.5pt] plot coordinates{
	      (1.000000e-05,3.514912e-10)(1.778279e-05,1.111513e-09)(3.162278e-05,3.514912e-09)(5.623413e-05,1.111513e-08)(1.000000e-04,3.514911e-08)(1.778279e-04,1.111512e-07)(3.162278e-04,3.514907e-07)(5.623413e-04,1.111509e-06)(1.000000e-03,3.514873e-06)(1.778279e-03,1.111477e-05)(3.162278e-03,3.514585e-05)(5.623413e-03,1.111217e-04)(1.000000e-02,3.512270e-04)(1.778279e-02,1.109190e-03)(3.162278e-02,3.494906e-03)(5.623413e-02,1.094721e-02)(1.000000e-01,3.379111e-02)(1.778279e-01,1.007395e-01)(3.162278e-01,2.778622e-01)(5.623413e-01,6.549410e-01)(1.000000e+00,1.191487e+00)(1.778279e+00,1.629989e+00)(3.162278e+00,1.849139e+00)(5.623413e+00,1.931827e+00)(1.000000e+01,1.959599e+00)
      };
      \addplot[black,smooth,line width=0.5pt] plot coordinates{
	      (1.000000e-05,4.443516e-10)(1.778279e-05,1.405163e-09)(3.162278e-05,4.443516e-09)(5.623413e-05,1.405163e-08)(1.000000e-04,4.443516e-08)(1.778279e-04,1.405163e-07)(3.162278e-04,4.443515e-07)(5.623413e-04,1.405162e-06)(1.000000e-03,4.443505e-06)(1.778279e-03,1.405152e-05)(3.162278e-03,4.443407e-05)(5.623413e-03,1.405054e-04)(1.000000e-02,4.442423e-04)(1.778279e-02,1.404069e-03)(3.162278e-02,4.432594e-03)(5.623413e-02,1.394310e-02)(1.000000e-01,4.337095e-02)(1.778279e-01,1.304808e-01)(3.162278e-01,3.587098e-01)(5.623413e-01,8.090454e-01)(1.000000e+00,1.352231e+00)(1.778279e+00,1.721774e+00)(3.162278e+00,1.885667e+00)(5.623413e+00,1.944323e+00)(1.000000e+01,1.963653e+00)
      };
      \addplot[blue,smooth,line width=0.5pt] plot coordinates{
	      (1.000000e-05,6.916745e-10)(1.778279e-05,2.175550e-09)(3.162278e-05,6.835822e-09)(5.623413e-05,2.145125e-08)(1.000000e-04,6.721177e-08)(1.778279e-04,2.101949e-07)(3.162278e-04,6.558208e-07)(5.623413e-04,2.040587e-06)(1.000000e-03,6.326837e-06)(1.778279e-03,1.953680e-05)(3.162278e-03,6.000329e-05)(5.623413e-03,1.831234e-04)(1.000000e-02,5.538242e-04)(1.778279e-02,1.656911e-03)(3.162278e-02,4.880204e-03)(5.623413e-02,1.409713e-02)(1.000000e-01,3.951124e-02)(1.778279e-01,1.062592e-01)(3.162278e-01,2.665510e-01)(5.623413e-01,6.002508e-01)(1.000000e+00,1.112395e+00)(1.778279e+00,1.576625e+00)(3.162278e+00,1.826288e+00)(5.623413e+00,1.923799e+00)(1.000000e+01,1.956972e+00)
      };
      \addplot[red,smooth,densely dashed,line width=0.5pt] plot coordinates{
	      (1.000000e-05,9.033667e-11)(1.778279e-05,2.856696e-10)(3.162278e-05,9.033667e-10)(5.623413e-05,2.856696e-09)(1.000000e-04,9.033666e-09)(1.778279e-04,2.856695e-08)(3.162278e-04,9.033656e-08)(5.623413e-04,2.856686e-07)(1.000000e-03,9.033573e-07)(1.778279e-03,2.856610e-06)(3.162278e-03,9.032877e-06)(5.623413e-03,2.855981e-05)(1.000000e-02,9.027252e-05)(1.778279e-02,2.851010e-04)(3.162278e-02,8.983986e-04)(5.623413e-02,2.814276e-03)(1.000000e-01,8.685435e-03)(1.778279e-01,2.587409e-02)(3.162278e-01,7.125079e-02)(5.623413e-01,1.674847e-01)(1.000000e+00,3.035947e-01)(1.778279e+00,4.140976e-01)(3.162278e+00,4.690718e-01)(5.623413e+00,4.897708e-01)(1.000000e+01,4.967176e-01)
      };
      \addplot[black,densely dashed,densely dashed,line width=0.5pt] plot coordinates{
	      (1.000000e-05,1.142027e-10)(1.778279e-05,3.611407e-10)(3.162278e-05,1.142027e-09)(5.623413e-05,3.611407e-09)(1.000000e-04,1.142027e-08)(1.778279e-04,3.611406e-08)(3.162278e-04,1.142027e-07)(5.623413e-04,3.611404e-07)(1.000000e-03,1.142024e-06)(1.778279e-03,3.611377e-06)(3.162278e-03,1.141998e-05)(5.623413e-03,3.611116e-05)(1.000000e-02,1.141737e-04)(1.778279e-02,3.608515e-04)(3.162278e-02,1.139146e-03)(5.623413e-02,3.582824e-03)(1.000000e-01,1.114071e-02)(1.778279e-01,3.348767e-02)(3.162278e-01,9.189018e-02)(5.623413e-01,2.065936e-01)(1.000000e+00,3.440777e-01)(1.778279e+00,4.370834e-01)(3.162278e+00,4.781956e-01)(5.623413e+00,4.928889e-01)(1.000000e+01,4.977288e-01)
      };
      \addplot[blue,densely dashed,densely dashed,line width=0.5pt] plot coordinates{
	      (1.000000e-05,1.777501e-10)(1.778279e-05,5.590723e-10)(3.162278e-05,1.756625e-09)(5.623413e-05,5.512191e-09)(1.000000e-04,1.727048e-08)(1.778279e-04,5.400904e-08)(3.162278e-04,1.685098e-07)(5.623413e-04,5.243117e-07)(1.000000e-03,1.625636e-06)(1.778279e-03,5.019600e-06)(3.162278e-03,1.541469e-05)(5.623413e-03,4.703207e-05)(1.000000e-02,1.422172e-04)(1.778279e-02,4.254794e-04)(3.162278e-02,1.253115e-03)(5.623413e-02,3.618945e-03)(1.000000e-01,1.013891e-02)(1.778279e-01,2.724796e-02)(3.162278e-01,6.826975e-02)(5.623413e-01,1.534697e-01)(1.000000e+00,2.835587e-01)(1.778279e+00,4.006794e-01)(3.162278e+00,4.633470e-01)(5.623413e+00,4.877623e-01)(1.000000e+01,4.960605e-01)
      };
      \legend{ {$\sigma=.5=\sqrt{|\alpha|}$}, {$\sigma=.7$}, {$\sigma=.3$} }
      \draw[->,thick] (axis cs:1e-1,1) -- (axis cs:4e-1,1) node [left,pos=0,font=\footnotesize] {Test};
      \draw[->,thick] (axis cs:3,.03) -- (axis cs:3,.3) node [below,pos=0,font=\footnotesize] {Train};
      \coordinate (spypoint) at (axis cs:2e-4,1e-7);
      \coordinate (magnifyglass) at (axis cs:3e-1,1e-6);
    \end{loglogaxis}
    \spy [gray, size=2.5cm] on (spypoint) in node[fill=white] at (magnifyglass);
  \end{tikzpicture}
  \caption{Optimal $\sigma$ choice for $r_t=\sum_{i\geq 0} u_{t-i}b_i$, $b_i=\alpha^{i-1}$ with $\alpha=-.25$, $u$ i.i.d.\@ zero mean Gaussian, $W$ Haar distributed, $n=200$, $T=\hat{T}=400$.}
  \label{fig:optimal_sigma}
\end{figure}

The discussion above leads to interesting practical considerations that may help improve the design of an ESN.
\begin{remark}[Selecting $W$ based on delayed correlations]
	Note that, in the aforementioned formulas, the quantity $b^\trans D^{-1}b$ with $b$ defined by $r=U^\trans b$ appears as a fundamental quantity bounding the training and testing MSE. In practical settings where $r$ is not a pure linear combinations of delayed versions of $u$, it may nonetheless be useful to obtain an estimate $\hat{b}$ of the closest approximation of $r$ by delays of $u$, in such a way that $\hat{b}^\trans D^{-1}b$ be small. One may for instance let
	\begin{align*}
		\hat{b} &= (UU^\trans+\gamma I_T)^{-1}Ur
	\end{align*}
	for some regularization parameter $\gamma\geq 0$ (if needed), and parametrize $W$ so that $\hat{b}^\trans D^{-1}\hat{b}$ is minimal. For instance, if $\hat{b}_i=\alpha^{i-1}$ for some $\alpha\in (-1,1)$, it is easily shown that an optimal choice for $W=\sigma Z$ with $Z$ Haar is to take $\sigma^2=|\alpha|$. This scenario is illustrated in Figure~\ref{fig:optimal_sigma}, where the theoretical approximations for the testing and training normalized MSE are depicted for various choices of $\sigma^2$. For less obvious values of $\hat{b}$, a more elaborate multi-memory matrix $W$, as introduced in Remark~\ref{rem:multimemory}, can be used, with proper setting of the parameters $n_i$ and $\sigma_i$. 
\end{remark}

\begin{remark}[Memory Capacity for Stationary Inputs]
	Let $W$ be orthogonally invariant and $m$ random, so that $D$ is diagonal in the limit. Further assume the sequence $u$ is an auto-regressive Gaussian process, so that we may write $u=C^{\frac12}\tilde{u}$ with $\tilde{u}$ having independent zero mean unit variance Gaussian entries and $C$ a Toeplitz covariance matrix with $C_{ab}=q^{|b-a|}$ for some $q\in[0,1)$. Then, for the $\tau$-delay memory task, i.e., $r_t=u_{t-\tau}$ with $\tau$ fixed, we find that
	\begin{align*}
		E_\eta(u,r) &\leftrightarrow \eta^2\frac{1-c}{D_{\tau+1,\tau+1}} \left[ 1 - \left[ \left( I_T + \frac1{\eta^2} \left\{ \sqrt{D_{ii}} q^{|i-j|} \sqrt{D_{jj}} \right\}_{i,j=0}^{T-1} \right)^{-1} \right]_{\tau+1,\tau+1}\right].
	\end{align*}
	Since $q<1$, the matrix $\{q^{|i-j|}\}_{ij}$ has its smallest eigenvalue asymptotically far from zero (see e.g., \cite{GRA06} for arguments) so that the right-hand side inner bracket vanishes as $\eta^2\to 0$ and we thus have, for small $\eta^2$ 
	\begin{align*}
		\eta^{-2} E_\eta(u,r) &\leftrightarrow \frac{1-c}{D_{\tau+1,\tau+1}} + o(\eta^2).
	\end{align*}
	As a consequence, the memory task is performed irrespective of the smoothness of $u$, so that $u$ can be assumed composed of i.i.d.\@ elements (i.e., $q=0$). Observe that this leads to the same performance as the memory task considering $u=\sqrt{T}{\bm\delta}_0$ defining the memory capacity in Remark~\ref{rem:MC}. Of course, if instead $q=1$, then the matrix in curly brackets would have unit rank and the previous conclusions would fail (in this case $u$ is a constant vector).
\end{remark}

\medskip 

Expression~\eqref{eq:hatE_vs_E} also provides us an opportunity to open a short parenthesis on the effect of $c$ on the training and testing MSE. From Corollary~\ref{cor:unitarily_invariant}, it appears that, while $E_\eta$ is minimal for $c=1$, $\hat{E}_\eta$ is minimal for $c=0$. The former observation is clear from the fact that $\omega$ is a least-square regressor, but the latter observation is less trivial. As a matter of fact, note that, even if $\hat{U}=U$ and $\hat{r}=r$, in the limit of $\eta>0$ fixed and $c\to 1$, $\hat{E}_\eta$ becomes arbitrarily large. The reason for this seemingly counter-intuitive effect (after all, we merely ask the ESN to reproduce the exact learned sequence) lies in the fact that $\omega$ is built upon the network noise realization during training, while during testing a new noise realization is produced. As such, training an ESN of size almost equal to $T$ produces dramatic effects on testing. However, this has the positive effect of strongly reducing over fitting. Of course, in practical settings, there exists an interplay between $\eta^2$ that drives both MSE's to zero as $\eta\to 0$ and $c$ that reduces overfitting as it tends to $1$.

\medskip

Coming back to the approximations of $E_\eta$ and $\hat{E}_\eta$, note now that if $D$ is a rank-one matrix, then we may write $D=dd^\trans$ for some vector $d\in\RR^T$ having exponentially vanishing entries. In this case, we find, again after standard algebraic calculus, that
\begin{align*}
	E_\eta(u,r) &\leftrightarrow (1-c) \left( \frac1T\|r\|^2 - \frac{  \frac1T |d^\trans U r|^2}{\eta^2+|d^\trans U|^2}\right).
\end{align*}
Taking as above $r=\sqrt{T}Ub$, this is $E_\eta(u,r)\leftrightarrow (1-c) \left( b^\trans UU^\trans b - \frac{ |d^\trans U U^\trans b|^2}{\eta^2+d^\trans UU^\trans d}\right)$. By Cauchy-Schwarz inequality, this quantity, even in the limit $\eta^2\to 0$, cannot vanish unless $b=d$. As such, the ESN will only adequately fulfill a single task, which depends on the network configuration itself through $d$. A similar reasoning can be made on $\hat{E}_\eta$ revealing the same shortcomings.

\medskip

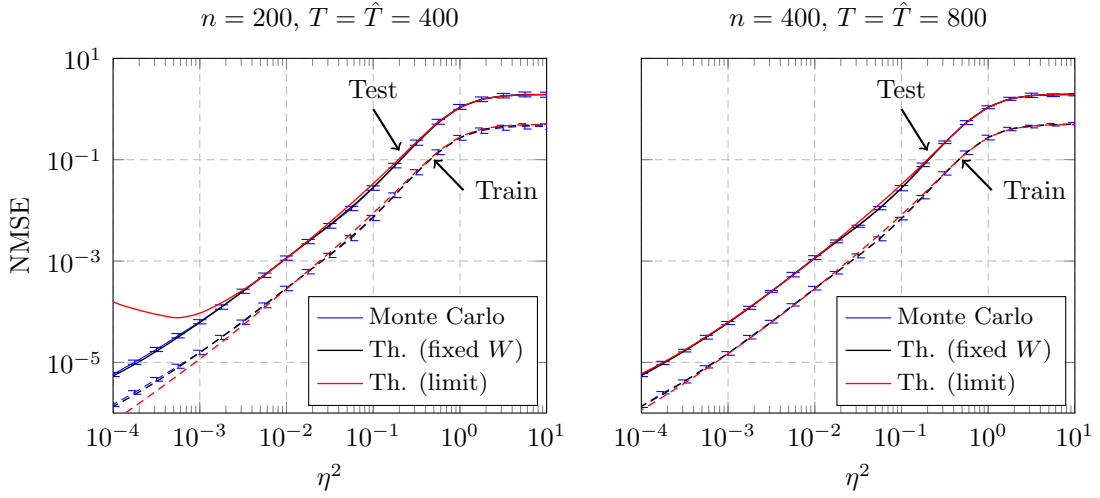
\begin{figure}[h!]
  \centering
  \begin{tabular}{cc}
  \begin{tikzpicture}[font=\footnotesize]
    \renewcommand{\axisdefaulttryminticks}{4} 
    \tikzstyle{every major grid}+=[style=densely dashed]       
    \tikzstyle{every axis y label}+=[yshift=-10pt] 
    \tikzstyle{every axis x label}+=[yshift=5pt]
    \tikzstyle{every axis legend}+=[cells={anchor=west},fill=white,
        at={(0.98,0.02)}, anchor=south east, font=\scriptsize ]
    \begin{loglogaxis}[
      title={$n=200$, $T=\hat{T}=400$},
      width=.45\linewidth,
      xmin=1e-4,
      ymin=1e-6,
      xmax=10,
      ymax=10,
      grid=major,
      scaled ticks=true,
      xlabel={$\eta^2$},
      mark options= {solid},
      ylabel={NMSE},
      ]
      \addplot[blue,error bars/.cd,y dir=both,y explicit, error bar style={mark size=2.5pt}] plot coordinates{
	      (1.000000e-05,8.426655e-07)+-(6.765713e-08,6.765713e-08)(1.778279e-05,1.277603e-06)+-(1.364289e-07,1.364289e-07)(3.162278e-05,2.053008e-06)+-(1.863787e-07,1.863787e-07)(5.623413e-05,3.418560e-06)+-(2.978738e-07,2.978738e-07)(1.000000e-04,5.802174e-06)+-(5.467665e-07,5.467665e-07)(1.778279e-04,1.011563e-05)+-(9.746692e-07,9.746692e-07)(3.162278e-04,1.804316e-05)+-(1.684913e-06,1.684913e-06)(5.623413e-04,3.372900e-05)+-(3.292288e-06,3.292288e-06)(1.000000e-03,6.358507e-05)+-(6.030959e-06,6.030959e-06)(1.778279e-03,1.241904e-04)+-(1.243791e-05,1.243791e-05)(3.162278e-03,2.544853e-04)+-(2.424503e-05,2.424503e-05)(5.623413e-03,5.265377e-04)+-(5.259113e-05,5.259113e-05)(1.000000e-02,1.149216e-03)+-(1.105034e-04,1.105034e-04)(1.778279e-02,2.433270e-03)+-(2.391696e-04,2.391696e-04)(3.162278e-02,4.972888e-03)+-(5.406908e-04,5.406908e-04)(5.623413e-02,1.094779e-02)+-(1.073743e-03,1.073743e-03)(1.000000e-01,2.771694e-02)+-(2.834033e-03,2.834033e-03)(1.778279e-01,7.681251e-02)+-(7.950787e-03,7.950787e-03)(3.162278e-01,2.183283e-01)+-(2.566712e-02,2.566712e-02)(5.623413e-01,5.640759e-01)+-(6.286384e-02,6.286384e-02)(1.000000e+00,1.107670e+00)+-(1.217388e-01,1.217388e-01)(1.778279e+00,1.563899e+00)+-(1.661549e-01,1.661549e-01)(3.162278e+00,1.842183e+00)+-(1.899369e-01,1.899369e-01)(5.623413e+00,1.952124e+00)+-(2.140676e-01,2.140676e-01)(1.000000e+01,1.928939e+00)+-(2.199082e-01,2.199082e-01)
      };
      \addplot[black,smooth,line width=0.5pt] plot coordinates{
	      (1.000000e-05,8.916354e-07)(1.778279e-05,1.344753e-06)(3.162278e-05,2.097912e-06)(5.623413e-05,3.317228e-06)(1.000000e-04,5.384810e-06)(1.778279e-04,9.175850e-06)(3.162278e-04,1.654771e-05)(5.623413e-04,3.141687e-05)(1.000000e-03,6.138055e-05)(1.778279e-03,1.231137e-04)(3.162278e-03,2.545264e-04)(5.623413e-03,5.403011e-04)(1.000000e-02,1.148029e-03)(1.778279e-02,2.423934e-03)(3.162278e-02,5.186653e-03)(5.623413e-02,1.166374e-02)(1.000000e-01,2.843311e-02)(1.778279e-01,7.608464e-02)(3.162278e-01,2.104761e-01)(5.623413e-01,5.314572e-01)(1.000000e+00,1.053514e+00)(1.778279e+00,1.534015e+00)(3.162278e+00,1.793219e+00)(5.623413e+00,1.894503e+00)(1.000000e+01,1.928961e+00)
      };
      \addplot[red,smooth,line width=0.5pt] plot coordinates{
      (1.000000e-05,1.881951e-03)(1.778279e-05,1.236034e-03)(3.162278e-05,7.082614e-04)(5.623413e-05,3.314769e-04)(1.000000e-04,1.584329e-04)(1.778279e-04,1.134160e-04)(3.162278e-04,8.872102e-05)(5.623413e-04,7.592914e-05)(1.000000e-03,9.356249e-05)(1.778279e-03,1.496411e-04)(3.162278e-03,2.691923e-04)(5.623413e-03,5.355922e-04)(1.000000e-02,1.142441e-03)(1.778279e-02,2.543727e-03)(3.162278e-02,5.835789e-03)(5.623413e-02,1.380695e-02)(1.000000e-01,3.382477e-02)(1.778279e-01,8.609830e-02)(3.162278e-01,2.252267e-01)(5.623413e-01,5.503717e-01)(1.000000e+00,1.071875e+00)(1.778279e+00,1.545657e+00)(3.162278e+00,1.798975e+00)(5.623413e+00,1.897548e+00)(1.000000e+01,1.931032e+00)
      };
      \addplot[blue,densely dashed,error bars/.cd,y dir=both,y explicit, error bar style={mark size=2.5pt}] plot coordinates{
	      (1.000000e-05,2.048376e-07)+-(1.813605e-08,1.813605e-08)(1.778279e-05,3.124490e-07)+-(2.560596e-08,2.560596e-08)(3.162278e-05,5.064581e-07)+-(4.696776e-08,4.696776e-08)(5.623413e-05,8.475481e-07)+-(7.382821e-08,7.382821e-08)(1.000000e-04,1.470097e-06)+-(1.275085e-07,1.275085e-07)(1.778279e-04,2.559966e-06)+-(2.075058e-07,2.075058e-07)(3.162278e-04,4.647501e-06)+-(3.903707e-07,3.903707e-07)(5.623413e-04,8.429807e-06)+-(7.712185e-07,7.712185e-07)(1.000000e-03,1.578552e-05)+-(1.597893e-06,1.597893e-06)(1.778279e-03,3.119271e-05)+-(2.974094e-06,2.974094e-06)(3.162278e-03,6.241851e-05)+-(6.083931e-06,6.083931e-06)(5.623413e-03,1.342819e-04)+-(1.466406e-05,1.466406e-05)(1.000000e-02,2.891896e-04)+-(2.940906e-05,2.940906e-05)(1.778279e-02,6.179837e-04)+-(6.045710e-05,6.045710e-05)(3.162278e-02,1.296750e-03)+-(1.235853e-04,1.235853e-04)(5.623413e-02,2.846013e-03)+-(3.253396e-04,3.253396e-04)(1.000000e-01,7.130681e-03)+-(8.339788e-04,8.339788e-04)(1.778279e-01,2.001278e-02)+-(2.137630e-03,2.137630e-03)(3.162278e-01,5.682788e-02)+-(6.452968e-03,6.452968e-03)(5.623413e-01,1.413413e-01)+-(1.536008e-02,1.536008e-02)(1.000000e+00,2.719667e-01)+-(2.983354e-02,2.983354e-02)(1.778279e+00,3.766999e-01)+-(4.591094e-02,4.591094e-02)(3.162278e+00,4.302408e-01)+-(4.900288e-02,4.900288e-02)(5.623413e+00,4.577639e-01)+-(5.493275e-02,5.493275e-02)(1.000000e+01,4.510498e-01)+-(5.781343e-02,5.781343e-02)
      };
      \addplot[black,densely dashed,smooth,line width=0.5pt] plot coordinates{
	      (1.000000e-05,2.116883e-07)(1.778279e-05,3.229594e-07)(3.162278e-05,5.101431e-07)(5.623413e-05,8.184185e-07)(1.000000e-04,1.348214e-06)(1.778279e-04,2.319519e-06)(3.162278e-04,4.197527e-06)(5.623413e-04,7.955856e-06)(1.000000e-03,1.556343e-05)(1.778279e-03,3.143715e-05)(3.162278e-03,6.525657e-05)(5.623413e-03,1.380009e-04)(1.000000e-02,2.916412e-04)(1.778279e-02,6.181524e-04)(3.162278e-02,1.338240e-03)(5.623413e-02,3.046947e-03)(1.000000e-01,7.476518e-03)(1.778279e-01,2.001949e-02)(3.162278e-01,5.519504e-02)(5.623413e-01,1.381847e-01)(1.000000e+00,2.702607e-01)(1.778279e+00,3.888540e-01)(3.162278e+00,4.517170e-01)(5.623413e+00,4.760773e-01)(1.000000e+01,4.843394e-01)
      };
      \addplot[red,densely dashed,smooth,line width=0.5pt] plot coordinates{
	      (1.000000e-05,4.884289e-08)(1.778279e-05,9.184731e-08)(3.162278e-05,1.764752e-07)(5.623413e-05,3.440188e-07)(1.000000e-04,6.776916e-07)(1.778279e-04,1.355303e-06)(3.162278e-04,2.745483e-06)(5.623413e-04,5.617490e-06)(1.000000e-03,1.167057e-05)(1.778279e-03,2.483388e-05)(3.162278e-03,5.458761e-05)(5.623413e-03,1.234440e-04)(1.000000e-02,2.808686e-04)(1.778279e-02,6.409030e-04)(3.162278e-02,1.483875e-03)(5.623413e-02,3.547751e-03)(1.000000e-01,8.810787e-03)(1.778279e-01,2.261085e-02)(3.162278e-01,5.914895e-02)(5.623413e-01,1.438820e-01)(1.000000e+00,2.782927e-01)(1.778279e+00,3.989113e-01)(3.162278e+00,4.628400e-01)(5.623413e+00,4.876126e-01)(1.000000e+01,4.960145e-01)
      };
      \legend{ {Monte Carlo}, {Th. (fixed $W$)}, {Th. (limit)} }
    \draw[->,thick] (axis cs:.1,1) -- (axis cs:.2,.15) node [above,pos=0,font=\footnotesize] {Test};
    \draw[->,thick] (axis cs:1.1,.025) -- (axis cs:.5,.1) node [right,pos=0,font=\footnotesize] {Train};
    \end{loglogaxis}
  \end{tikzpicture}
  &
  \begin{tikzpicture}[font=\footnotesize]
    \renewcommand{\axisdefaulttryminticks}{4} 
    \tikzstyle{every major grid}+=[style=densely dashed]       
    \tikzstyle{every axis y label}+=[yshift=-10pt] 
    \tikzstyle{every axis x label}+=[yshift=5pt]
    \tikzstyle{every axis legend}+=[cells={anchor=west},fill=white,
        at={(0.98,0.02)}, anchor=south east, font=\scriptsize ]
    \begin{loglogaxis}[
      title={$n=400$, $T=\hat{T}=800$},
      width=.45\linewidth,
      xmin=1e-4,
      ymin=1e-6,
      xmax=10,
      ymax=10,
      grid=major,
      scaled ticks=true,
      xlabel={$\eta^2$},
      mark options= {solid},
      yticklabels = {},
      ]
      \addplot[blue,error bars/.cd,y dir=both,y explicit, error bar style={mark size=2.5pt}] plot coordinates{
(1.000000e-05,8.422052e-07)+-(6.524848e-08,6.524848e-08)(1.778279e-05,1.310984e-06)+-(8.186030e-08,8.186030e-08)(3.162278e-05,2.014595e-06)+-(1.260683e-07,1.260683e-07)(5.623413e-05,3.148998e-06)+-(1.198106e-07,1.198106e-07)(1.000000e-04,5.661969e-06)+-(4.208252e-07,4.208252e-07)(1.778279e-04,9.772401e-06)+-(7.015599e-07,7.015599e-07)(3.162278e-04,1.710596e-05)+-(1.005179e-06,1.005179e-06)(5.623413e-04,3.147299e-05)+-(2.737348e-06,2.737348e-06)(1.000000e-03,6.003702e-05)+-(4.200177e-06,4.200177e-06)(1.778279e-03,1.197236e-04)+-(8.217470e-06,8.217470e-06)(3.162278e-03,2.522378e-04)+-(1.057917e-05,1.057917e-05)(5.623413e-03,5.428120e-04)+-(5.479171e-05,5.479171e-05)(1.000000e-02,1.173869e-03)+-(1.007669e-04,1.007669e-04)(1.778279e-02,2.443529e-03)+-(1.595853e-04,1.595853e-04)(3.162278e-02,4.864905e-03)+-(2.300957e-04,2.300957e-04)(5.623413e-02,1.111841e-02)+-(8.348499e-04,8.348499e-04)(1.000000e-01,2.762239e-02)+-(3.254012e-03,3.254012e-03)(1.778279e-01,7.948377e-02)+-(6.298685e-03,6.298685e-03)(3.162278e-01,2.209199e-01)+-(1.331910e-02,1.331910e-02)(5.623413e-01,5.419610e-01)+-(4.812230e-02,4.812230e-02)(1.000000e+00,1.084289e+00)+-(7.121442e-02,7.121442e-02)(1.778279e+00,1.594187e+00)+-(1.094259e-01,1.094259e-01)(3.162278e+00,1.868634e+00)+-(1.846411e-01,1.846411e-01)(5.623413e+00,1.877041e+00)+-(1.152448e-01,1.152448e-01)(1.000000e+01,1.916547e+00)+-(9.746331e-02,9.746331e-02)
      };
      \addplot[black,smooth,line width=0.5pt] plot coordinates{
	      (1.000000e-05,8.611346e-07)(1.778279e-05,1.293537e-06)(3.162278e-05,2.018885e-06)(5.623413e-05,3.266502e-06)(1.000000e-04,5.457413e-06)(1.778279e-04,9.431432e-06)(3.162278e-04,1.690283e-05)(5.623413e-04,3.129976e-05)(1.000000e-03,6.007859e-05)(1.778279e-03,1.210725e-04)(3.162278e-03,2.507394e-04)(5.623413e-03,5.283186e-04)(1.000000e-02,1.124072e-03)(1.778279e-02,2.374621e-03)(3.162278e-02,4.999127e-03)(5.623413e-02,1.102354e-02)(1.000000e-01,2.740014e-02)(1.778279e-01,7.594613e-02)(3.162278e-01,2.142175e-01)(5.623413e-01,5.441262e-01)(1.000000e+00,1.078449e+00)(1.778279e+00,1.568084e+00)(3.162278e+00,1.831500e+00)(5.623413e+00,1.934311e+00)(1.000000e+01,1.969275e+00)
      };
      \addplot[red,smooth,line width=0.5pt] plot coordinates{
	      (1.000000e-05,2.158324e-06)(1.778279e-05,2.222249e-06)(3.162278e-05,2.699693e-06)(5.623413e-05,3.821587e-06)(1.000000e-04,5.967865e-06)(1.778279e-04,9.952312e-06)(3.162278e-04,1.756008e-05)(5.623413e-04,3.244042e-05)(1.000000e-03,6.228394e-05)(1.778279e-03,1.235916e-04)(3.162278e-03,2.526804e-04)(5.623413e-03,5.307038e-04)(1.000000e-02,1.144492e-03)(1.778279e-02,2.541588e-03)(3.162278e-02,5.780350e-03)(5.623413e-02,1.348636e-02)(1.000000e-01,3.264252e-02)(1.778279e-01,8.305261e-02)(3.162278e-01,2.192235e-01)(5.623413e-01,5.421906e-01)(1.000000e+00,1.071025e+00)(1.778279e+00,1.563089e+00)(3.162278e+00,1.830874e+00)(5.623413e+00,1.935982e+00)(1.000000e+01,1.971802e+00)
      };
      \addplot[blue,densely dashed,error bars/.cd,y dir=both,y explicit, error bar style={mark size=2.5pt}] plot coordinates{
	      (1.000000e-05,2.225730e-07)+-(1.414404e-08,1.414404e-08)(1.778279e-05,3.261020e-07)+-(2.611786e-08,2.611786e-08)(3.162278e-05,5.041492e-07)+-(3.895086e-08,3.895086e-08)(5.623413e-05,8.295161e-07)+-(5.765633e-08,5.765633e-08)(1.000000e-04,1.357699e-06)+-(6.862783e-08,6.862783e-08)(1.778279e-04,2.518469e-06)+-(1.346322e-07,1.346322e-07)(3.162278e-04,4.216069e-06)+-(2.905335e-07,2.905335e-07)(5.623413e-04,8.167107e-06)+-(4.140376e-07,4.140376e-07)(1.000000e-03,1.506296e-05)+-(1.228451e-06,1.228451e-06)(1.778279e-03,3.149063e-05)+-(2.572049e-06,2.572049e-06)(3.162278e-03,6.365640e-05)+-(3.934095e-06,3.934095e-06)(5.623413e-03,1.367864e-04)+-(7.072534e-06,7.072534e-06)(1.000000e-02,2.927333e-04)+-(2.394582e-05,2.394582e-05)(1.778279e-02,5.895630e-04)+-(3.459073e-05,3.459073e-05)(3.162278e-02,1.268788e-03)+-(1.123942e-04,1.123942e-04)(5.623413e-02,2.750956e-03)+-(2.422120e-04,2.422120e-04)(1.000000e-01,6.891512e-03)+-(4.003337e-04,4.003337e-04)(1.778279e-01,1.829411e-02)+-(1.391216e-03,1.391216e-03)(3.162278e-01,5.373809e-02)+-(4.018550e-03,4.018550e-03)(5.623413e-01,1.377743e-01)+-(1.083300e-02,1.083300e-02)(1.000000e+00,2.746502e-01)+-(2.914742e-02,2.914742e-02)(1.778279e+00,4.041681e-01)+-(3.083648e-02,3.083648e-02)(3.162278e+00,4.515468e-01)+-(2.759966e-02,2.759966e-02)(5.623413e+00,4.914574e-01)+-(2.857694e-02,2.857694e-02)(1.000000e+01,5.044026e-01)+-(3.604905e-02,3.604905e-02)
      };
      \addplot[black,densely dashed,smooth,line width=0.5pt] plot coordinates{
	      (1.000000e-05,2.101540e-07)(1.778279e-05,3.167775e-07)(3.162278e-05,4.965713e-07)(5.623413e-05,8.074174e-07)(1.000000e-04,1.357553e-06)(1.778279e-04,2.359798e-06)(3.162278e-04,4.245073e-06)(5.623413e-04,7.869750e-06)(1.000000e-03,1.505200e-05)(1.778279e-03,3.011060e-05)(3.162278e-03,6.223480e-05)(5.623413e-03,1.320493e-04)(1.000000e-02,2.833325e-04)(1.778279e-02,6.001873e-04)(3.162278e-02,1.259712e-03)(5.623413e-02,2.773427e-03)(1.000000e-01,6.913970e-03)(1.778279e-01,1.921972e-02)(3.162278e-01,5.428689e-02)(5.623413e-01,1.379698e-01)(1.000000e+00,2.735331e-01)(1.778279e+00,3.977900e-01)(3.162278e+00,4.646506e-01)(5.623413e+00,4.907487e-01)(1.000000e+01,4.996243e-01)
      };
      \addplot[red,densely dashed,smooth,line width=0.5pt] plot coordinates{
	      (1.000000e-05,1.345435e-07)(1.778279e-05,2.212593e-07)(3.162278e-05,3.737093e-07)(5.623413e-05,6.476740e-07)(1.000000e-04,1.149961e-06)(1.778279e-04,2.093569e-06)(3.162278e-04,3.907282e-06)(5.623413e-04,7.482073e-06)(1.000000e-03,1.475163e-05)(1.778279e-03,2.986478e-05)(3.162278e-03,6.181445e-05)(5.623413e-03,1.309140e-04)(1.000000e-02,2.847311e-04)(1.778279e-02,6.364142e-04)(3.162278e-02,1.449149e-03)(5.623413e-02,3.370791e-03)(1.000000e-01,8.145209e-03)(1.778279e-01,2.077736e-02)(3.162278e-01,5.503957e-02)(5.623413e-01,1.363639e-01)(1.000000e+00,2.694294e-01)(1.778279e+00,3.931052e-01)(3.162278e+00,4.603533e-01)(5.623413e+00,4.867376e-01)(1.000000e+01,4.957279e-01)
      };
      \legend{ {Monte Carlo}, {Th. (fixed $W$)}, {Th. (limit)} }
    \draw[->,thick] (axis cs:.1,1) -- (axis cs:.2,.15) node [above,pos=0,font=\footnotesize] {Test};
    \draw[->,thick] (axis cs:1.1,.025) -- (axis cs:.5,.1) node [right,pos=0,font=\footnotesize] {Train};
    \end{loglogaxis}
  \end{tikzpicture}
  \end{tabular}
  \caption{Training and testing (normalized) MSE for the Mackey Glass one-step ahead task, $W$ fixed and defined as in Figure~\ref{fig:multimemory}, $n=200$, $T=\hat{T}=400$ (left) and $n=400$, $T=\hat{T}=800$ (right). Comparison between Monte Carlo simulations (Monte Carlo), deterministic approximation assuming $W$ fixed (Th. (fixed $W$)) as per Corollaries~\ref{cor:MSEtrain} and \ref{cor:MSEtest}, and assuming $W$ random in the large $n$ limit (Th. (limit)) as per Corollary~\ref{cor:unitarily_invariant}. Error bars indicate one standard deviation of the Monte Carlo simulations.}
  \label{fig:mackeyglass}
\end{figure}

As a practical example, we provide in Figure~\ref{fig:mackeyglass} Monte Carlo simulations versus theory curves of the training and testing performances of networks of $n=200$ and $n=400$ nodes, for training and testing times $T=\hat{T}=2n$, on the Mackey Glass one-step ahead anticipation task \cite{GLA79}. The network is chosen to be the multi-memory model introduced in Remark~\ref{rem:multimemory} and following the description of Figure~\ref{fig:multimemory}. The NMSE is defined here as the ratio between the MSE and the output vector squared norm $\|r\|^2/T$ or $\|\hat{r}\|^2/\hat{T}$. Simulations are run for a single $W$ but different noise realizations and comparison is made against theory for either this $W$ or its approximated asymptotic limit. Observe the extremely accurate match between theory and practice, with increasing precision as $n,T$ grow large.

\subsubsection{Case $c>1$}

The case $c>1$ is slightly more involved as it does not lend itself to a purely explicit expression. Precisely, following the same steps as for $c<1$, we find that in the large $n$ limit
\begin{align*}
	\mathcal R &\leftrightarrow \alpha I_T \\
	\tilde{\mathcal R} &\leftrightarrow \frac1{\alpha} S_0 \\
	\mathcal G^{[B]} &\leftrightarrow c \alpha^2 \frac{\frac1n\tr S_0\left( \alpha I_n + S_0 \right)^{-1}B\left( \alpha I_n + S_0 \right)^{-1}}{1 - c \frac1n\tr S_0^2\left( \alpha I_n + S_0 \right)^{-2} } I_T \\
	\tilde{\mathcal G}^{[B]} &\leftrightarrow c \frac{\frac1n\tr S_0\left( \alpha I_n + S_0 \right)^{-1}B\left( \alpha I_n + S_0 \right)^{-1}}{1 - c \frac1n\tr S_0^2\left( \alpha I_n + S_0 \right)^{-2} } S_0
\end{align*}
where $\alpha>0$ is the unique solution to the equation
\begin{align*}
	1 &= c \frac1n\tr S_0 \left( \alpha I_n + S_0 \right)^{-1}.
\end{align*}

With these notations, we have the following counterpart to Corollary~\ref{cor:unitarily_invariant}.
\begin{corollary}[Orthogonally invariant case, $c>1$]
	\label{cor:unitarily_invariant_c>1}
	Let $W$ be random and left and right independently orthogonally invariant and let $\alpha>0$ be the unique solution to $1=c \frac1n\tr S_0 \left( \alpha I_n + S_0 \right)^{-1}$. Then, under Assumptions~\ref{ass:spectral_radius}--\ref{ass:rmthat} and with $c>1$, the following holds
	\begin{align*}
		\hat{E}_{\eta}(u,r;\hat{u},\hat{r}) &\leftrightarrow \left\| \eta^{-2}\hat{U}^\trans \hat{D}U \left( I_T + \eta^{-2} U^\trans DU \right)^{-1}\frac{r}{\sqrt{T}} - \frac1{\sqrt{\hat T}} \hat{r} \right\|^2 - \frac1Tr^\trans \left( I_T + \eta^{-2}U^\trans DU \right)^{-1}r \nonumber \\ 
		&+ \frac{\frac1Tr^\trans \left( I_T + \eta^{-2} U^\trans D U \right)^{-1} \left[ I_T + \eta^{-2} U^\trans D_2 U \right] \left( I_T + \eta^{-2} U^\trans D U \right)^{-1}  r}{1-c\frac1n\tr S_0^2\left( \alpha I_T + S_0 \right)^{-2} } 
	\end{align*}
	where $D \equiv \left\{ m^\trans (W^i)^\trans (\alpha I_n + S_0)^{-1} W^j m \right\}_{i,j=0}^{T-1}$, $\hat{D} \equiv \left\{ m^\trans (W^i)^\trans ( \alpha I_n + S_0)^{-1} W^j m \right\}_{i,j=0}^{\hat{T}-1,T-1}$, and $D_2\equiv \left\{ m^\trans (W^i)^\trans (\alpha I_n + S_0)^{-1}S_0(\alpha I_n + S_0)^{-1} W^j m \right\}_{i,j=0}^{T-1}$.
\end{corollary}
Of course here $E_{\eta}(u,r)=0$.

\begin{remark}[Haar $W$, random $m$ for $c>1$]
	\label{rem:Haar_c>1}
Although seemingly less tractable, for $W$ following a Haar model, Corollary~\ref{cor:unitarily_invariant_c>1} takes a much simpler form. Indeed, for $W$ and $m$ as defined in Remark~\ref{rem:Haar}, we find that $\alpha=(c-1)(1-\sigma^2)^{-1}$ and $S_0=(1-\sigma^2)^{-1}I_n$ which then leads to
\begin{align*}
	\hat{E}_{\eta}(u,r;\hat{u},\hat{r}) &\leftrightarrow \left\| (c\eta^2)^{-1} \hat{U}^\trans \hat{D}U \left( I_T + (c\eta^2)^{-1} U^\trans DU \right)^{-1}\frac{r}{\sqrt{T}}  - \frac1{\sqrt{\hat T}} \hat{r}  \right\|^2 \nonumber \\
	&+ \frac1{c-1}\frac1Tr^\trans \left( I_T + (c\eta^2)^{-1} U^\trans DU \right)^{-1} r
\end{align*}
where $D$ is diagonal with $D_{ii}\equiv (1-\sigma^2)\sigma^{2(i-1)}$. 
\end{remark}

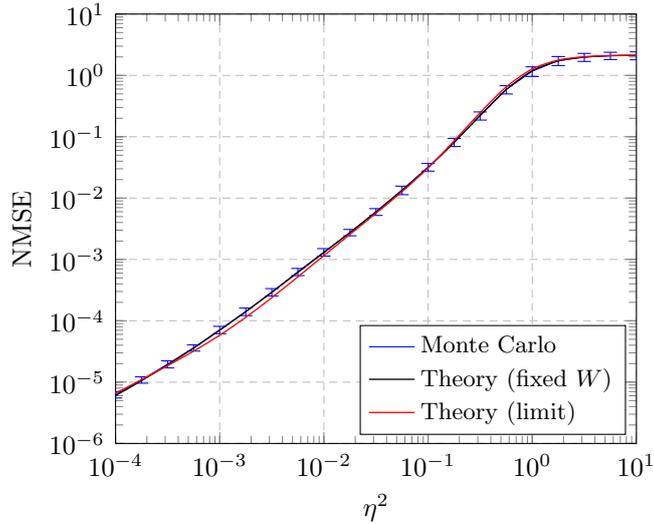
\begin{figure}[h!]
  \centering
  \begin{tikzpicture}[font=\footnotesize]
    \renewcommand{\axisdefaulttryminticks}{4} 
    \tikzstyle{every major grid}+=[style=densely dashed]       
    \tikzstyle{every axis y label}+=[yshift=-10pt] 
    \tikzstyle{every axis x label}+=[yshift=5pt]
    \tikzstyle{every axis legend}+=[cells={anchor=west},fill=white,
        at={(0.98,0.02)}, anchor=south east, font=\scriptsize ]
    \begin{loglogaxis}[
      xmin=1e-4,
      ymin=1e-6,
      xmax=10,
      ymax=10,
      grid=major,
      scaled ticks=true,
      xlabel={$\eta^2$},
      mark options= {solid},
      ylabel={NMSE},
      ]
      \addplot[blue,error bars/.cd,y dir=both,y explicit, error bar style={mark size=2.5pt}] plot coordinates{
	      (1.000000e-05,1.022313e-06)+-(1.193708e-07,1.193708e-07)(1.778279e-05,1.542200e-06)+-(1.814080e-07,1.814080e-07)(3.162278e-05,2.341479e-06)+-(2.706662e-07,2.706662e-07)(5.623413e-05,3.696993e-06)+-(3.979865e-07,3.979865e-07)(1.000000e-04,6.256436e-06)+-(7.427454e-07,7.427454e-07)(1.778279e-04,1.089714e-05)+-(1.308622e-06,1.308622e-06)(3.162278e-04,1.965093e-05)+-(2.609600e-06,2.609600e-06)(5.623413e-04,3.630379e-05)+-(4.299182e-06,4.299182e-06)(1.000000e-03,7.138167e-05)+-(9.866558e-06,9.866558e-06)(1.778279e-03,1.413494e-04)+-(2.008625e-05,2.008625e-05)(3.162278e-03,2.943402e-04)+-(4.038488e-05,4.038488e-05)(5.623413e-03,6.290129e-04)+-(8.435741e-05,8.435741e-05)(1.000000e-02,1.318243e-03)+-(1.840048e-04,1.840048e-04)(1.778279e-02,2.755762e-03)+-(3.294008e-04,3.294008e-04)(3.162278e-02,5.976069e-03)+-(7.601881e-04,7.601881e-04)(5.623413e-02,1.350437e-02)+-(2.146915e-03,2.146915e-03)(1.000000e-01,3.208806e-02)+-(4.574139e-03,4.574139e-03)(1.778279e-01,8.163532e-02)+-(1.235951e-02,1.235951e-02)(3.162278e-01,2.204436e-01)+-(3.333996e-02,3.333996e-02)(5.623413e-01,5.889225e-01)+-(9.140408e-02,9.140408e-02)(1.000000e+00,1.170412e+00)+-(2.107111e-01,2.107111e-01)(1.778279e+00,1.737226e+00)+-(2.914852e-01,2.914852e-01)(3.162278e+00,1.985548e+00)+-(2.976866e-01,2.976866e-01)(5.623413e+00,2.098238e+00)+-(2.846659e-01,2.846659e-01)(1.000000e+01,2.108441e+00)+-(3.117946e-01,3.117946e-01)
      };
      \addplot[black,smooth,line width=0.5pt] plot coordinates{
	      (1.000000e-05,9.799879e-07)(1.778279e-05,1.453250e-06)(3.162278e-05,2.244968e-06)(5.623413e-05,3.599483e-06)(1.000000e-04,6.040589e-06)(1.778279e-04,1.057724e-05)(3.162278e-04,1.916255e-05)(5.623413e-04,3.600850e-05)(1.000000e-03,7.027545e-05)(1.778279e-03,1.409113e-04)(3.162278e-03,2.909125e-04)(5.623413e-03,6.156290e-04)(1.000000e-02,1.295870e-03)(1.778279e-02,2.735443e-03)(3.162278e-02,5.951360e-03)(5.623413e-02,1.336240e-02)(1.000000e-01,3.130739e-02)(1.778279e-01,8.049595e-02)(3.162278e-01,2.234641e-01)(5.623413e-01,5.864300e-01)(1.000000e+00,1.189511e+00)(1.778279e+00,1.719378e+00)(3.162278e+00,1.989083e+00)(5.623413e+00,2.091123e+00)(1.000000e+01,2.125404e+00)
      };
      \addplot[red,smooth,line width=0.5pt] plot coordinates{
	      (1.000000e-05,3.056760e-05)(1.778279e-05,1.373076e-05)(3.162278e-05,6.030860e-06)(5.623413e-05,4.781465e-06)(1.000000e-04,6.685307e-06)(1.778279e-04,1.098848e-05)(3.162278e-04,1.855778e-05)(5.623413e-04,3.204596e-05)(1.000000e-03,5.812687e-05)(1.778279e-03,1.138368e-04)(3.162278e-03,2.382591e-04)(5.623413e-03,5.202929e-04)(1.000000e-02,1.155667e-03)(1.778279e-02,2.556578e-03)(3.162278e-02,5.654798e-03)(5.623413e-02,1.272168e-02)(1.000000e-01,3.111843e-02)(1.778279e-01,8.561140e-02)(3.162278e-01,2.473391e-01)(5.623413e-01,6.492486e-01)(1.000000e+00,1.272330e+00)(1.778279e+00,1.776005e+00)(3.162278e+00,2.018645e+00)(5.623413e+00,2.108244e+00)(1.000000e+01,2.138084e+00)
      };
      \legend{ {Monte Carlo}, {Theory (fixed $W$)}, {Theory (limit)} }
    \end{loglogaxis}
  \end{tikzpicture}
  \caption{Testing (normalized) MSE for the Mackey Glass one-step ahead task, $W$ fixed and defined as in Figure~\ref{fig:multimemory}, $n=400$, $T=\hat{T}=200$. Error bars indicate one standard deviation of the Monte Carlo simulations.}
  \label{fig:mackeyglass_c>1}
\end{figure}

Aside from obtaining a shorter form expression for $D$ and $D_2$, the multi memory model of Remark~\ref{rem:multimemory} does not lead to an explicit formulation as in Remark~\ref{rem:Haar_c>1}, but it is nonetheless instructive to observe the performance achieved on the Mackey Glass model from Figure~\ref{fig:mackeyglass}, now in the setting where $c>1$. This is depicted here in Figure~\ref{fig:mackeyglass_c>1}.

\subsection{Normal $W$}
\label{sec:Hermitian}

We now turn to the case of normal matrices. Let then $W$ be normal (i.e., diagonalizable in orthogonal basis) and having an eigenvalue decomposition of the type $W=V\Lambda V^\trans$ with $V$ orthogonal and $\Lambda$ diagonal with largest absolute entry less than one. For simplicity, we shall further assume that, as $n\to\infty$, the normalized counting measure of the diagonal elements of $\Lambda$ ($n^{-1}\sum_i {\bm\delta}_{\Lambda_{ii}}$) converges in law to a probability measure $\mu$. We do not make any assumption here on $V$.

For instance, real Gaussian Wigner matrices $W$, that is with i.i.d.\@ zero mean variance $\frac14\sigma^2$ Gaussian entries on and above the diagonal, and symmetrized below the diagonal, is an example of such a matrix. In this case, $\mu$ corresponds (almost surely) to the well-known semi-circular distribution, with density $\mu(d\lambda)=2(\pi\sigma^2)^{-1}\sqrt{(\sigma^2-\lambda^2)^+}d\lambda$. Another example is when $\mu(d\lambda)=\frac12 [{\bm\delta}_\sigma+{\bm\delta}_{-\sigma}]d\lambda$, so that $W$ is the sum of two ($\sigma$-scaled) projection matrices on orthogonal subspaces. In particular here, $W^2=\sigma^2 I_n$, so that $W^{2k}=\sigma^{2k}I_n$ and $W^{2k+1}=\sigma^{2k}W$, for all $k\geq 0$.

Because of the symmetry property, it is no longer true that $\frac1n\tr W^i(W^j)^\trans=\frac1n\tr W^{i+j}$ vanished for $i\neq j$, and we then obtain more involved results. To keep this discussion short and since the results take here more involved forms, we shall only deal here with the case $c<1$ and focus on the training performance. In this case, solving Proposition~\ref{prop:fundeq} for $\mathcal R$ and $\tilde{\mathcal R}$, we have the following result. As $n\to\infty$, $\mathcal R$ has a limit (which for simplicity we keep calling $\mathcal R$) which is solution to
\begin{align}
	\label{eq:R_normal}
	\mathcal R_{ab} &= c \int \frac{t^{|a-b|} \mu(dt)}{\sum_{q=-\infty}^\infty \frac1T\tr (J^q(I_T+\mathcal R)^{-1}) t^{|q|}}
\end{align}
for all $a,b\in\{1,\ldots,T\}$.
Remember that $\mathcal R$ is Toeplitz with fast decaying values off the diagonal, so that \eqref{eq:R_normal} is computationally easy to solve. Similar conclusions can be drawn on the matrices $\mathcal G^{[B]}$ and $\tilde{\mathcal G}^{[B]}$, that however do not lead to simple expressions.

\begin{figure}
	\centering
	\begin{tabular}{cc}
\begin{tikzpicture}
	\begin{axis}[view={0}{-90},
			xlabel={$\mathcal R$ [i.i.d.]},
			xtick={.5,1.5,2.5,3.5,4.5,5.5,6.5,7.5,8.5},
			ytick={.5,1.5,2.5,3.5,4.5,5.5,6.5,7.5,8.5},
			xticklabels={$1$,$2$,$3$,$4$,$5$,$6$,$7$,$8$,$9$},
			yticklabels={$1$,$2$,$3$,$4$,$5$,$6$,$7$,$8$,$9$},
			yticklabel pos=right,
			width=.4\linewidth,
			colormap={whiteblack}{color(0cm)=(white);color(1cm)=(black);},
		]
		\addplot3[surf,shader=flat corner] file {noncheckerboard.dat};
	\end{axis}
\end{tikzpicture}
&
\begin{tikzpicture}
	\begin{axis}[view={0}{-90},
			xlabel={$\mathcal R$ [Wigner]},
			xtick={.5,1.5,2.5,3.5,4.5,5.5,6.5,7.5,8.5},
			ytick={.5,1.5,2.5,3.5,4.5,5.5,6.5,7.5,8.5},
			xticklabels={$1$,$2$,$3$,$4$,$5$,$6$,$7$,$8$,$9$},
			yticklabels={$1$,$2$,$3$,$4$,$5$,$6$,$7$,$8$,$9$},			
			width=.4\linewidth,
			colormap={whiteblack}{color(0cm)=(white);color(1cm)=(black);},
		]
		\addplot3[surf,shader=flat corner] file {checkerboard.dat};
	\end{axis}
\end{tikzpicture}
\end{tabular}
\caption{Upper $9\times 9$ part of $\mathcal R$ for $c=1/2$ and $\sigma=0.9$ for $W$ with i.i.d.\@ zero mean Gaussian entries (left) and $W$ Gaussian Wigner (right). Linear grayscale representation with black being $1$ and white being $0$.}
\label{fig:checkerboard}
\end{figure}
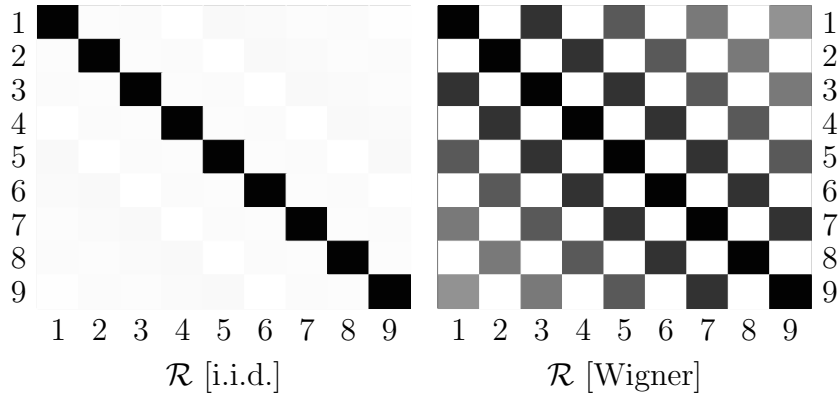

\begin{remark}[Symmetric $\mu$]
	An interesting scenario is when $\mu$ is symmetric, i.e., $\mu(-t)=\mu(t)$, which is the case of both aforementioned (Wigner and projection matrix) examples. From \eqref{eq:R_normal}, we find in this case that $[\mathcal R]_{ab}$ is zero if $a-b$ is odd and positive if $a-b$ is even. As such, $\mathcal R$, takes the form of a checkerboard matrix. Figure~\ref{fig:checkerboard} provides a representation of $\mathcal R$ in both normal and non-normal Gaussian $W$ cases.
\end{remark}

\begin{remark}[Projection $W$]
	\label{rem:proj_W}
	Let $W=V\Lambda V^\trans$ with the normalized counting measure of $\Lambda$ converging to $\mu(d\lambda)=\frac12 [{\bm\delta}_\sigma+{\bm\delta}_{-\sigma}]d\lambda$ and $c<1$. Then, $\mathcal R_{ab}\leftrightarrow \sigma^{|b-a|}r_0 {\bm\delta}_{|b-a|\in 2\mathbb{N}} $ and $\tilde{\mathcal R}\leftrightarrow -(1-\sigma^2)^{-1}cr_0^{-1}I_n$ where
	\begin{align*}
		r_0 &= c \left( \sum_{q=-\infty}^\infty \frac1T\tr J^q \left( I_T + \mathcal R \right)^{-1} \right)^{-1}.
	\end{align*}
	As a consequence, letting $m$ be random, we find that
	\begin{align*}
		E_\eta(u,r) &\leftrightarrow \frac1Tr^\trans \left( I_T + r_0 \left\{ \sigma^{|j-i|}{\bm\delta}_{|j-i|\in 2\mathbb{N}} \right\}_{i,j=0}^{T-1} + \frac{r_0(1-\alpha^2)}{\eta^2c} U^\trans \left\{ \sigma^{j+i}{\bm\delta}_{|j-i|\in 2\mathbb{N}} \right\}_{i,j=0}^{T-1} U \right)^{-1}r.
	\end{align*}
	Note in particular that the matrix $\{\sigma^{j+i}{\bm\delta}_{|j-i|\in 2\mathbb{N}}\}_{i,j=0}^{T-1}$ can be decomposed as the sum of two matrices: (i) the rank-one matrix $vv^\trans$ with $v=(1,0,\alpha^2,0,\alpha^4,\ldots)^{\trans}$ and the diagonal matrix $\diag(0,\alpha^2,0,\alpha^4,\ldots)$. Recalling that rank-one matrices in this position do not allow for efficient training (see the final discussions in Section~\ref{sec:non-Hermitian}, $c<1$ case), only the diagonal component $\diag(0,\alpha^2,0,\alpha^4,\ldots)$ really matters here. This diagonal misses half its entries and thus intuitively does not allow for efficient retrieval of odd past steps. This remark generally prefigures a weaker performance of normal matrices with symmetric spectrum than their non-normal counterparts.
\end{remark}

\begin{figure}[h!]
  \centering
  \begin{tabular}{cc}
  \begin{tikzpicture}[font=\footnotesize]
    \renewcommand{\axisdefaulttryminticks}{4} 
    \tikzstyle{every major grid}+=[style=densely dashed]       
    \tikzstyle{every axis y label}+=[yshift=-10pt] 
    \tikzstyle{every axis x label}+=[yshift=5pt]
    \tikzstyle{every axis legend}+=[cells={anchor=west},fill=white,
        at={(0.98,0.02)}, anchor=south east, font=\scriptsize ]
    \begin{loglogaxis}[
      width=0.45\linewidth,	    
      xmin=1e-5,
      ymin=1e-10,
      xmax=10,
      ymax=10,
      grid=major,
      scaled ticks=true,
      xlabel={$\eta^2$},
      mark options= {solid},
      ylabel={NMSE},
      ]
      \addplot[blue,densely dashed,smooth,line width=0.5pt] plot coordinates{
	      (1.000000e-05,3.279204e-08)(1.778279e-05,8.932814e-08)(3.162278e-05,2.391611e-07)(5.623413e-05,6.391022e-07)(1.000000e-04,1.693857e-06)(1.778279e-04,4.464806e-06)(3.162278e-04,1.169784e-05)(5.623413e-04,3.004562e-05)(1.000000e-03,7.724589e-05)(1.778279e-03,1.972436e-04)(3.162278e-03,4.955575e-04)(5.623413e-03,1.226097e-03)(1.000000e-02,2.985582e-03)(1.778279e-02,7.258118e-03)(3.162278e-02,1.690769e-02)(5.623413e-02,3.693155e-02)(1.000000e-01,7.507757e-02)(1.778279e-01,1.420801e-01)(3.162278e-01,2.505502e-01)(5.623413e-01,3.726522e-01)(1.000000e+00,4.529770e-01)(1.778279e+00,4.879866e-01)(3.162278e+00,5.004272e-01)(5.623413e+00,5.045168e-01)(1.000000e+01,5.058263e-01)
      };
      \addplot[red,densely dashed,smooth,line width=0.5pt] plot coordinates{
(1.000000e-05,2.326673e-10)(1.778279e-05,7.357585e-10)(3.162278e-05,2.326673e-09)(5.623413e-05,7.357585e-09)(1.000000e-04,2.326673e-08)(1.778279e-04,7.357584e-08)(3.162278e-04,2.326672e-07)(5.623413e-04,7.357573e-07)(1.000000e-03,2.326661e-06)(1.778279e-03,7.357467e-06)(3.162278e-03,2.326556e-05)(5.623413e-03,7.356428e-05)(1.000000e-02,2.325528e-04)(1.778279e-02,7.346267e-04)(3.162278e-02,2.315515e-03)(5.623413e-02,7.248204e-03)(1.000000e-01,2.221518e-02)(1.778279e-01,6.405365e-02)(3.162278e-01,1.585275e-01)(5.623413e-01,2.973303e-01)(1.000000e+00,4.113013e-01)(1.778279e+00,4.680707e-01)(3.162278e+00,4.894408e-01)(5.623413e+00,4.966118e-01)(1.000000e+01,4.989236e-01)
      };
      \addplot[blue,densely dashed,smooth,line width=0.5pt] plot coordinates{
	      (1.000000e-05,1.552887e-06)(1.778279e-05,3.670944e-06)(3.162278e-05,8.740572e-06)(5.623413e-05,2.080029e-05)(1.000000e-04,4.957519e-05)(1.778279e-04,1.194960e-04)(3.162278e-04,2.853558e-04)(5.623413e-04,6.598024e-04)(1.000000e-03,1.486719e-03)(1.778279e-03,3.296619e-03)(3.162278e-03,7.029794e-03)(5.623413e-03,1.439980e-02)(1.000000e-02,2.811381e-02)(1.778279e-02,5.137816e-02)(3.162278e-02,9.042274e-02)(5.623413e-02,1.511208e-01)(1.000000e-01,2.344142e-01)(1.778279e-01,3.343156e-01)(3.162278e-01,4.227321e-01)(5.623413e-01,4.737579e-01)(1.000000e+00,4.951002e-01)(1.778279e+00,5.026943e-01)(3.162278e+00,5.052148e-01)(5.623413e+00,5.060266e-01)(1.000000e+01,5.062850e-01)
      };
      \addplot[red,densely dashed,smooth,line width=0.5pt] plot coordinates{
	      (1.000000e-05,4.645488e-10)(1.778279e-05,1.469032e-09)(3.162278e-05,4.645488e-09)(5.623413e-05,1.469032e-08)(1.000000e-04,4.645488e-08)(1.778279e-04,1.469032e-07)(3.162278e-04,4.645483e-07)(5.623413e-04,1.469027e-06)(1.000000e-03,4.645439e-06)(1.778279e-03,1.468984e-05)(3.162278e-03,4.645010e-05)(5.623413e-03,1.468561e-04)(1.000000e-02,4.640832e-04)(1.778279e-02,1.464433e-03)(3.162278e-02,4.600286e-03)(5.623413e-02,1.425307e-02)(1.000000e-01,4.239472e-02)(1.778279e-01,1.130325e-01)(3.162278e-01,2.394062e-01)(5.623413e-01,3.710964e-01)(1.000000e+00,4.500290e-01)(1.778279e+00,4.829142e-01)(3.162278e+00,4.944489e-01)(5.623413e+00,4.982290e-01)(1.000000e+01,4.994384e-01)
      };
      \addplot[blue,densely dashed,smooth,line width=0.5pt] plot coordinates{
	      (1.000000e-05,3.268859e-05)(1.778279e-05,7.220227e-05)(3.162278e-05,1.520748e-04)(5.623413e-05,3.155534e-04)(1.000000e-04,6.436541e-04)(1.778279e-04,1.300832e-03)(3.162278e-04,2.705439e-03)(5.623413e-04,5.489474e-03)(1.000000e-03,1.064699e-02)(1.778279e-03,1.974771e-02)(3.162278e-03,3.507790e-02)(5.623413e-03,6.084674e-02)(1.000000e-02,1.000723e-01)(1.778279e-02,1.563638e-01)(3.162278e-02,2.310505e-01)(5.623413e-02,3.135775e-01)(1.000000e-01,3.928596e-01)(1.778279e-01,4.514035e-01)(3.162278e-01,4.826160e-01)(5.623413e-01,4.968335e-01)(1.000000e+00,5.029458e-01)(1.778279e+00,5.053512e-01)(3.162278e+00,5.062116e-01)(5.623413e+00,5.064987e-01)(1.000000e+01,5.065912e-01)
      };
      \addplot[red,densely dashed,smooth,line width=0.5pt] plot coordinates{
	      (1.000000e-05,8.174081e-10)(1.778279e-05,2.584871e-09)(3.162278e-05,8.174081e-09)(5.623413e-05,2.584871e-08)(1.000000e-04,8.174080e-08)(1.778279e-04,2.584870e-07)(3.162278e-04,8.174066e-07)(5.623413e-04,2.584856e-06)(1.000000e-03,8.173928e-06)(1.778279e-03,2.584721e-05)(3.162278e-03,8.172606e-05)(5.623413e-03,2.583419e-04)(1.000000e-02,8.159745e-04)(1.778279e-02,2.570759e-03)(3.162278e-02,8.036365e-03)(5.623413e-02,2.453551e-02)(1.000000e-01,7.001408e-02)(1.778279e-01,1.694699e-01)(3.162278e-01,3.083761e-01)(5.623413e-01,4.171940e-01)(1.000000e+00,4.702353e-01)(1.778279e+00,4.901445e-01)(3.162278e+00,4.968352e-01)(5.623413e+00,4.989942e-01)(1.000000e+01,4.996814e-01)
      };
      \addplot[blue,densely dashed,smooth,line width=0.5pt] plot coordinates{
	      (1.000000e-05,2.938582e-04)(1.778279e-05,6.052923e-04)(3.162278e-05,1.238962e-03)(5.623413e-05,2.456351e-03)(1.000000e-04,4.805769e-03)(1.778279e-04,9.195568e-03)(3.162278e-04,1.690793e-02)(5.623413e-04,3.069549e-02)(1.000000e-03,5.228897e-02)(1.778279e-03,8.378885e-02)(3.162278e-03,1.306401e-01)(5.623413e-03,1.900895e-01)(1.000000e-02,2.573212e-01)(1.778279e-02,3.220412e-01)(3.162278e-02,3.753501e-01)(5.623413e-02,4.189391e-01)(1.000000e-01,4.504991e-01)(1.778279e-01,4.731314e-01)(3.162278e-01,4.894385e-01)(5.623413e-01,4.986936e-01)(1.000000e+00,5.031747e-01)(1.778279e+00,5.052746e-01)(3.162278e+00,5.061435e-01)(5.623413e+00,5.064540e-01)(1.000000e+01,5.065567e-01)
      };
      \addplot[red,densely dashed,smooth,line width=0.5pt] plot coordinates{
	      (1.000000e-05,1.143654e-09)(1.778279e-05,3.616551e-09)(3.162278e-05,1.143654e-08)(5.623413e-05,3.616551e-08)(1.000000e-04,1.143654e-07)(1.778279e-04,3.616548e-07)(3.162278e-04,1.143651e-06)(5.623413e-04,3.616520e-06)(1.000000e-03,1.143623e-05)(1.778279e-03,3.616252e-05)(3.162278e-03,1.143361e-04)(5.623413e-03,3.613668e-04)(1.000000e-02,1.140814e-03)(1.778279e-02,3.588639e-03)(3.162278e-02,1.116543e-02)(5.623413e-02,3.362089e-02)(1.000000e-01,9.252169e-02)(1.778279e-01,2.080378e-01)(3.162278e-01,3.448409e-01)(5.623413e-01,4.365652e-01)(1.000000e+00,4.775552e-01)(1.778279e+00,4.925689e-01)(3.162278e+00,4.976103e-01)(5.623413e+00,4.992400e-01)(1.000000e+01,4.997592e-01)
      };
      \legend{ {Wigner $W$},{i.i.d.\@ $W$} }
    \draw[->,thick] (axis cs:1e-4,1e-8) -- (axis cs:6e-5,1e-7) node [right,pos=0,font=\footnotesize] {$\tau=1,\ldots,4$};
    \draw[->,thick] (axis cs:1e-4,5e-7) -- (axis cs:2e-5,1e-2) node [above,pos=0,font=\footnotesize] {};
    \end{loglogaxis}
  \end{tikzpicture}
  &
  \begin{tikzpicture}[font=\footnotesize]
    \renewcommand{\axisdefaulttryminticks}{4} 
    \tikzstyle{every major grid}+=[style=densely dashed]       
    \tikzstyle{every axis y label}+=[yshift=-10pt] 
    \tikzstyle{every axis x label}+=[yshift=5pt]
    \tikzstyle{every axis legend}+=[cells={anchor=west},fill=white,
        at={(0.98,0.02)}, anchor=south east, font=\scriptsize ]
    \begin{loglogaxis}[
      width=0.45\linewidth,	    
      xmin=1e-5,
      ymin=1e-10,
      xmax=10,
      ymax=10,
      yticklabels = {},
      grid=major,
      scaled ticks=true,
      xlabel={$\eta^2$},
      mark options= {solid},
      ]
      \addplot[blue,smooth,line width=0.5pt] plot coordinates{
	      (1.000000e-05,1.422030e-07)(1.778279e-05,3.874947e-07)(3.162278e-05,1.038069e-06)(5.623413e-05,2.771892e-06)(1.000000e-04,7.348656e-06)(1.778279e-04,1.935932e-05)(3.162278e-04,5.063893e-05)(5.623413e-04,1.299926e-04)(1.000000e-03,3.341686e-04)(1.778279e-03,8.527597e-04)(3.162278e-03,2.140344e-03)(5.623413e-03,5.297477e-03)(1.000000e-02,1.290206e-02)(1.778279e-02,3.134860e-02)(3.162278e-02,7.304880e-02)(5.623413e-02,1.594906e-01)(1.000000e-01,3.236807e-01)(1.778279e-01,6.087636e-01)(3.162278e-01,1.062991e+00)(5.623413e-01,1.566168e+00)(1.000000e+00,1.892561e+00)(1.778279e+00,2.033649e+00)(3.162278e+00,2.083609e+00)(5.623413e+00,2.100013e+00)(1.000000e+01,2.105263e+00)
      };
      \addplot[red,smooth,line width=0.5pt] plot coordinates{
	      (1.000000e-05,9.777337e-10)(1.778279e-05,3.091865e-09)(3.162278e-05,9.777337e-09)(5.623413e-05,3.091865e-08)(1.000000e-04,9.777337e-08)(1.778279e-04,3.091865e-07)(3.162278e-04,9.777333e-07)(5.623413e-04,3.091861e-06)(1.000000e-03,9.777291e-06)(1.778279e-03,3.091819e-05)(3.162278e-03,9.776870e-05)(5.623413e-03,3.091396e-04)(1.000000e-02,9.772642e-04)(1.778279e-02,3.087180e-03)(3.162278e-02,9.730712e-03)(5.623413e-02,3.045773e-02)(1.000000e-01,9.331209e-02)(1.778279e-01,2.686134e-01)(3.162278e-01,6.618718e-01)(5.623413e-01,1.232645e+00)(1.000000e+00,1.694890e+00)(1.778279e+00,1.922949e+00)(3.162278e+00,2.008417e+00)(5.623413e+00,2.037050e+00)(1.000000e+01,2.046276e+00)
      };
      \addplot[blue,smooth,line width=0.5pt] plot coordinates{
	      (1.000000e-05,6.693060e-06)(1.778279e-05,1.578727e-05)(3.162278e-05,3.756807e-05)(5.623413e-05,8.934901e-05)(1.000000e-04,2.124942e-04)(1.778279e-04,5.110105e-04)(3.162278e-04,1.218022e-03)(5.623413e-04,2.811410e-03)(1.000000e-03,6.314557e-03)(1.778279e-03,1.400922e-02)(3.162278e-03,2.989686e-02)(5.623413e-03,6.119040e-02)(1.000000e-02,1.194778e-01)(1.778279e-02,2.175926e-01)(3.162278e-02,3.818153e-01)(5.623413e-02,6.354871e-01)(1.000000e-01,9.806659e-01)(1.778279e-01,1.398406e+00)(3.162278e-01,1.768746e+00)(5.623413e-01,1.978347e+00)(1.000000e+00,2.061259e+00)(1.778279e+00,2.087681e+00)(3.162278e+00,2.095456e+00)(5.623413e+00,2.097785e+00)(1.000000e+01,2.098504e+00)
      };
      \addplot[red,smooth,line width=0.5pt] plot coordinates{
	      (1.000000e-05,1.936637e-09)(1.778279e-05,6.124184e-09)(3.162278e-05,1.936637e-08)(5.623413e-05,6.124184e-08)(1.000000e-04,1.936637e-07)(1.778279e-04,6.124182e-07)(3.162278e-04,1.936635e-06)(5.623413e-04,6.124166e-06)(1.000000e-03,1.936619e-05)(1.778279e-03,6.124002e-05)(3.162278e-03,1.936454e-04)(5.623413e-03,6.122349e-04)(1.000000e-02,1.934795e-03)(1.778279e-02,6.105723e-03)(3.162278e-02,1.918274e-02)(5.623413e-02,5.944443e-02)(1.000000e-01,1.767894e-01)(1.778279e-01,4.706111e-01)(3.162278e-01,9.924096e-01)(5.623413e-01,1.529162e+00)(1.000000e+00,1.845699e+00)(1.778279e+00,1.975539e+00)(3.162278e+00,2.020625e+00)(5.623413e+00,2.035335e+00)(1.000000e+01,2.040033e+00)
      };
      \addplot[blue,smooth,line width=0.5pt] plot coordinates{
	      (1.000000e-05,1.400395e-04)(1.778279e-05,3.093919e-04)(3.162278e-05,6.521045e-04)(5.623413e-05,1.350657e-03)(1.000000e-04,2.754152e-03)(1.778279e-04,5.550186e-03)(3.162278e-04,1.147916e-02)(5.623413e-04,2.321007e-02)(1.000000e-03,4.500755e-02)(1.778279e-03,8.325297e-02)(3.162278e-03,1.467743e-01)(5.623413e-03,2.535053e-01)(1.000000e-02,4.148464e-01)(1.778279e-02,6.437616e-01)(3.162278e-02,9.449201e-01)(5.623413e-02,1.270543e+00)(1.000000e-01,1.582475e+00)(1.778279e-01,1.818590e+00)(3.162278e-01,1.955476e+00)(5.623413e-01,2.031519e+00)(1.000000e+00,2.070400e+00)(1.778279e+00,2.086559e+00)(3.162278e+00,2.092321e+00)(5.623413e+00,2.094227e+00)(1.000000e+01,2.094839e+00)
      };
      \addplot[red,smooth,line width=0.5pt] plot coordinates{
	      (1.000000e-05,3.407982e-09)(1.778279e-05,1.077699e-08)(3.162278e-05,3.407982e-08)(5.623413e-05,1.077698e-07)(1.000000e-04,3.407981e-07)(1.778279e-04,1.077698e-06)(3.162278e-04,3.407977e-06)(5.623413e-04,1.077693e-05)(1.000000e-03,3.407927e-05)(1.778279e-03,1.077643e-04)(3.162278e-03,3.407417e-04)(5.623413e-03,1.077130e-03)(1.000000e-02,3.402275e-03)(1.778279e-02,1.072003e-02)(3.162278e-02,3.351701e-02)(5.623413e-02,1.023408e-01)(1.000000e-01,2.918806e-01)(1.778279e-01,7.046626e-01)(3.162278e-01,1.275437e+00)(5.623413e-01,1.715577e+00)(1.000000e+00,1.926230e+00)(1.778279e+00,2.004124e+00)(3.162278e+00,2.030082e+00)(5.623413e+00,2.038430e+00)(1.000000e+01,2.041084e+00)
      };
      \addplot[blue,smooth,line width=0.5pt] plot coordinates{
	      (1.000000e-05,1.268604e-03)(1.778279e-05,2.606455e-03)(3.162278e-05,5.313942e-03)(5.623413e-05,1.052978e-02)(1.000000e-04,2.053842e-02)(1.778279e-04,3.910443e-02)(3.162278e-04,7.146890e-02)(5.623413e-04,1.288180e-01)(1.000000e-03,2.177280e-01)(1.778279e-03,3.459774e-01)(3.162278e-03,5.369133e-01)(5.623413e-03,7.772701e-01)(1.000000e-02,1.042256e+00)(1.778279e-02,1.292073e+00)(3.162278e-02,1.488479e+00)(5.623413e-02,1.644991e+00)(1.000000e-01,1.774613e+00)(1.778279e-01,1.896449e+00)(3.162278e-01,2.000439e+00)(5.623413e-01,2.061467e+00)(1.000000e+00,2.089214e+00)(1.778279e+00,2.100860e+00)(3.162278e+00,2.105320e+00)(5.623413e+00,2.106860e+00)(1.000000e+01,2.107364e+00)
      };
      \addplot[red,smooth,line width=0.5pt] plot coordinates{
	      (1.000000e-05,4.792194e-09)(1.778279e-05,1.515425e-08)(3.162278e-05,4.792194e-08)(5.623413e-05,1.515425e-07)(1.000000e-04,4.792193e-07)(1.778279e-04,1.515424e-06)(3.162278e-04,4.792183e-06)(5.623413e-04,1.515414e-05)(1.000000e-03,4.792086e-05)(1.778279e-03,1.515315e-04)(3.162278e-03,4.791090e-04)(5.623413e-03,1.514312e-03)(1.000000e-02,4.780998e-03)(1.778279e-02,1.504228e-02)(3.162278e-02,4.681850e-02)(5.623413e-02,1.410409e-01)(1.000000e-01,3.880068e-01)(1.778279e-01,8.698710e-01)(3.162278e-01,1.433417e+00)(5.623413e-01,1.803491e+00)(1.000000e+00,1.963787e+00)(1.778279e+00,2.020383e+00)(3.162278e+00,2.038901e+00)(5.623413e+00,2.044817e+00)(1.000000e+01,2.046693e+00)
      };
      \legend{ {Wigner $W$},{i.i.d.\@ $W$} }
    \draw[->,thick] (axis cs:1e-4,5e-8) -- (axis cs:6e-5,6e-7) node [right,pos=0,font=\footnotesize] {$\tau=1,\ldots,4$};
    \draw[->,thick] (axis cs:1e-4,2e-6) -- (axis cs:2e-5,2e-2) node [above,pos=0,font=\footnotesize] {};
    \end{loglogaxis}
  \end{tikzpicture}
  \end{tabular}
  \caption{Training (left) and testing (right) performance of a $\tau$-delay task for $\tau\in\{1,\ldots,4\}$ compared for i.i.d.\@ $W$ versus Wigner $W$, $\sigma=.9$ and $n=200$, $T=\hat{T}=400$ in both cases (here on the Mackey-Glass dataset).}
  \label{fig:MG_delays}
\end{figure}
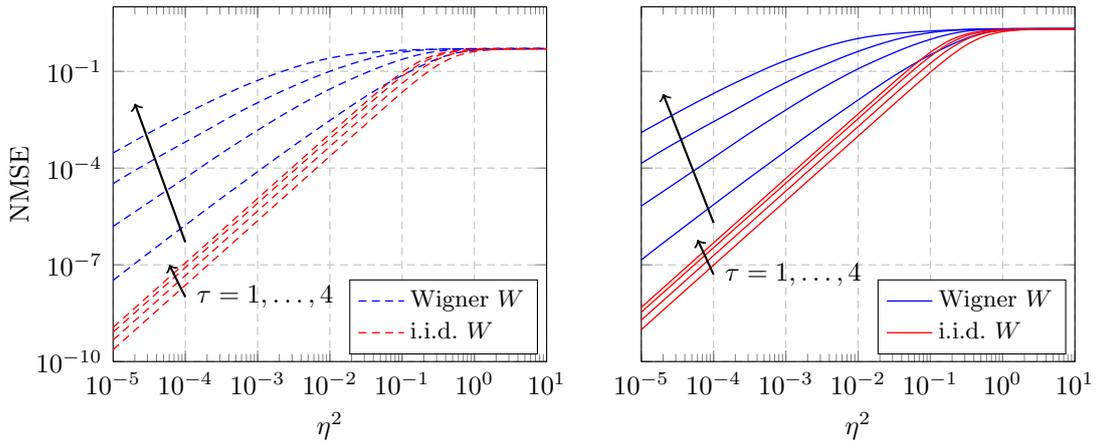

The final discussion in Remark~\ref{rem:proj_W} motivates a deeper comparative study of the performances of non-normal versus normal connectivity matrices. From \eqref{eq:R_normal}, we may in particular evaluate the memory curve (as defined here in Remark~\ref{rem:MC}) for $W$ a Wigner random matrix. The performance figures are displayed in Table~\ref{tab:MC}, which show a dramatic decay of the memory curve for the Wigner connectivity matrix as compared to an i.i.d.\@ Gaussian non-normal matrix. In Figure~\ref{fig:MG_delays}, a practical scenario of a $\tau$-delay task is depicted comparatively for Haar versus Wigner matrices (the input data being extracted from Mackey-Glass processes but the general results hold for any non-trivial input dataset); there we confirm that, for increasing values of the delay $\tau$, the ESN performance strongly decays for Wigner matrices as compared to Haar matrices, as predicted by the theoretical results of Table~\ref{tab:MC}.

\begin{table}[h]
	\centering
	\begin{tabular}{rrr}
		\toprule
		$\tau$ & \multicolumn{1}{l}{i.i.d.} &  \multicolumn{1}{l}{Wigner} \\
		\midrule
		$0$	& $5.2\cdot 10^{-1}$ & $4.8\cdot 10^{-1}$ \\
		$1$	& $2.0\cdot 10^{-1}$ & $1.6\cdot 10^{-2}$ \\
		$2$	& $1.0\cdot 10^{-1}$ & $1.3\cdot 10^{-3}$ \\
		$3$	& $6.0\cdot 10^{-2}$ & $2.0\cdot 10^{-4}$ \\
		$4$	& $3.9\cdot 10^{-2}$ & $5.7\cdot 10^{-5}$ \\
		\bottomrule
	\end{tabular}
	\medskip
	\caption{Memory curve ${\rm MC}(\tau)$ for i.i.d.\@ versus Wigner matrices, $c=.5$.}
	\label{tab:MC}
\end{table}

\begin{figure}[h!]
  \centering
  \begin{tikzpicture}[font=\footnotesize]
    \renewcommand{\axisdefaulttryminticks}{4} 
    \tikzstyle{every major grid}+=[style=densely dashed]       
    \tikzstyle{every axis y label}+=[yshift=-10pt] 
    \tikzstyle{every axis x label}+=[yshift=5pt]
    \tikzstyle{every axis legend}+=[cells={anchor=west},fill=white,
        at={(0.02,0.98)}, anchor=north west, font=\scriptsize ]
    \begin{loglogaxis}[
      xmin=1e-4,
      ymin=7e-3,
      xmax=10,
      ymax=5e-2,
      grid=major,
      scaled ticks=true,
      xlabel={$\eta^2$},
      mark options= {solid},
      ylabel={NMSE},
      ]
      \addplot[blue] plot coordinates{
	      (1.000000e-05,7.831683e-03)+-(7.576869e-04,7.576869e-04)(1.778279e-05,7.878959e-03)+-(7.114449e-04,7.114449e-04)(3.162278e-05,7.895727e-03)+-(7.173025e-04,7.173025e-04)(5.623413e-05,7.895771e-03)+-(7.450159e-04,7.450159e-04)(1.000000e-04,7.934525e-03)+-(7.431089e-04,7.431089e-04)(1.778279e-04,7.966876e-03)+-(7.427746e-04,7.427746e-04)(3.162278e-04,8.010324e-03)+-(7.302348e-04,7.302348e-04)(5.623413e-04,8.003719e-03)+-(7.389155e-04,7.389155e-04)(1.000000e-03,8.035890e-03)+-(7.480975e-04,7.480975e-04)(1.778279e-03,8.060107e-03)+-(7.402630e-04,7.402630e-04)(3.162278e-03,8.124300e-03)+-(7.741002e-04,7.741002e-04)(5.623413e-03,8.138224e-03)+-(7.686716e-04,7.686716e-04)(1.000000e-02,8.159249e-03)+-(7.693252e-04,7.693252e-04)(1.778279e-02,8.148573e-03)+-(7.656356e-04,7.656356e-04)(3.162278e-02,8.194528e-03)+-(7.684807e-04,7.684807e-04)(5.623413e-02,8.326500e-03)+-(7.700802e-04,7.700802e-04)(1.000000e-01,8.419457e-03)+-(7.793504e-04,7.793504e-04)(1.778279e-01,8.642996e-03)+-(8.242613e-04,8.242613e-04)(3.162278e-01,9.185964e-03)+-(8.712385e-04,8.712385e-04)(5.623413e-01,1.044713e-02)+-(1.040481e-03,1.040481e-03)(1.000000e+00,1.275010e-02)+-(1.311493e-03,1.311493e-03)(1.778279e+00,1.730252e-02)+-(1.769712e-03,1.769712e-03)(3.162278e+00,2.683664e-02)+-(2.822190e-03,2.822190e-03)(5.623413e+00,4.634975e-02)+-(5.379646e-03,5.379646e-03)(1.000000e+01,9.349567e-02)+-(1.298931e-02,1.298931e-02)
      };
      \addplot[red,smooth,line width=0.5pt] plot coordinates{
	      (1.000000e-05,7.869946e-03)(1.778279e-05,7.863820e-03)(3.162278e-05,7.876622e-03)(5.623413e-05,7.897969e-03)(1.000000e-04,7.922837e-03)(1.778279e-04,7.943443e-03)(3.162278e-04,7.957970e-03)(5.623413e-04,7.976727e-03)(1.000000e-03,8.007071e-03)(1.778279e-03,8.044203e-03)(3.162278e-03,8.077414e-03)(5.623413e-03,8.108169e-03)(1.000000e-02,8.137390e-03)(1.778279e-02,8.166865e-03)(3.162278e-02,8.209707e-03)(5.623413e-02,8.276183e-03)(1.000000e-01,8.383962e-03)(1.778279e-01,8.611868e-03)(3.162278e-01,9.146459e-03)(5.623413e-01,1.030454e-02)(1.000000e+00,1.260972e-02)(1.778279e+00,1.717153e-02)(3.162278e+00,2.636467e-02)(5.623413e+00,4.559095e-02)(1.000000e+01,9.166513e-02)
      };

      \addplot[blue,densely dashed] plot coordinates{
	      (1.000000e-05,8.851101e-03)+-(8.987121e-04,8.987121e-04)(1.778279e-05,8.873459e-03)+-(9.100718e-04,9.100718e-04)(3.162278e-05,8.867254e-03)+-(8.752238e-04,8.752238e-04)(5.623413e-05,8.827131e-03)+-(9.061901e-04,9.061901e-04)(1.000000e-04,8.926629e-03)+-(8.887444e-04,8.887444e-04)(1.778279e-04,8.888881e-03)+-(8.927794e-04,8.927794e-04)(3.162278e-04,8.990664e-03)+-(8.865274e-04,8.865274e-04)(5.623413e-04,8.991438e-03)+-(8.861730e-04,8.861730e-04)(1.000000e-03,9.018321e-03)+-(9.321128e-04,9.321128e-04)(1.778279e-03,9.016182e-03)+-(9.445166e-04,9.445166e-04)(3.162278e-03,9.077215e-03)+-(9.030429e-04,9.030429e-04)(5.623413e-03,9.108081e-03)+-(9.452377e-04,9.452377e-04)(1.000000e-02,9.097353e-03)+-(9.142751e-04,9.142751e-04)(1.778279e-02,9.103073e-03)+-(9.495407e-04,9.495407e-04)(3.162278e-02,9.110311e-03)+-(8.954745e-04,8.954745e-04)(5.623413e-02,9.196287e-03)+-(9.069689e-04,9.069689e-04)(1.000000e-01,9.241063e-03)+-(9.365010e-04,9.365010e-04)(1.778279e-01,9.464387e-03)+-(9.604261e-04,9.604261e-04)(3.162278e-01,9.903943e-03)+-(9.921715e-04,9.921715e-04)(5.623413e-01,1.096649e-02)+-(1.142882e-03,1.142882e-03)(1.000000e+00,1.329608e-02)+-(1.308743e-03,1.308743e-03)(1.778279e+00,1.763825e-02)+-(1.741627e-03,1.741627e-03)(3.162278e+00,2.491289e-02)+-(2.583210e-03,2.583210e-03)(5.623413e+00,3.861636e-02)+-(4.151125e-03,4.151125e-03)(1.000000e+01,7.066071e-02)+-(7.824110e-03,7.824110e-03)
      };
      \addplot[red,densely dashed,smooth,line width=0.5pt] plot coordinates{
	      (1.000000e-05,8.790946e-03)(1.778279e-05,8.801654e-03)(3.162278e-05,8.806500e-03)(5.623413e-05,8.820259e-03)(1.000000e-04,8.850242e-03)(1.778279e-04,8.889229e-03)(3.162278e-04,8.928817e-03)(5.623413e-04,8.955638e-03)(1.000000e-03,8.965594e-03)(1.778279e-03,8.967148e-03)(3.162278e-03,8.972461e-03)(5.623413e-03,8.987221e-03)(1.000000e-02,9.006785e-03)(1.778279e-02,9.028313e-03)(3.162278e-02,9.059660e-03)(5.623413e-02,9.103174e-03)(1.000000e-01,9.169815e-03)(1.778279e-01,9.346808e-03)(3.162278e-01,9.839574e-03)(5.623413e-01,1.097056e-02)(1.000000e+00,1.325185e-02)(1.778279e+00,1.762217e-02)(3.162278e+00,2.512707e-02)(5.623413e+00,3.811309e-02)(1.000000e+01,6.474811e-02)
      };
      \legend{ {Monte Carlo}, {Theory (limit)} }
    \draw[->,thick] (axis cs:0.2,.015) -- (axis cs:0.2,.01) node [above,pos=0,font=\footnotesize] {$W$ normal};
    \draw[->,thick] (axis cs:1e-3,.012) -- (axis cs:1e-3,.0082) node [above,pos=0,font=\footnotesize] {$W$ non-normal};
    \end{loglogaxis}
  \end{tikzpicture}
  \caption{Testing (normalized) MSE for the PM10 one-step ahead task, $W$ i.i.d.\@ Gaussian or Gaussian Wigner ($\sigma=.9$), $n=200$, $T=\hat{T}=400$.}
  \label{fig:PM10}
\end{figure}
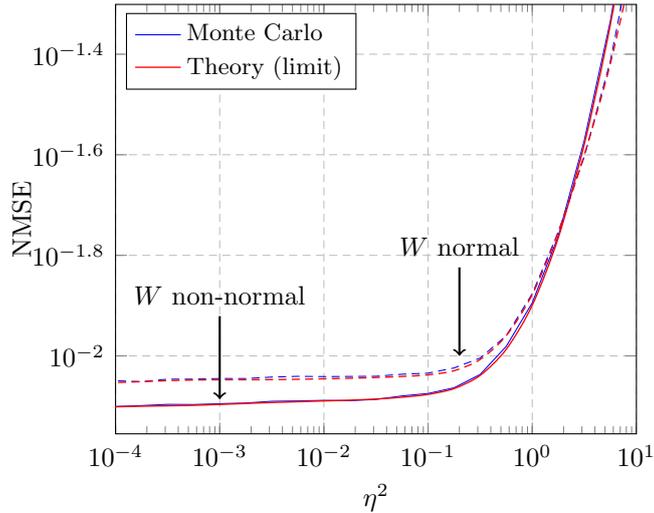

An application example in a less artificial context is devised in Figure~\ref{fig:PM10}, where, on a real dataset of daily pollution (PM10) records, we provide the one-day ahead interpolation performance of neural networks assuming $m$ random i.i.d.\@ and either (i) $W$ with i.i.d.\@ Gaussian entries or (ii) $W$ Gaussian Wigner. We observe again a better performance achieved by the ESN with non-normal matrix $W$ which, accordingly with the fact that ESN's rely heavily on past input retrieval, is coherent with the previous remark.
 
\subsection{Further Experiments}
\label{sec:experiments}

In this section, we provide further noticeable results of interest to neural network optimization. 

To start with, we consider a scenario where the testing dataset is polluted by an additional impulsive white Gaussian noise arising independently with probability $p$. This is depicted in Figure~\ref{fig:robustness} for the Mackey--Glass one-step ahead task. It is observed here that the in-network noise is valuable in bringing the normalized MSE down to acceptable values. It is in particular seen that the more the noise impulsion probability the larger the variance $\eta^2$ should be chosen. A particular realization of the noisy Mackey-Glass output is provided in Figure~\ref{fig:robustness_oneshot}, where it is observed that a visually small noise impulsion in the input vector drives a large fluctuation of the output for a too small-$\eta^2$ ESN. 

This phenomenon can be theoretically anticipated in simple settings. Let us consider the scenario of Section~\ref{sec:non-Hermitian} with $W$ orthogonally invariant, where $r=U^\trans b$ for a vector $b\in\RR^T$ having only its last $T-k$ entries identically zero for some fixed $k$; let us now assume that $\hat{u}=\hat{u}_0+\hat{e}$ for some noise vector $\hat{e}$ made of i.i.d.\@ zero mean and variance $s^2$ entries, and suppose that $\hat{r}=\hat{U}_0$ for $\{\hat{U}_0\}_{ij}=[\hat{u}_0]_{i-j}$. Then, an application of Corollary~\ref{cor:unitarily_invariant} leads to $\hat{E}_\eta(u,r,\hat{u},\hat{r})$ asymptotically equal to \eqref{eq:hatE_r=Ub} plus an additional term given by (after calculus)
\begin{align}
	\label{eq:MSE_impulsive_noise}
	s^2 \left\| \left( \eta^2I_T + D^{\frac12}UU^\trans D^{\frac12} \right)^{-1} D^{\frac12}UU^\trans D^{\frac12} (D^{-\frac12}b) \right\|^2.
\end{align}
From the inequality $\|(\eta^2I_T + D^{\frac12}UU^\trans D^{\frac12})^{-1}\|\leq \eta^{-2}$ and the fact that $\|D^{-\frac12}b\|$ remains bounded, we get that the term \eqref{eq:MSE_impulsive_noise} can be made arbitrarily small by letting $\eta^2\to\infty$. Therefore, $\eta^2$ induces robustness in this scenario. Since $\eta^2\to 0$ was shown to be optimal in the scenario where $s^2=0$, there must exist an MSE minimizing choice of $\eta^2\in(0,\infty)$.

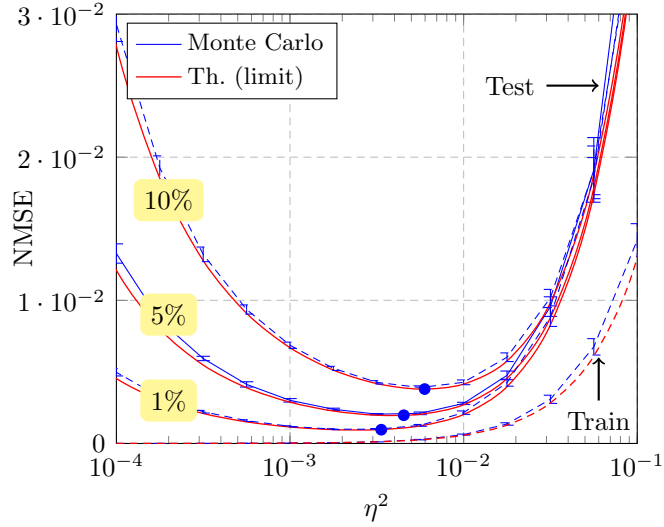
\begin{figure}[h!]
  \centering
  \begin{tikzpicture}[font=\footnotesize]
    \renewcommand{\axisdefaulttryminticks}{4} 
    \tikzstyle{every major grid}+=[style=densely dashed]       
    \tikzstyle{every axis y label}+=[yshift=-10pt] 
    \tikzstyle{every axis x label}+=[yshift=5pt]
    \tikzstyle{every axis legend}+=[cells={anchor=west},fill=white,
        at={(0.02,0.98)}, anchor=north west, font=\scriptsize ]
    \begin{semilogxaxis}[
      xmin=1e-4,
      ymin=0,
      xmax=.1,
      ymax=3e-2,
      grid=major,
      scaled ticks=true,
      xlabel={$\eta^2$},
      mark options= {solid},
      ylabel={NMSE},
      ]
      \addplot[blue,error bars/.cd,y dir=both,y explicit, error bar style={mark size=2.5pt}] plot coordinates{
	      (1.000000e-05,9.266942e-02)+-(2.205364e-03,2.205364e-03)(1.778279e-05,5.344331e-02)+-(1.899157e-03,1.899157e-03)(3.162278e-05,3.237209e-02)+-(9.907475e-04,9.907475e-04)(5.623413e-05,2.020548e-02)+-(9.076916e-04,9.076916e-04)(1.000000e-04,1.327173e-02)+-(6.738140e-04,6.738140e-04)(1.778279e-04,8.602802e-03)+-(2.383578e-04,2.383578e-04)(3.162278e-04,5.938515e-03)+-(1.523721e-04,1.523721e-04)(5.623413e-04,4.126662e-03)+-(1.847970e-04,1.847970e-04)(1.000000e-03,3.013037e-03)+-(1.223416e-04,1.223416e-04)(1.778279e-03,2.389597e-03)+-(7.628705e-05,7.628705e-05)(3.162278e-03,2.046092e-03)+-(4.241632e-05,4.241632e-05)(5.623413e-03,2.105776e-03)+-(9.246081e-05,9.246081e-05)(1.000000e-02,2.798513e-03)+-(7.988405e-05,7.988405e-05)(1.778279e-02,4.758675e-03)+-(3.020511e-04,3.020511e-04)(3.162278e-02,9.561720e-03)+-(6.859567e-04,6.859567e-04)(5.623413e-02,1.911203e-02)+-(2.258611e-03,2.258611e-03)(1.000000e-01,4.306590e-02)+-(2.382971e-03,2.382971e-03)(1.778279e-01,9.316174e-02)+-(9.192438e-03,9.192438e-03)(3.162278e-01,2.197055e-01)+-(1.555560e-02,1.555560e-02)(5.623413e-01,5.163905e-01)+-(6.774560e-02,6.774560e-02)(1.000000e+00,1.065848e+00)+-(1.195074e-01,1.195074e-01)(1.778279e+00,1.409089e+00)+-(1.564828e-01,1.564828e-01)(3.162278e+00,1.561604e+00)+-(1.377846e-01,1.377846e-01)(5.623413e+00,1.694480e+00)+-(3.203572e-01,3.203572e-01)(1.000000e+01,1.718221e+00)+-(1.831747e-01,1.831747e-01)
      };
      \addplot[red,smooth,line width=0.5pt] plot coordinates{
	      (1.000000e-05,1.902730e-01)(1.778279e-05,1.121210e-01)(3.162278e-05,6.826918e-02)(5.623413e-05,4.290060e-02)(1.000000e-04,2.775432e-02)(1.778279e-04,1.852713e-02)(3.162278e-04,1.277629e-02)(5.623413e-04,9.092024e-03)(1.000000e-03,6.690072e-03)(1.778279e-03,5.125467e-03)(3.162278e-03,4.182371e-03)(5.623413e-03,3.800434e-03)(1.000000e-02,4.146723e-03)(1.778279e-02,5.763292e-03)(3.162278e-02,9.845574e-03)(5.623413e-02,1.880355e-02)(1.000000e-01,3.779206e-02)(1.778279e-01,8.163702e-02)(3.162278e-01,1.949001e-01)(5.623413e-01,4.621556e-01)(1.000000e+00,9.014267e-01)(1.778279e+00,1.312463e+00)(3.162278e+00,1.537113e+00)(5.623413e+00,1.625476e+00)(1.000000e+01,1.655615e+00)
      };
      \addplot[blue,densely dashed,error bars/.cd,y dir=both,y explicit, error bar style={mark size=2.5pt}] plot coordinates{
	      (1.000000e-05,1.999122e-01)+-(1.165274e-02,1.165274e-02)(1.778279e-05,1.179620e-01)+-(6.207318e-03,6.207318e-03)(3.162278e-05,7.230353e-02)+-(3.372454e-03,3.372454e-03)(5.623413e-05,4.533758e-02)+-(1.758874e-03,1.758874e-03)(1.000000e-04,2.928841e-02)+-(1.204343e-03,1.204343e-03)(1.778279e-04,1.926200e-02)+-(8.316652e-04,8.316652e-04)(3.162278e-04,1.321246e-02)+-(5.100372e-04,5.100372e-04)(5.623413e-04,9.359216e-03)+-(3.198242e-04,3.198242e-04)(1.000000e-03,6.860732e-03)+-(2.181628e-04,2.181628e-04)(1.778279e-03,5.246398e-03)+-(1.444253e-04,1.444253e-04)(3.162278e-03,4.269821e-03)+-(1.188203e-04,1.188203e-04)(5.623413e-03,3.907819e-03)+-(1.169897e-04,1.169897e-04)(1.000000e-02,4.272148e-03)+-(1.652352e-04,1.652352e-04)(1.778279e-02,6.025817e-03)+-(3.022262e-04,3.022262e-04)(3.162278e-02,1.018876e-02)+-(5.757718e-04,5.757718e-04)(5.623413e-02,1.908097e-02)+-(1.700886e-03,1.700886e-03)(1.000000e-01,3.756942e-02)+-(3.493657e-03,3.493657e-03)(1.778279e-01,7.910042e-02)+-(7.360902e-03,7.360902e-03)(3.162278e-01,1.860484e-01)+-(2.184569e-02,2.184569e-02)(5.623413e-01,4.398440e-01)+-(3.964191e-02,3.964191e-02)(1.000000e+00,8.885452e-01)+-(1.088560e-01,1.088560e-01)(1.778279e+00,1.302938e+00)+-(1.387563e-01,1.387563e-01)(3.162278e+00,1.541573e+00)+-(1.737746e-01,1.737746e-01)(5.623413e+00,1.643376e+00)+-(1.810153e-01,1.810153e-01)(1.000000e+01,1.662714e+00)+-(1.809860e-01,1.809860e-01)
      };
      \addplot[red,smooth,line width=0.5pt] plot coordinates{
	      (1.000000e-05,8.735758e-02)(1.778279e-05,5.081684e-02)(3.162278e-05,3.052923e-02)(5.623413e-05,1.893997e-02)(1.000000e-04,1.210857e-02)(1.778279e-04,8.001003e-03)(3.162278e-04,5.479209e-03)(5.623413e-04,3.892266e-03)(1.000000e-03,2.884493e-03)(1.778279e-03,2.266923e-03)(3.162278e-03,1.971719e-03)(5.623413e-03,2.039840e-03)(1.000000e-02,2.702739e-03)(1.778279e-02,4.548313e-03)(3.162278e-02,8.797904e-03)(5.623413e-02,1.787517e-02)(1.000000e-01,3.695681e-02)(1.778279e-01,8.093401e-02)(3.162278e-01,1.945379e-01)(5.623413e-01,4.625415e-01)(1.000000e+00,9.027725e-01)(1.778279e+00,1.314390e+00)(3.162278e+00,1.539227e+00)(5.623413e+00,1.627639e+00)(1.000000e+01,1.657790e+00)
      };
      \addplot[blue,densely dashed,error bars/.cd,y dir=both,y explicit, error bar style={mark size=2.5pt}] plot coordinates{
	      (1.000000e-05,2.017514e-07)+-(8.343298e-09,8.343298e-09)(1.778279e-05,3.034840e-07)+-(9.628908e-09,9.628908e-09)(3.162278e-05,5.325296e-07)+-(1.989360e-08,1.989360e-08)(5.623413e-05,9.276843e-07)+-(4.748887e-08,4.748887e-08)(1.000000e-04,1.625561e-06)+-(9.327995e-08,9.327995e-08)(1.778279e-04,3.131558e-06)+-(1.695778e-07,1.695778e-07)(3.162278e-04,6.108542e-06)+-(2.898126e-07,2.898126e-07)(5.623413e-04,1.275250e-05)+-(6.297786e-07,6.297786e-07)(1.000000e-03,2.584861e-05)+-(6.885644e-07,6.885644e-07)(1.778279e-03,5.453654e-05)+-(1.877178e-06,1.877178e-06)(3.162278e-03,1.195194e-04)+-(8.633280e-06,8.633280e-06)(5.623413e-03,2.681091e-04)+-(1.653968e-05,1.653968e-05)(1.000000e-02,6.200346e-04)+-(3.440783e-05,3.440783e-05)(1.778279e-02,1.361309e-03)+-(8.383084e-05,8.383084e-05)(3.162278e-02,3.085953e-03)+-(2.855628e-04,2.855628e-04)(5.623413e-02,6.750764e-03)+-(5.724781e-04,5.724781e-04)(1.000000e-01,1.421327e-02)+-(1.141856e-03,1.141856e-03)(1.778279e-01,3.375805e-02)+-(5.645159e-03,5.645159e-03)(3.162278e-01,7.630956e-02)+-(1.318271e-02,1.318271e-02)(5.623413e-01,1.960117e-01)+-(3.133028e-02,3.133028e-02)(1.000000e+00,3.500434e-01)+-(3.355249e-02,3.355249e-02)(1.778279e+00,4.993116e-01)+-(4.854789e-02,4.854789e-02)(3.162278e+00,5.245462e-01)+-(8.559600e-02,8.559600e-02)(5.623413e+00,5.651148e-01)+-(5.805380e-02,5.805380e-02)(1.000000e+01,5.857640e-01)+-(6.262577e-02,6.262577e-02)
      };
      \addplot[red,smooth,line width=0.5pt] plot coordinates{
	      (1.000000e-05,3.176783e-02)(1.778279e-05,1.860223e-02)(3.162278e-05,1.125956e-02)(5.623413e-05,7.042176e-03)(1.000000e-04,4.540162e-03)(1.778279e-04,3.026383e-03)(3.162278e-04,2.093164e-03)(5.623413e-04,1.508973e-03)(1.000000e-03,1.151600e-03)(1.778279e-03,9.665767e-04)(3.162278e-03,9.617400e-04)(5.623413e-03,1.227507e-03)(1.000000e-02,2.026655e-03)(1.778279e-02,3.969819e-03)(3.162278e-02,8.288984e-03)(5.623413e-02,1.741262e-02)(1.000000e-01,3.652544e-02)(1.778279e-01,8.052373e-02)(3.162278e-01,1.941457e-01)(5.623413e-01,4.621874e-01)(1.000000e+00,9.025116e-01)(1.778279e+00,1.314255e+00)(3.162278e+00,1.539176e+00)(5.623413e+00,1.627624e+00)(1.000000e+01,1.657788e+00)
      };
      \addplot[blue,densely dashed,error bars/.cd,y dir=both,y explicit, error bar style={mark size=2.5pt}] plot coordinates{
	      (1.000000e-05,3.741626e-02)+-(2.885524e-03,2.885524e-03)(1.778279e-05,2.153592e-02)+-(1.284286e-03,1.284286e-03)(3.162278e-05,1.275157e-02)+-(7.430913e-04,7.430913e-04)(5.623413e-05,7.850338e-03)+-(4.507832e-04,4.507832e-04)(1.000000e-04,4.972364e-03)+-(2.545621e-04,2.545621e-04)(1.778279e-04,3.244505e-03)+-(1.508225e-04,1.508225e-04)(3.162278e-04,2.216842e-03)+-(9.470151e-05,9.470151e-05)(5.623413e-04,1.593366e-03)+-(5.969698e-05,5.969698e-05)(1.000000e-03,1.195221e-03)+-(4.103728e-05,4.103728e-05)(1.778279e-03,9.974005e-04)+-(3.423749e-05,3.423749e-05)(3.162278e-03,9.956364e-04)+-(4.513701e-05,4.513701e-05)(5.623413e-03,1.265420e-03)+-(6.287766e-05,6.287766e-05)(1.000000e-02,2.153750e-03)+-(1.192312e-04,1.192312e-04)(1.778279e-02,4.246629e-03)+-(2.425275e-04,2.425275e-04)(3.162278e-02,8.837714e-03)+-(6.602952e-04,6.602952e-04)(5.623413e-02,1.853698e-02)+-(1.436749e-03,1.436749e-03)(1.000000e-01,3.850639e-02)+-(3.164984e-03,3.164984e-03)(1.778279e-01,8.640454e-02)+-(8.033846e-03,8.033846e-03)(3.162278e-01,2.110883e-01)+-(2.113059e-02,2.113059e-02)(5.623413e-01,4.966386e-01)+-(4.884711e-02,4.884711e-02)(1.000000e+00,9.412486e-01)+-(1.054522e-01,1.054522e-01)(1.778279e+00,1.344678e+00)+-(1.553212e-01,1.553212e-01)(3.162278e+00,1.573836e+00)+-(1.914283e-01,1.914283e-01)(5.623413e+00,1.666478e+00)+-(1.759271e-01,1.759271e-01)(1.000000e+01,1.665490e+00)+-(1.955402e-01,1.955402e-01)
      };
      \addplot[red,densely dashed,smooth,line width=0.5pt] plot coordinates{
	      (1.000000e-05,1.976085e-07)(1.778279e-05,3.158408e-07)(3.162278e-05,5.285188e-07)(5.623413e-05,9.238109e-07)(1.000000e-04,1.677355e-06)(1.778279e-04,3.158504e-06)(3.162278e-04,6.167902e-06)(5.623413e-04,1.242447e-05)(1.000000e-03,2.569662e-05)(1.778279e-03,5.429364e-05)(3.162278e-03,1.170870e-04)(5.623413e-03,2.573151e-04)(1.000000e-02,5.740745e-04)(1.778279e-02,1.285295e-03)(3.162278e-02,2.833325e-03)(5.623413e-02,6.114514e-03)(1.000000e-01,1.303857e-02)(1.778279e-01,2.891029e-02)(3.162278e-01,6.962227e-02)(5.623413e-01,1.655433e-01)(1.000000e+00,3.233430e-01)(1.778279e+00,4.712169e-01)(3.162278e+00,5.521287e-01)(5.623413e+00,5.839723e-01)(1.000000e+01,5.948356e-01)
      };
      \legend{ {Monte Carlo}, {Th.\@ (limit)} }
      \draw[->,thick] (axis cs:.03,.025) -- (axis cs:.06,.025) node [left,pos=0,font=\footnotesize] {Test};
      \draw[->,thick] (axis cs:.06,.003) -- (axis cs:.06,.006) node [below,pos=0,font=\footnotesize] {Train};
      \node[color=blue] at (axis cs:3.162278e-03,9.617400e-04) {\pgfuseplotmark{*}};
      \node[color=blue] at (axis cs:4.262278e-03,1.971719e-03) {\pgfuseplotmark{*}};
      \node[color=blue] at (axis cs:5.623413e-03,3.800434e-03) {\pgfuseplotmark{*}};
      \node[fill=yellow!50!white,rectangle,rounded corners=3pt,font=\footnotesize] at (axis cs:.0002,.003) {$1\%$};
      \node[fill=yellow!50!white,rectangle,rounded corners=3pt,font=\footnotesize] at (axis cs:.0002,.009) {$5\%$};
      \node[fill=yellow!50!white,rectangle,rounded corners=3pt,font=\footnotesize] at (axis cs:.0002,.017) {$10\%$};
    \end{semilogxaxis}
  \end{tikzpicture}
  \caption{Testing (normalized) MSE for the Mackey-Glass one-step ahead task with $1\%$ or $10\%$ impulsive $\mathcal N(0,.01)$ noise pollution in test data inputs, $W$ Haar with $\sigma=.9$, $n=400$, $T=\hat{T}=1000$. Circles indicate the NMSE theoretical minima. Error bars indicate one standard deviation of the Monte Carlo simulations.}
  \label{fig:robustness}
\end{figure}
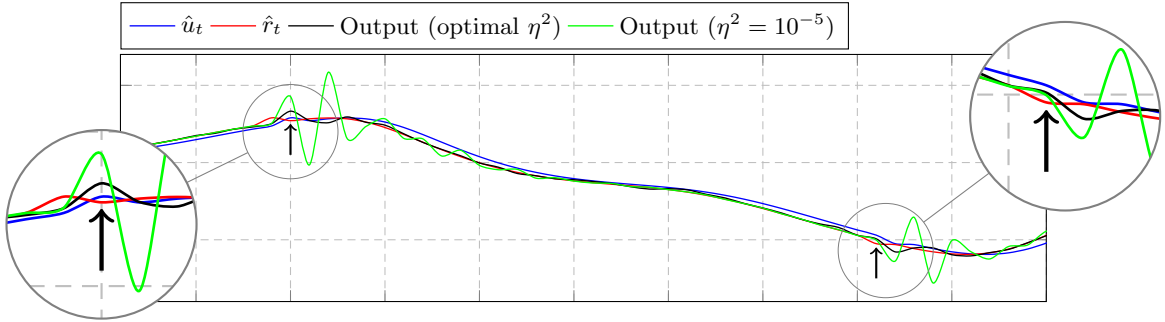
\begin{figure}[h!]
  \centering
  \begin{tikzpicture}[font=\footnotesize,spy using outlines={circle, magnification=2, connect spies}]
    \renewcommand{\axisdefaulttryminticks}{4} 
    \tikzstyle{every major grid}+=[style=densely dashed]       
    \tikzstyle{every axis y label}+=[yshift=-10pt] 
    \tikzstyle{every axis x label}+=[yshift=5pt]
    \tikzstyle{every axis legend}+=[cells={anchor=west},fill=white,
        at={(0,1.02)}, anchor=south west, font=\scriptsize ]
    \begin{axis}[
      legend columns=-1,
      width=0.85\linewidth,
      height=0.3\linewidth,
      xmin=71,
      ymin=0.1,
      xmax=120,
      ymax=1.7,
      xticklabels={},
      yticklabels = {},
      grid=major,
      scaled ticks=true,
      xlabel={},
      ylabel={},
      mark options= {solid},
      ]
      \addplot[blue,smooth,line width=0.5pt] plot coordinates{
(1,0.569152)(2,0.543927)(3,0.524108)(4,0.509134)(5,0.497825)(6,0.488414)(7,0.478874)(8,0.467469)(9,0.453298)(10,0.436525)(11,0.418179)(12,0.399724)(13,0.382685)(14,0.368495)(15,0.358505)(16,0.354047)(17,0.356501)(18,0.367353)(19,0.388258)(20,0.420999)(21,0.467080)(22,0.526612)(23,0.596876)(24,0.672074)(25,0.745263)(26,0.810917)(27,0.866051)(28,0.909887)(29,0.943026)(30,0.966770)(31,0.982709)(32,0.992481)(33,0.997629)(34,0.999479)(35,0.999027)(36,0.996842)(37,0.993064)(38,0.987489)(39,0.979759)(40,0.969583)(41,0.956892)(42,0.941902)(43,0.925098)(44,0.907194)(45,0.889093)(46,0.871866)(47,0.856746)(48,0.845129)(49,0.838594)(50,0.838904)(51,0.847925)(52,0.867298)(53,0.897716)(54,0.937839)(55,0.983313)(56,1.027177)(57,1.062638)(58,1.086196)(59,1.098164)(60,1.101059)(61,1.097932)(62,1.091460)(63,1.083657)(64,1.075894)(65,1.069042)(66,1.063672)(67,1.060240)(68,1.059224)(69,1.061162)(70,1.066585)(71,1.075891)(72,1.089192)(73,1.106214)(74,1.126281)(75,1.148408)(76,1.171470)(77,1.194409)(78,1.216383)(79,1.236812)(80,1.289570)(81,1.271191)(82,1.283254)(83,1.288649)(84,1.282959)(85,1.262299)(86,1.226897)(87,1.181458)(88,1.131939)(89,1.083084)(90,1.037923)(91,0.998091)(92,0.964221)(93,0.936255)(94,0.913680)(95,0.895712)(96,0.881411)(97,0.869733)(98,0.859542)(99,0.849605)(100,0.838627)(101,0.825344)(102,0.808685)(103,0.787962)(104,0.763009)(105,0.734202)(106,0.702332)(107,0.668398)(108,0.633405)(109,0.598235)(110,0.563605)(111,0.530102)(112,0.474709)(113,0.468865)(114,0.442912)(115,0.422079)(116,0.408666)(117,0.405475)(118,0.415266)(119,0.439739)(120,0.478491)(121,0.528845)(122,0.586810)(123,0.648390)(124,0.710417)(125,0.770757)(126,0.828162)(127,0.882026)(128,0.932168)(129,0.978667)(130,1.021714)(131,1.061476)(132,1.097933)(133,1.130751)(134,1.159262)(135,1.182598)(136,1.199937)(137,1.210737)(138,1.214833)(139,1.212420)(140,1.203966)(141,1.190109)(142,1.171602)(143,1.149300)(144,1.124203)(145,1.097539)(146,1.070863)(147,1.046120)(148,1.025592)(149,1.011665)(150,1.006404)(151,1.011113)(152,1.026039)(153,1.050204)(154,1.081187)(155,1.114887)(156,1.145643)(157,1.167262)(158,1.174824)(159,1.166160)(160,1.142011)(161,1.105078)(162,1.058855)(163,1.006813)(164,0.951986)(165,0.896837)(166,0.843262)(167,0.792677)(168,0.746142)(169,0.704483)(170,0.668410)(171,0.638610)(172,0.615806)(173,0.600769)(174,0.594249)(175,0.596769)(176,0.608301)(177,0.627895)(178,0.653474)(179,0.652321)(180,0.709801)(181,0.733163)(182,0.748553)(183,0.753260)(184,0.746231)(185,0.728708)(186,0.703908)(187,0.675928)(188,0.648817)(189,0.626348)(190,0.612280)(191,0.610613)(192,0.625217)(193,0.658279)(194,0.708078)(195,0.768399)(196,0.831080)(197,0.889400)(198,0.939563)(199,0.980313)(200,1.012014)
      };
      \addplot[red,smooth,line width=0.5pt] plot coordinates{
	      (1,0.543927)(2,0.524108)(3,0.509134)(4,0.497825)(5,0.488414)(6,0.478874)(7,0.467469)(8,0.453298)(9,0.436525)(10,0.418179)(11,0.399724)(12,0.382685)(13,0.368495)(14,0.358505)(15,0.354047)(16,0.356501)(17,0.367353)(18,0.388258)(19,0.420999)(20,0.467080)(21,0.526612)(22,0.596876)(23,0.672074)(24,0.745263)(25,0.810917)(26,0.866051)(27,0.909887)(28,0.943026)(29,0.966770)(30,0.982709)(31,0.992481)(32,0.997629)(33,0.999479)(34,0.999027)(35,0.996842)(36,0.993064)(37,0.987489)(38,0.979759)(39,0.969583)(40,0.956892)(41,0.941902)(42,0.925098)(43,0.907194)(44,0.889093)(45,0.871866)(46,0.856746)(47,0.845129)(48,0.838594)(49,0.838904)(50,0.847925)(51,0.867298)(52,0.897716)(53,0.937839)(54,0.983313)(55,1.027177)(56,1.062638)(57,1.086196)(58,1.098164)(59,1.101059)(60,1.097932)(61,1.091460)(62,1.083657)(63,1.075894)(64,1.069042)(65,1.063672)(66,1.060240)(67,1.059224)(68,1.061162)(69,1.066585)(70,1.075891)(71,1.089192)(72,1.106214)(73,1.126281)(74,1.148408)(75,1.171470)(76,1.194409)(77,1.216383)(78,1.236812)(79,1.289570)(80,1.271191)(81,1.283254)(82,1.288649)(83,1.282959)(84,1.262299)(85,1.226897)(86,1.181458)(87,1.131939)(88,1.083084)(89,1.037923)(90,0.998091)(91,0.964221)(92,0.936255)(93,0.913680)(94,0.895712)(95,0.881411)(96,0.869733)(97,0.859542)(98,0.849605)(99,0.838627)(100,0.825344)(101,0.808685)(102,0.787962)(103,0.763009)(104,0.734202)(105,0.702332)(106,0.668398)(107,0.633405)(108,0.598235)(109,0.563605)(110,0.530102)(111,0.474709)(112,0.468865)(113,0.442912)(114,0.422079)(115,0.408666)(116,0.405475)(117,0.415266)(118,0.439739)(119,0.478491)(120,0.528845)(121,0.586810)(122,0.648390)(123,0.710417)(124,0.770757)(125,0.828162)(126,0.882026)(127,0.932168)(128,0.978667)(129,1.021714)(130,1.061476)(131,1.097933)(132,1.130751)(133,1.159262)(134,1.182598)(135,1.199937)(136,1.210737)(137,1.214833)(138,1.212420)(139,1.203966)(140,1.190109)(141,1.171602)(142,1.149300)(143,1.124203)(144,1.097539)(145,1.070863)(146,1.046120)(147,1.025592)(148,1.011665)(149,1.006404)(150,1.011113)(151,1.026039)(152,1.050204)(153,1.081187)(154,1.114887)(155,1.145643)(156,1.167262)(157,1.174824)(158,1.166160)(159,1.142011)(160,1.105078)(161,1.058855)(162,1.006813)(163,0.951986)(164,0.896837)(165,0.843262)(166,0.792677)(167,0.746142)(168,0.704483)(169,0.668410)(170,0.638610)(171,0.615806)(172,0.600769)(173,0.594249)(174,0.596769)(175,0.608301)(176,0.627895)(177,0.653474)(178,0.652321)(179,0.709801)(180,0.733163)(181,0.748553)(182,0.753260)(183,0.746231)(184,0.728708)(185,0.703908)(186,0.675928)(187,0.648817)(188,0.626348)(189,0.612280)(190,0.610613)(191,0.625217)(192,0.658279)(193,0.708078)(194,0.768399)(195,0.831080)(196,0.889400)(197,0.939563)(198,0.980313)(199,1.012014)(200,1.035896)
      };
      \addplot[black,smooth,line width=0.5pt] plot coordinates{
	      (1,0.549587)(2,0.550250)(3,0.504403)(4,0.486784)(5,0.498421)(6,0.476763)(7,0.471121)(8,0.469949)(9,0.433671)(10,0.433813)(11,0.397131)(12,0.393783)(13,0.370524)(14,0.349978)(15,0.360400)(16,0.349802)(17,0.365824)(18,0.386213)(19,0.401054)(20,0.467238)(21,0.526350)(22,0.583734)(23,0.669708)(24,0.749549)(25,0.828743)(26,0.870354)(27,0.912994)(28,0.950842)(29,0.974370)(30,0.980962)(31,0.993256)(32,0.993442)(33,1.014604)(34,0.985360)(35,1.000064)(36,0.993082)(37,0.978305)(38,0.971865)(39,0.963293)(40,0.962590)(41,0.947192)(42,0.913264)(43,0.906625)(44,0.892287)(45,0.865859)(46,0.861244)(47,0.839906)(48,0.828868)(49,0.839458)(50,0.842841)(51,0.859293)(52,0.892363)(53,0.927447)(54,0.980381)(55,1.032462)(56,1.070246)(57,1.089929)(58,1.103093)(59,1.094354)(60,1.094885)(61,1.088243)(62,1.080663)(63,1.070317)(64,1.060731)(65,1.056084)(66,1.055738)(67,1.058648)(68,1.050508)(69,1.069627)(70,1.074682)(71,1.075778)(72,1.102154)(73,1.126154)(74,1.148853)(75,1.174836)(76,1.189077)(77,1.216103)(78,1.230294)(79,1.252014)(80,1.332680)(81,1.272362)(82,1.257802)(83,1.295529)(84,1.260382)(85,1.242127)(86,1.186137)(87,1.130659)(88,1.085238)(89,1.044214)(90,0.997277)(91,0.970074)(92,0.929287)(93,0.914580)(94,0.894694)(95,0.888921)(96,0.877611)(97,0.863910)(98,0.843443)(99,0.828178)(100,0.830755)(101,0.823266)(102,0.794479)(103,0.770958)(104,0.735813)(105,0.706090)(106,0.673073)(107,0.633598)(108,0.603049)(109,0.579947)(110,0.529484)(111,0.506260)(112,0.421821)(113,0.446971)(114,0.447484)(115,0.403164)(116,0.395474)(117,0.416790)(118,0.437378)(119,0.480327)(120,0.526203)(121,0.579221)(122,0.648683)(123,0.711065)(124,0.784456)(125,0.822340)(126,0.875755)(127,0.933566)(128,0.978488)(129,1.024326)(130,1.054923)(131,1.089251)(132,1.141903)(133,1.170559)(134,1.184199)(135,1.199391)(136,1.203530)(137,1.208079)(138,1.216287)(139,1.216732)(140,1.204690)(141,1.168597)(142,1.151495)(143,1.123085)(144,1.112424)(145,1.074929)(146,1.037710)(147,1.024887)(148,1.012754)(149,1.004337)(150,1.011280)(151,1.018463)(152,1.046840)(153,1.076157)(154,1.114672)(155,1.135391)(156,1.172759)(157,1.182555)(158,1.174590)(159,1.146013)(160,1.098141)(161,1.066544)(162,1.012785)(163,0.959522)(164,0.898601)(165,0.842847)(166,0.777596)(167,0.758966)(168,0.701141)(169,0.670260)(170,0.644017)(171,0.616720)(172,0.591408)(173,0.592066)(174,0.581743)(175,0.607323)(176,0.627332)(177,0.660937)(178,0.682645)(179,0.662125)(180,0.761619)(181,0.774946)(182,0.757452)(183,0.751075)(184,0.711271)(185,0.701860)(186,0.677422)(187,0.641199)(188,0.641545)(189,0.602047)(190,0.608775)(191,0.628390)(192,0.645802)(193,0.692323)(194,0.762059)(195,0.836693)(196,0.899469)(197,0.938796)(198,0.977495)(199,1.022206)(200,1.033014)
      };
      \addplot[green,smooth,line width=0.5pt] plot coordinates{
	      (1,0.601405)(2,0.418529)(3,0.588600)(4,0.464337)(5,0.492596)(6,0.493376)(7,0.441319)(8,0.473208)(9,0.428211)(10,0.427997)(11,0.392522)(12,0.371251)(13,0.385573)(14,0.360869)(15,0.341174)(16,0.362728)(17,0.363397)(18,0.397710)(19,0.406068)(20,0.474303)(21,0.525068)(22,0.592807)(23,0.681061)(24,0.734598)(25,0.816557)(26,0.868966)(27,0.905592)(28,0.944108)(29,0.965029)(30,0.987011)(31,0.987846)(32,0.998214)(33,1.000282)(34,0.999651)(35,0.997295)(36,0.990952)(37,0.990033)(38,0.979059)(39,0.970110)(40,0.957263)(41,0.941770)(42,0.925765)(43,0.907342)(44,0.888574)(45,0.871592)(46,0.856953)(47,0.844533)(48,0.838076)(49,0.839135)(50,0.847141)(51,0.865729)(52,0.897002)(53,0.938076)(54,0.983746)(55,1.027471)(56,1.062974)(57,1.085338)(58,1.098806)(59,1.100636)(60,1.098126)(61,1.090316)(62,1.083701)(63,1.076776)(64,1.068702)(65,1.063910)(66,1.060047)(67,1.060330)(68,1.060842)(69,1.066775)(70,1.075868)(71,1.088407)(72,1.106585)(73,1.125547)(74,1.148399)(75,1.171048)(76,1.194417)(77,1.216524)(78,1.236649)(79,1.253685)(80,1.427311)(81,0.984507)(82,1.584613)(83,1.155182)(84,1.234115)(85,1.283245)(86,1.156323)(87,1.153482)(88,1.042585)(89,1.079602)(90,0.981599)(91,0.953434)(92,0.956201)(93,0.898555)(94,0.902138)(95,0.879987)(96,0.867627)(97,0.864446)(98,0.844814)(99,0.840245)(100,0.824307)(101,0.809767)(102,0.790094)(103,0.761163)(104,0.734328)(105,0.705047)(106,0.667640)(107,0.635379)(108,0.596299)(109,0.565611)(110,0.527893)(111,0.498150)(112,0.361547)(113,0.645728)(114,0.220633)(115,0.494865)(116,0.423482)(117,0.375708)(118,0.456378)(119,0.465280)(120,0.555291)(121,0.558420)(122,0.658836)(123,0.718657)(124,0.756430)(125,0.837232)(126,0.878292)(127,0.932612)(128,0.979756)(129,1.018188)(130,1.064457)(131,1.097081)(132,1.131701)(133,1.158872)(134,1.181336)(135,1.202035)(136,1.212036)(137,1.212844)(138,1.213614)(139,1.204197)(140,1.191732)(141,1.171071)(142,1.150705)(143,1.124884)(144,1.097843)(145,1.071951)(146,1.045849)(147,1.026691)(148,1.011693)(149,1.005213)(150,1.011407)(151,1.025563)(152,1.050117)(153,1.081390)(154,1.114891)(155,1.146570)(156,1.168156)(157,1.175088)(158,1.165363)(159,1.142919)(160,1.105029)(161,1.058821)(162,1.006163)(163,0.952460)(164,0.897190)(165,0.843912)(166,0.793123)(167,0.747046)(168,0.704529)(169,0.668548)(170,0.638702)(171,0.616178)(172,0.600501)(173,0.593969)(174,0.597759)(175,0.608148)(176,0.628157)(177,0.653377)(178,0.682450)(179,0.574713)(180,0.992625)(181,0.494396)(182,0.864323)(183,0.771969)(184,0.676965)(185,0.726594)(186,0.659158)(187,0.683231)(188,0.589615)(189,0.626363)(190,0.619353)(191,0.607642)(192,0.671136)(193,0.704498)(194,0.769788)(195,0.833335)(196,0.884611)(197,0.943163)(198,0.979268)(199,1.013292)(200,1.034219)
      };
      \legend{ {$\hat{u}_t$}, {$\hat{r}_t$}, {Output (optimal $\eta^2$)}, {Output ($\eta^2=10^{-5}$)} }
      \coordinate (spypoint) at (axis cs:80,1.2);
      \coordinate (magnifyglass) at (axis cs:70,0.6);
      \coordinate (spypoint2) at (axis cs:111.5,0.43);
      \coordinate (magnifyglass2) at (axis cs:121,1.3);
      \draw[->,thick] (axis cs:80,1.05) -- (axis cs:80,1.25) node {};
      \draw[->,thick] (axis cs:111,0.25) -- (axis cs:111,0.45) node {};
    \end{axis}
    \spy [gray, size=2.5cm] on (spypoint) in node[fill=white] at (magnifyglass);
    \spy [gray, size=2.5cm] on (spypoint2) in node[fill=white] at (magnifyglass2);
  \end{tikzpicture}
  \caption{Realization of a $1\%$ $\mathcal N(0,.01)$-noisy Mackey-Glass sequence versus network output, $W$ Haar with $\sigma=.9$, $n=400$, $T=\hat{T}=1000$. In magnifying lenses, points of added impulsive noise.}
  \label{fig:robustness_oneshot}
\end{figure}

\medskip

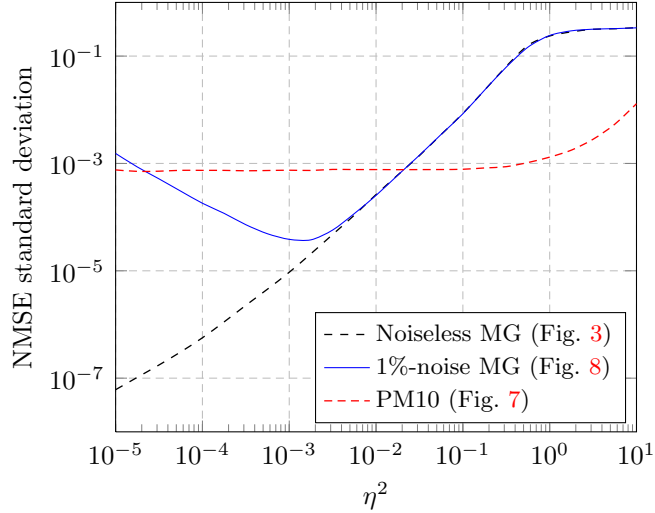
\begin{figure}[h!]
  \centering
  \begin{tikzpicture}[font=\footnotesize]
    \renewcommand{\axisdefaulttryminticks}{4} 
    \tikzstyle{every major grid}+=[style=densely dashed]       
    \tikzstyle{every axis y label}+=[yshift=-10pt] 
    \tikzstyle{every axis x label}+=[yshift=5pt]
    \tikzstyle{every axis legend}+=[cells={anchor=west},fill=white,
        at={(0.98,0.02)}, anchor=south east, font=\scriptsize ]
    \begin{loglogaxis}[
      xmin=1e-5,
      ymin=1e-8,
      xmax=10,
      ymax=1,
      grid=major,
      scaled ticks=true,
      xlabel={$\eta^2$},
      mark options= {solid},
      ylabel={NMSE standard deviation},
      ]
      \addplot[black,dashed,smooth,line width=0.5pt] plot coordinates{
	      (1.000000e-05,6.100363e-08)(1.778279e-05,1.036768e-07)(3.162278e-05,1.761085e-07)(5.623413e-05,3.083740e-07)(1.000000e-04,5.651353e-07)(1.778279e-04,1.117686e-06)(3.162278e-04,2.276309e-06)(5.623413e-04,4.559275e-06)(1.000000e-03,9.396235e-06)(1.778279e-03,2.082187e-05)(3.162278e-03,4.749948e-05)(5.623413e-03,1.088788e-04)(1.000000e-02,2.636316e-04)(1.778279e-02,6.085220e-04)(3.162278e-02,1.401763e-03)(5.623413e-02,3.459957e-03)(1.000000e-01,8.542125e-03)(1.778279e-01,2.318069e-02)(3.162278e-01,6.446564e-02)(5.623413e-01,1.589422e-01)(1.000000e+00,2.370290e-01)(1.778279e+00,2.828967e-01)(3.162278e+00,3.155633e-01)(5.623413e+00,3.233031e-01)(1.000000e+01,3.390413e-01)
};
      \addplot[blue,smooth] plot coordinates{
	      (1.000000e-05,1.534929e-03)(1.778279e-05,8.598815e-04)(3.162278e-05,5.158875e-04)(5.623413e-05,3.046356e-04)(1.000000e-04,1.813197e-04)(1.778279e-04,1.175292e-04)(3.162278e-04,7.224379e-05)(5.623413e-04,4.895190e-05)(1.000000e-03,3.854961e-05)(1.778279e-03,3.774558e-05)(3.162278e-03,5.746786e-05)(5.623413e-03,1.164968e-04)(1.000000e-02,2.540897e-04)(1.778279e-02,6.038796e-04)(3.162278e-02,1.424590e-03)(5.623413e-02,3.478343e-03)(1.000000e-01,8.409451e-03)(1.778279e-01,2.321613e-02)(3.162278e-01,6.234533e-02)(5.623413e-01,1.468013e-01)(1.000000e+00,2.422018e-01)(1.778279e+00,2.956426e-01)(3.162278e+00,3.170252e-01)(5.623413e+00,3.248236e-01)(1.000000e+01,3.362394e-01)
      };
      \addplot[red,densely dashed,smooth,line width=0.5pt] plot coordinates{
	      (1.000000e-05,7.576869e-04)(1.778279e-05,7.114449e-04)(3.162278e-05,7.173025e-04)(5.623413e-05,7.450159e-04)(1.000000e-04,7.431089e-04)(1.778279e-04,7.427746e-04)(3.162278e-04,7.302348e-04)(5.623413e-04,7.389155e-04)(1.000000e-03,7.480975e-04)(1.778279e-03,7.402630e-04)(3.162278e-03,7.741002e-04)(5.623413e-03,7.686716e-04)(1.000000e-02,7.693252e-04)(1.778279e-02,7.656356e-04)(3.162278e-02,7.684807e-04)(5.623413e-02,7.700802e-04)(1.000000e-01,7.793504e-04)(1.778279e-01,8.242613e-04)(3.162278e-01,8.712385e-04)(5.623413e-01,1.040481e-03)(1.000000e+00,1.311493e-03)(1.778279e+00,1.769712e-03)(3.162278e+00,2.822190e-03)(5.623413e+00,5.379646e-03)(1.000000e+01,1.298931e-02)
      };
      \legend{ {Noiseless MG (Fig.~\ref{fig:mackeyglass})}, {$1\%$-noise MG (Fig.~\ref{fig:robustness})}, {PM10 (Fig.~\ref{fig:PM10})} }
    \end{loglogaxis}
  \end{tikzpicture}
  \caption{Standard deviation of testing NMSE for different testbeds (exemplifying the resurgence of noise effect). MG in legend stands for Mackey--Glass. In all scenarios, $n=200$, $T=\hat{T}=400$.}
  \label{fig:noise_resurgence}
\end{figure}

In a second experiment, we shall illustrate the ``noise resurgence'' effect discussed earlier in Remark~\ref{rem:low_noise}. In Figure~\ref{fig:noise_resurgence}, we specifically draw the curves of the testing MSE variances for various experiments conducted earlier in the article. It is observed, as discussed in Remark~\ref{rem:low_noise} that, somewhat counter-intuitively, smaller $\eta^2$ values may lead to increased variances solely due to the in-network noise realization itself (recall that in all our experiments, the connectivity matrix $W$ and the input-output pairs $(u,r)$ and $(\hat u,\hat r)$ are fixed across all Monte Carlo realizations). It is even more interesting to observe here each of the three possible behaviors: a ``natural'' MSE variance decay as $\eta^2\to 0$, a surprising MSE increase, and even an MSE stabilization. Further theoretical analysis to understand those strikingly different behaviors would be appreciable, which would demand more advanced technical considerations.

\medskip

\begin{figure}[h!]
  \centering
  \begin{tikzpicture}[font=\footnotesize]
    \renewcommand{\axisdefaulttryminticks}{4} 
    \tikzstyle{every major grid}+=[style=densely dashed]       
    \tikzstyle{every axis y label}+=[yshift=-10pt] 
    \tikzstyle{every axis x label}+=[yshift=5pt]
    \tikzstyle{every axis legend}+=[cells={anchor=west},fill=white,
        at={(0.98,0.02)}, anchor=south east, font=\scriptsize ]
    \begin{loglogaxis}[
      xmin=1e-4,
      ymin=1e-6,
      xmax=10,
      ymax=10,
      grid=major,
      scaled ticks=true,
      xlabel={$\eta^2$},
      mark options= {solid},
      ylabel={NMSE},
      ]
      \addplot[black,mark=*,smooth,line width=0.5pt] plot coordinates{
(1.000000e-05,7.316997e-06)(1.778279e-05,8.175289e-06)(3.162278e-05,9.764805e-06)(5.623413e-05,1.359670e-05)(1.000000e-04,2.222945e-05)(1.778279e-04,4.045238e-05)(3.162278e-04,7.852946e-05)(5.623413e-04,1.598380e-04)(1.000000e-03,3.353831e-04)(1.778279e-03,7.146774e-04)(3.162278e-03,1.531195e-03)(5.623413e-03,3.274989e-03)(1.000000e-02,6.940455e-03)(1.778279e-02,1.443822e-02)(3.162278e-02,2.899907e-02)(5.623413e-02,5.527345e-02)(1.000000e-01,9.989824e-02)(1.778279e-01,1.760688e-01)(3.162278e-01,3.271895e-01)(5.623413e-01,6.473157e-01)(1.000000e+00,1.150197e+00)(1.778279e+00,1.604699e+00)(3.162278e+00,1.847264e+00)(5.623413e+00,1.941588e+00)(1.000000e+01,1.973622e+00)
      };
      \addplot[black,mark=o,smooth,line width=0.5pt] plot coordinates{
	      (1.000000e-05,6.781638e-07)(1.778279e-05,1.080848e-06)(3.162278e-05,1.802401e-06)(5.623413e-05,3.138874e-06)(1.000000e-04,5.682019e-06)(1.778279e-04,1.065664e-05)(3.162278e-04,2.069542e-05)(5.623413e-04,4.150687e-05)(1.000000e-03,8.565809e-05)(1.778279e-03,1.809023e-04)(3.162278e-03,3.900722e-04)(5.623413e-03,8.562200e-04)(1.000000e-02,1.909435e-03)(1.778279e-02,4.282029e-03)(3.162278e-02,9.450857e-03)(5.623413e-02,2.034997e-02)(1.000000e-01,4.317234e-02)(1.778279e-01,9.539156e-02)(3.162278e-01,2.296726e-01)(5.623413e-01,5.466761e-01)(1.000000e+00,1.069321e+00)(1.778279e+00,1.560296e+00)(3.162278e+00,1.829430e+00)(5.623413e+00,1.935445e+00)(1.000000e+01,1.971623e+00)
      };
      \addplot[black,mark=+,smooth,line width=0.5pt] plot coordinates{
	      (1.000000e-05,1.069007e-06)(1.778279e-05,1.586302e-06)(3.162278e-05,2.449765e-06)(5.623413e-05,3.904976e-06)(1.000000e-04,6.460878e-06)(1.778279e-04,1.098978e-05)(3.162278e-04,1.952638e-05)(5.623413e-04,3.576106e-05)(1.000000e-03,6.781893e-05)(1.778279e-03,1.384535e-04)(3.162278e-03,2.814102e-04)(5.623413e-03,5.560819e-04)(1.000000e-02,1.162927e-03)(1.778279e-02,2.461466e-03)(3.162278e-02,4.891914e-03)(5.623413e-02,1.022856e-02)(1.000000e-01,2.534190e-02)(1.778279e-01,7.099134e-02)(3.162278e-01,2.021319e-01)(5.623413e-01,5.197455e-01)(1.000000e+00,1.048399e+00)(1.778279e+00,1.548955e+00)(3.162278e+00,1.824888e+00)(5.623413e+00,1.933881e+00)(1.000000e+01,1.971115e+00)
      };
      \addplot[red,line width=1pt,error bars/.cd,y dir=both,y explicit, error bar style={mark size=2.5pt}] plot coordinates{
	      (1.000000e-05,2.158324e-06)(1.778279e-05,2.222249e-06)(3.162278e-05,2.699693e-06)(5.623413e-05,3.821587e-06)(1.000000e-04,5.967865e-06)(1.778279e-04,9.952312e-06)(3.162278e-04,1.756008e-05)(5.623413e-04,3.244042e-05)(1.000000e-03,6.228394e-05)(1.778279e-03,1.235916e-04)(3.162278e-03,2.526804e-04)(5.623413e-03,5.307038e-04)(1.000000e-02,1.144492e-03)(1.778279e-02,2.541588e-03)(3.162278e-02,5.780350e-03)(5.623413e-02,1.348636e-02)(1.000000e-01,3.264252e-02)(1.778279e-01,8.305261e-02)(3.162278e-01,2.192235e-01)(5.623413e-01,5.421906e-01)(1.000000e+00,1.071025e+00)(1.778279e+00,1.563089e+00)(3.162278e+00,1.830874e+00)(5.623413e+00,1.935982e+00)(1.000000e+01,1.971802e+00)
      };
      \legend{{Haar $W$, $\sigma=.99$}, {Haar $W$, $\sigma=.9$}, {Haar $W$, $\sigma=.5$},  {Multimemory $W$} }
    \end{loglogaxis}
  \end{tikzpicture}
  \caption{Testing (normalized) MSE for the Mackey Glass one-step ahead task, $W$ (multimemory) versus $W_1^+=.99 Z_1^+$, $W_2^+=.9 Z_2^+$, $W_3^+=.5 Z_3^+$ (with $Z_i^+$ Haar distributed) all defined as in Figure~\ref{fig:multimemory}, $n=400$, $T=\hat{T}=800$.}
  \label{fig:mackeyglass_multimemory_comp}
\end{figure}
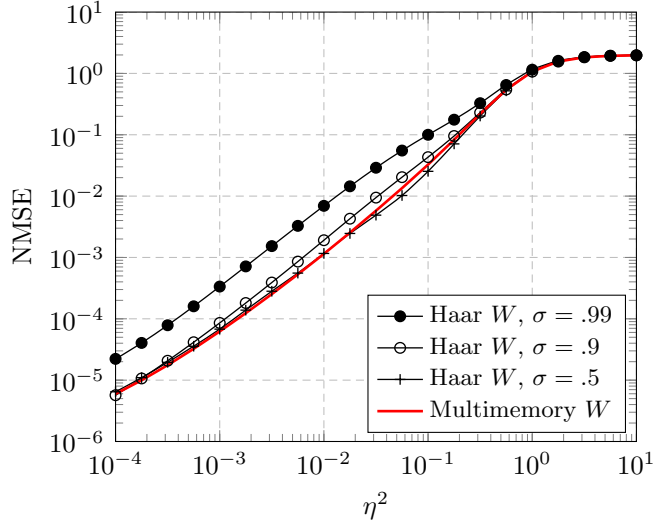

We complete this section by a last comparative experiment of the performance of the multi-memory matrix $W$ defined in Remark~\ref{rem:multimemory} specialized to the setting of Figure~\ref{fig:multimemory} (that is, with three rates $\sigma_1=.99$, $\sigma_2=.9$, and $\sigma_3=.5$) versus Haar matrices for the different $\sigma_i$ values, for the Mackey-Glass model. This is depicted in Figure~\ref{fig:mackeyglass_multimemory_comp}, which shows a valuable performance gain versus ill-chosen individual hypotheses of $\sigma$ and a rather fair match to the best individual $\sigma$ value.

\section{Concluding Remarks}
\label{sec:conclusion}

One of the main outcomes of the present study is a better understanding of the ESN instability to low internal noise variance described by Jaeger in \cite{JAE01b}. We made it clear here that, when the noise variance is sufficiently large compared to the inverse square root of the network size, the ESN tends to have a deterministic behavior (that is, independent of the noise realization) as both time and network size grow large. This deterministic behavior was characterized here through new results from random matrix theory, with the main consequences to ESN's being encapsulated in Corollary~\ref{cor:MSEtrain} and Corollary~\ref{cor:MSEtest}. When the noise variance is however too small, random matrix theory cannot guarantee in general the aforementioned deterministic network behavior in the large system asymptotic. Although difficult to read, the asymptotic performances revolve around a critical matrix that contains the exponential memory decay information and may be use to generalize Ganguli's notion of memory curve (see Remark~\ref{rem:MC}). This generalized memory curve draws improved conclusions on the ESN performance (with sometimes opposite outcomes as compared to the conclusions drawn upon the former memory curve notion).

In the particular case of some standard random matrix models for the neural connectivity matrix, we further simplified the rather involved generic expressions from Corollary~\ref{cor:MSEtrain} and Corollary~\ref{cor:MSEtest}. Of particular interest is the case of bi-orthogonally invariant random connectivity matrices for which the mean square error performances of learning and testing take on the explicit expressions of Corollary~\ref{cor:unitarily_invariant} or Corollary~\ref{cor:unitarily_invariant_c>1} from which much can be inferred. Among other results, we understood the importance of random input weights for the network performance as compared to input weights that match the leading eigenvectors of the connectivity matrix and we made it clear that the ESN testing performance is asymptotically optimal for arbitrary low noise variances {\it when} the task to fulfill is a mere linear combination of the last few past inputs. In additional experiments, we also understood the role of a non-trivial noise level as a robustness-to-outliers enhancer.

\bigskip

Beyond their theoretical value, note also that the results of this article may be used in practice to anticipate the behavior of ESN's on real-life datasets, thereby saving one from the painstaking task of running long Monte Carlo simulations. For instance, one may consider retrieving the theoretical MSE outputs corresponding to successive sequences of training and testing inputs so to better tune the ESN parameters. This is all the more precious that the network size and time windows are large since then the formulas of, say Corollary~\ref{cor:unitarily_invariant}, can be retrieved extremely fast.

\bigskip

One frustrating aspect of the work nonetheless remains that, for low noise variances (typically of practical interest), our analysis leads to large mismatches when the network size is kept moderate. This is observed in Figure~\ref{fig:mackeyglass} in particular. There is as such no theoretical control of this regime. This being said, in some scenarios where the limiting singularity at zero noise can be avoided, we showed an accurate fit of our theoretical findings at all noise levels. But the main limitation of the analysis so far lies in its dealing with linear activation functions only. We believe that the (much more interesting) question of non-linear activation functions may be addressed by the exploitation of results from the mean field theory which prefigures an asymptotic joint Gaussian behavior of the state vectors at each time instant, which may allow for an adaptation of the present random matrix analysis; alternatively, the Gaussian tools presented here may be adapted to Taylor expansions of the activation functions. This investigation is left to future work.

\appendix

\section{Proof of Theorem~\ref{th:deteq1}}
\label{app:Th1}

The present and next sections are dedicated to the proofs of the main results Theorems~\ref{th:deteq1}--\ref{th:deteq2} of the article. The proofs rely on now well-established tools from random matrix theory, with an additional specificity due to the ``infinitely long'' time dependence between the columns of the random matrices involved; however, as the time dependence is {\it effectively} short (of order $o(T^{\alpha})$ for any $\alpha>0$), these matrices can be handled as if dependence was among only a few next and previous columns. We shall not deeply elaborate on all technical arguments for the sake of readability and concision. The reader more interested in the proof techniques and in more advanced time dependence considerations may refer to \cite{PAS11,HAC06} on the Gaussian methods and \cite{BAN13} for stationary processes in random matrix theory.

\medskip

In the present section, $W$ is considered a deterministic matrix with operator norm less than unity. We recall that $X=\{x_j\}_{j=0}^{T-1}\in\RR^{n\times T}$, for the infinite time series $\ldots,x_{-1},x_0,x_1,\ldots\in\RR^n$, defined recursively through
\begin{align*}
	x_{t+1} &= Wx_t + mu_{t+1} + \eta\varepsilon_{t+1}
\end{align*}
with $m$ of bounded norm and $\ldots,u_{-1},u_0,u_1,\ldots\in\RR$ some time series. We additionally denote $A=MU$ where $M=\{W^jm\}_{j=0}^{T-1}$ and $U=T^{-\frac12}\{u_{j-i}\}_{i,j=0}^{T-1}$. Also, let $Z=\eta T^{-\frac12} \{\sum_{k\geq 0}W^k \varepsilon_{j-k}\}_{j=0}^{T-1}$ the concatenated noise vectors, with $\varepsilon_i\sim\mathcal N(0,I_n)$. With these notations and normalization, we have $X=\sqrt{T}(A+Z)$, where $A$ and $Z$ are expected to have operator norm of order $O(1)$ with respect to $n,T\to\infty$ as per Assumption~\ref{ass:c} and thus so should $\frac1TXX^\trans$.

For $\gamma>0$, denoting $Q_\gamma=(\frac1TXX^\trans+\gamma I_n)^{-1}$, our objective is to obtain an approximation of $Q_\gamma$ in the sense of the equivalence $\leftrightarrow$ using the so-called {\it Gaussian method} introduced by Pastur in \cite{PAS11}. This method consists in two ingredients: (i) an integration by parts formula for Gaussian random variables (also called Stein's lemma) that stipulates that, for $x\sim \mathcal N(0,1)$ and a polynomially bounded differentiable $f$, $\EE[xf(x)]=\EE[f'(x)]$, and (ii) concentration inequalities or moment based bounds (such as the Nash--Poincar\'e inequality) to control small terms. The idea here is to expand terms of the type $\EE[ [\varepsilon_i]_j[Q_\gamma]_{kl}]$ using the Gaussian integration by parts formula in order to retrieve an implicit {\it but deterministic} expression for $Q_\gamma$, up to small random terms. Then, thanks to concentration or moment bounds, the aforementioned small terms are shown to vanish at a sufficient speed to ensure almost sure convergence of $Q_\gamma$ to the deterministic solution of the implicit equation in the sense of the equivalence $\leftrightarrow$.

\medskip

We start by noticing that $Q_\gamma = \frac1\gamma I_n - \frac1\gamma \frac1TXX^\trans Q_\gamma$, a relation often referred to as the {\it resolvent identity}. This allows one to write $\EE[Q_\gamma]$ as a function of $\EE[XX^\trans Q_\gamma]$ which lends itself to the integration by parts approach since $X$ is a linear function of the Gaussian variables $[\varepsilon_i]_j$. 

In what follows, for readability, we shall denote $Q=Q_\gamma$ (and thus $Q_{ij}=[Q_\gamma]_{ij}$) and $\varepsilon_{ij}=[\varepsilon_i]_j$. Then we have
\begin{align}
	\label{eq:terms_1-4}
	\EE[Q_{ij}] &= \frac1\gamma {\bm\delta}_{ij} - \frac1\gamma \Big( \underbrace{\EE[ [ZZ^\trans Q]_{ij} ]}_{(I)} + \underbrace{\EE[[ZA^\trans Q]_{ij}]}_{(II)} + \underbrace{\EE[ [AZ^\trans Q]_{ij}]}_{(III)} + \underbrace{\EE[ [AA^\trans Q]_{ij} ]}_{(IV)}  \Big).
\end{align}
Each of the four braced terms needs be treated independently. Note first that term $(IV)$ is simply $\sum_k [AA^\trans]_{ik} \EE[Q_{kj}]$ and is thus treated similar to $\EE[Q_{ij}]$ itself. It then remains to handle terms $(I)$--$(III)$. Before handling each term, let us first introduce a few elementary results of constant use in what follows. First, by a mere development, we have
\begin{align*}
	Z_{ab} &= \frac{\eta}{\sqrt{T}} \sum_{k\geq 0} \sum_{p=1}^n [W^k]_{ap} \varepsilon_{p,b-k}
\end{align*}
from which
\begin{align}
	\label{eq:partial_Z}
	\frac{\partial Z_{ab}}{\partial \varepsilon_{il}} &= \frac{\eta}{\sqrt{T}} \sum_{k\geq 0} \sum_{p=1}^n {\bm\delta}_{pi}{\bm\delta}_{l,b-k} [W^k]_{ap}.
\end{align}
Expanding $X$ in the expression of $Q$ and using $\partial Q=-Q(\partial Q^{-1})Q$, we then find
\begin{align}
	\label{eq:partial_Q}
	\frac{\partial Q_{mj}}{\partial \varepsilon_{il}} &= -\frac{\eta}{\sqrt{T}} \sum_{p=1}^n  {\bm\delta}_{l\leq p} \left(\left[ Q(Z+A) \right]_{mp} \left[ (W^{p-l})^\trans Q \right]_{ij} + \left[ (Z+A)^\trans Q \right]_{pj} \left[ Q W^{p-l} \right]_{mi}\right).
\end{align}
It is important at this point to bring some insight from random matrix theory. If $\varepsilon_{il}$ were a {\it complex} rather than real standard Gaussian random variable, the second term in the right-hand side parenthesis would not have appeared. Since first order deterministic equivalents (which is what we are proceeding to here) are usually valid irrespective of the i.i.d.\@ distribution (real or complex) of the $\varepsilon_{il}$'s, it is expected that this second term will lead to vanishing terms in what follows. 

With these preliminary results and this remark in mind, we can tackle the calculus of terms $(I)$--$(III)$ from \eqref{eq:terms_1-4}. Let us first focus on term $(I)$. Developing $\EE[ [ZZ^\trans Q]_{ij}]$ as a function of the $\varepsilon_{kl}$'s and applying the Gaussian integration-by-parts formula, we find
\begin{align*}
	\EE[ [ZZ^\trans Q]_{ij}] &= \eta \sum_{l=1}^T\sum_{m=1}^n\sum_{o=1}^n \sum_{k\geq 0} \EE \left[ \varepsilon_{o,l-k} Z_{m,l} Q_{mj} \right] \left[W^k\right]_{io} \\
	&= \eta \sum_{l=1}^T\sum_{m=1}^n\sum_{o=1}^n \sum_{k\geq 0} \left( \EE \left[ \frac{\partial Z_{m,k}}{\partial \varepsilon_{o,l-k}} Q_{mj}\right] + \EE\left[ \frac{\partial Q_{mj}}{\partial \varepsilon_{o,l-k}} Z_{ml} \right] \right) \left[W^k\right]_{io}.
\end{align*}
Substituting the derivatives by the forms \eqref{eq:partial_Z} and \eqref{eq:partial_Q}, we obtain after full development and simplifications
\begin{align*}
	\EE[ [ZZ^\trans Q]_{ij} ] &= \eta^2 \sum_{k\geq 0} \EE \left[ \left[ W^k (W^k)^\trans Q \right]_{ij} \right] \nonumber \\
	&-\eta^2 \sum_{k\geq 0} \sum_{q=-k}^{T-1} \sum_{l=1}^{T-q^+} \EE \left[ \left[ W^k (W^{k+q})^\trans Q \right]_{ij}  \frac1T[Z^\trans (Z+A)\tilde{Q}]_{l,q+l}  \right] + \EE[\zeta^{[1]}_{ij}]
\end{align*}
where we defined $\tilde{Q}=\tilde{Q}_\gamma=(\frac1TX^\trans X + \gamma I_n)^{-1}$ and where the term $\zeta^{[1]}_{ij}$ arises from the development of the aforementioned second term in the parentheses of \eqref{eq:partial_Q} and can be shown to satisfy $\zeta^{[1]} \leftrightarrow 0$. Similarly, addressing the term $(II)$ in \eqref{eq:terms_1-4}, we find
\begin{align*}
	\EE[ [ZA^\trans Q]_{ij} ] &= - \eta^2 \sum_{k\geq 0} \sum_{q=-k}^{T-1} \sum_{l=1}^{T-q^+} \EE \left[ \frac1T \left[ A^\trans (Z+A)\tilde{Q} \right]_{l,q+l} \left[ W^k (W^{k+q})^\trans Q \right]_{ij}\right] + \EE[\zeta^{[2]}_{ij}]
\end{align*}
where again we can show that $\zeta^{[2]} \leftrightarrow 0$. Summing the approximations for $(I)$ and $(II)$, from the resolvent identity $(Z+A)^\trans (Z+A)\tilde{Q}=I_n -\gamma \tilde{Q}$, we find
\begin{align*}
	\EE[ [Z(Z+A)^\trans Q]_{ij} ] &= \eta^2 \sum_{k\geq 0} \EE \left[ \left[ W^k (W^k)^\trans Q \right]_{ij} \right] \nonumber \\
	&- \eta^2 \sum_{k\geq 0} \sum_{q=-k}^{T-1} \sum_{l=1}^{T-q^+} \EE \left[ \left[ W^k (W^{k+q})^\trans Q \right]_{ij}  \frac1T[ I_n - \gamma \tilde{Q}]_{l,q+l}  \right] + \EE[\zeta^{[1]}_{ij}+\zeta^{[2]}_{ij}].
\end{align*}
Since $[I_n]_{l,q+l}={\bm\delta}_{q=0}$, the first right-hand side term cancels with the part of the second term involved with matrix $\frac1TI_n$, and we find
\begin{align}
	\label{eq:I+II}
	\EE[ [Z(Z+A)^\trans Q]_{ij} ] &= \eta^2\gamma \sum_{k\geq 0} \sum_{q=-k}^{T-1} \sum_{l=1}^{T-q^+} \EE \left[ \left[ W^k (W^{k+q})^\trans Q \right]_{ij}  \frac1T\tilde{Q}_{l,q+l} \right] + \EE[\zeta^{[1]}_{ij}+\zeta^{[2]}_{ij}].
\end{align}
Moving to term $(III)$ in \eqref{eq:terms_1-4}, since $A$ is deterministic, we first find the interesting expression
\begin{align}
	\label{eq:IIIa}
	\EE[ [Z^\trans Q]_{ij} ] &= -\eta^2 \sum_{k\geq 0} \sum_{q=-k}^{T-1} \sum_{l=1}^{T-q^+} \EE \left[ \frac1T \tr ( W^{k} (W^{k+q})^\trans Q ) [(Z+A)^\trans Q]_{q+i,j}\right] + \EE [\zeta^{[3]}_{ij}]
\end{align}
with $\zeta^{[3]} \leftrightarrow 0$ from which immediately we get
\begin{align*}
	\EE[ [AZ^\trans Q]_{ij} ] &= -\eta^2 \sum_{k\geq 0} \sum_{q=-k}^{T-1} \sum_{l=1}^{T-q^+} \EE \left[ \frac1T \tr ( W^{k} (W^{k+q})^\trans Q ) A_{il}[(Z+A)^\trans Q]_{q+l,j}\right] + \EE [ [A\zeta^{[3]}]_{ij}]
\end{align*}
and we of course still have $A\zeta^{[3]}\leftrightarrow 0$.

We must discuss at this point the next key idea of the Gaussian method. In term $(III)$, the right-hand side expectation is taken over the product of the trace $\frac1T \tr ( W^{k} (W^{k+q})^\trans Q )$ and of the quantity $A_{il} [(Z+A)^\trans Q]_{q+l,j}$. Writing $\frac1T \tr ( W^{k} (W^{k+q})^\trans Q )=\EE[\frac1T \tr ( W^{k} (W^{k+q})^\trans Q )]+(\frac1T \tr ( W^{k} (W^{k+q})^\trans Q )-\EE[\frac1T \tr ( W^{k} (W^{k+q})^\trans Q )])$, it can be shown, using Cauchy--Schwarz and the Nash--Poicar\'e inequalities \cite{PAS11}, along with the Borel--Cantelli lemma \cite{BIL08}, that 
\begin{align*}
	\sum_{k\geq 0} \sum_{q=-k}^{T-1} \sum_{l=1}^{T-q^+} \left(\frac1T \tr ( W^{k} (W^{k+q})^\trans Q )-\EE\left[\frac1T \tr ( W^{k} (W^{k+q})^\trans Q )\right]\right)A_{il} [(Z+A)^\trans Q]_{q+l,j} &\leftrightarrow 0
\end{align*}
which unfolds from $\frac1T \tr ( W^{k} (W^{k+q})^\trans Q )$ concentrating around its mean in the large $n,T$ regime, a standard result of random matrix theory. The main non-classical difficulty in showing this result lies here in the fact that the summation over up to $T$ values of the dummy variable $q$ involves both terms in and outside the bracket. Nonetheless, since $\rho(W)<1$, $\|W^q\|$ vanishes at exponential speed and thus only $O(\log(T))$ values of $q$ are effectively playing a role. The aforementioned Nash--Poicar\'e inequality argument ensures a control of the residual terms with a $O(1/T^2)$ variance for each $q$-summand which can then be summed over the non-trivial values of $q$ to bring a total variance bounded by $O(\log(T)/T^2)$, which is summable, and then allows for Borel--Cantelli to be applied. 

The same reasoning applies to the main expectation in the expression of $(I)+(II)$, where here the term that concentrates around its mean is $\frac1T\sum_{l=1}^{T-q^+}\tilde{Q}_{l,q+l}$, which is more easily seen as $\frac1T\tr (J^q\tilde{Q})$.

The relation \eqref{eq:IIIa} in itself is quite instructive. Indeed, with the previous remark on the concentration of $\frac1T \tr ( W^{k} (W^{k+q})^\trans Q )$, we may break the right-hand expectation as well as the term $(Z+A)^\trans Q$ into $Z^\trans Q+A^\trans Q$ to retrieve a connection between left- and right-hand sides. Precisely, we find that
\begin{align*}
	& \left[\left( I_T + \eta^2 \sum_{k\geq 0} \sum_{q=-k}^{T-1} \EE \left[\frac1T\tr \left( W^kW^{k+q})^\trans Q \right)\right] J^q \right) \EE\left[ X^\trans Q \right] \right]_{ij} \nonumber \\
	&= - \eta^2 \sum_{k\geq 0} \sum_{q=-k}^{T-1} \EE\left[\frac1T \tr ( W^{k} (W^{k+q})^\trans Q ) \right] \EE\left[ [ J^q A^\trans Q]_{i,j} \right] + o(1)
\end{align*}
where we used $[B]_{q+i,j}=[J^q B]_{i,j}$. 
Remark now that 
\begin{align*}
	\sum_{k\geq 0}\sum_{q=-k}^{T-1} \frac1T\tr \left( W^k(W^{k+q})^\trans Q \right)J^q &= \sum_{k\geq 0} \left\{ \frac1T\tr \left( W^{k+(b-a)^+}(W^{k+(a-b)^+})^\trans Q \right) \right\}_{a,b=1}^T \\
	&= \left\{ \frac1T\tr \left( S_{a-b} Q \right) \right\}_{a,b=1}^T.
\end{align*}
Denoting $\bar{R}= \EE[\{ \frac1T\tr ( S_{a-b} Q) \}_{a,b=1}^T ] $ and using concentration arguments (Nash--Poincar\'e inequality in particular) entails 
\begin{align}
	\label{eq:ZQ}
	Z^\trans Q &\leftrightarrow -\eta^2 \left( I_T + \eta^2 \bar{R} \right)^{-1}\bar{R} A^\trans Q.
\end{align}
From the definition of the equivalence relation $\leftrightarrow$, this entails 
\begin{align}
	\label{eq:AZQ}
	AZ^\trans Q &\leftrightarrow -\eta^2 A\left( I_T + \eta^2 \bar{R} \right)^{-1}\bar{R} A^\trans Q.
\end{align}
Similarly, recalling \eqref{eq:I+II}, we have
\begin{align*}
	Z(Z+A)^\trans Q &\leftrightarrow \eta^2 \gamma \sum_{k\geq 0} \sum_{q=-k}^{T-1} \frac1T\tr (J^q \tilde{Q}) W^k(W^{k+q})^\trans Q \\
	&= \eta^2 \gamma \sum_{q=-\infty}^\infty \frac1T\tr (J^q \tilde{Q}) S_q Q.
\end{align*}
We may then define $\bar{\tilde{R}}=\sum_{q=-\infty}^\infty \EE [\frac1T\tr (J^q \tilde{Q}) S_q]$. Added to \eqref{eq:AZQ} and $AA^\trans Q$, this is
\begin{align*}
	(Z+A)(Z+A)^\trans Q &\leftrightarrow -\eta^2 \gamma \bar{\tilde{R}} - \eta^2 A(I_T+\eta^2\bar{R})^{-1}\bar{R}A^\trans Q + AA^\trans Q.
\end{align*}
With $AA^\trans=A(I_T+\eta^2\bar{R})^{-1}(I_T+\eta\bar{R})A^\trans$ and $(Z+A)(Z+A)^\trans Q=I_n-\gamma Q$, this further reads
\begin{align*}
	Q &\leftrightarrow \frac1\gamma I_n - \eta^2 \bar{\tilde{R}}Q - \frac1\gamma A(I_T+\eta^2\bar{R})^{-1}A^\trans Q.
\end{align*}
which, after gathering the factors of $Q$ together, finally gives the first identity
\begin{align}
	\label{eq:Q_bR_btR}
	Q &\leftrightarrow \frac1\gamma \left( I_n + \eta^2 \bar{\tilde R} + \frac1\gamma A(I_T+\eta^2 \bar{R})^{-1}A^\trans \right)^{-1}.
\end{align}

To pursue our investigation, we need to proceed to the same development for the matrix $\tilde{Q}$ which appears in the definition of $\bar{\tilde R}$. The idea is to express $\tilde{Q}$ under a form involving $Q$ itself, then closing the loop. The analysis is extremely similar to that of $Q$ and it is not surprising (from the symmetry between $Q$ and $\tilde{Q}$) to finally obtain
\begin{align}
	\label{eq:tQ_bR_btR}
	\tilde{Q} &\leftrightarrow \frac1\gamma \left( I_T + \eta^2 \bar{R} + \frac1\gamma A^\trans (I_n+\eta^2 \bar{\tilde R})^{-1}A^\trans \right)^{-1}.
\end{align}

At this point, however, both $\bar{R}$ and $\tilde{\bar R}$ are non explicit quantities that depend on the statistics of $Q$ and $\tilde{Q}$. From \eqref{eq:Q_bR_btR}, we get that, for each $a,b$,
\begin{align*}
	\frac1T\tr (S_{a-b}Q) &\leftrightarrow \frac1\gamma \frac1T\tr S_{a-b} \left( I_n + \eta^2 \bar{\tilde R} + \frac1\gamma A(I_T+\eta^2 \bar{R})^{-1}A^\trans \right)^{-1}
\end{align*}
and this relation is shown to be uniform across $a,b$, as it involves only $O(\log(T))$ non-trivial coefficients. To freely identify $\bar{R}$ with $\{  \frac1{\gamma T}\tr S_{a-b} ( I_n + \eta^2 \bar{\tilde R} + \frac1\gamma A(I_T+\eta^2 \bar{R})^{-1}A^\trans)^{-1}\}_{a,b=1}^T$, one may ensure that the difference between both matrices vanishes in spectral norm almost surely (here the relation $\leftrightarrow$ may not be enough).\footnote{A typical counter-example is the case of $Z\in\RR^{n\times T}$ with i.i.d.\@ zero mean and unit variance entries for which $[\frac1TZZ^\trans]_{ab}\to {\bm \delta}_{a-b}$ uniformly over $a,b$ while clearly $(\frac1TZZ^\trans + \gamma I_n)^{-1} \not\leftrightarrow (1+\gamma)^{-1}I_T$.} Here the result holds true because both $\{\frac1T\tr (S_{a-b}Q)\}_{a,b=1}^T$ and $\bar{R}$ are Toeplitz matrices with exponentially decaying coefficients away from the main diagonal. Hence, we may essentially see each matrix as the sum of a circulant matrix and of a matrix with $O(\log(T))$ non-vanishing upper-right and lower-left entries (see \cite{GRA06} for such a construction). Circulant matrices being diagonalizable in the Fourier basis with eigenvalues equal to the Fourier transform of the concatenated first column and row, that the difference in spectral norm vanishes boils down to the convergence of the difference between these Fourier transforms, which is easily obtained through the joint entry convergence and exponential decrease. As for the remaining corner entries, being of $\log(T)$ number, we deal here with the difference in spectral norm of small rank matrices, which is obtained by direct uniform convergence. As such, generally speaking, if the entries of a Toeplitz matrix with exponentially vanishing profile converge jointly to given limits, then the limiting Toeplitz matrix is equivalent in the spectral norm sense.

Similarly, to identify $\bar{\tilde R}$ with $\sum_q \frac1{\gamma T}\tr (J^q ( I_T + \eta^2 \bar{R} + \frac1\gamma A^\trans (I_n+\eta^2 \bar{\tilde R})^{-1}A^\trans )^{-1}) S_q$, we need to show the spectral norm difference of these matrices vanishes almost surely. This is here obtained from the uniform convergence across the $O(\log(T))$ first trace coefficients (say for all $|q|\leq C\log(T)$) and from the corresponding exponentially vanishing spectral norm of $S_q$.

All said, we may then define $R_\gamma$, $\bar{R}_\gamma$, $\bar{Q}_\gamma$, and $\bar{\tilde Q}_\gamma$ as in Theorem~\ref{th:deteq1} and the results above ensure that $Q_\gamma\leftrightarrow \bar{Q}_\gamma$ and $\tilde{Q}_\gamma\leftrightarrow \bar{\tilde Q}_\gamma$. 

\begin{remark}[Result without washout period]
	Theorem~\ref{th:deteq1} assumes an infinite noise time series $(\ldots,\varepsilon_{-1},\varepsilon_0,\varepsilon_1,\ldots)$. One might have alternatively considered a scenario without washout period, that is, with $x_{-1}=0$ and first time instant being $t=0$. In this case, Theorem~\ref{th:deteq1} remains valid but for the following updated expressions of $R_\gamma$ and $\tilde{R}_\gamma$
	\begin{align*}
		R_\gamma &= \left\{  \sum_{k=0}^{\max(i,j)-1} \frac1T\tr W^{k+(j-i)^+} (W^{k+(i-j)^+})^\trans \bar{Q}_\gamma \right\}_{i,j=1}^T \\
		\tilde R_\gamma &= \sum_{q=-(T-1)}^{T-1}  \frac1T\tr \left( J^q\bar{\tilde Q}_\gamma \right) \sum_{k=0}^{T-1-|q|}W^{k+(-q)^+} (W^{k+q^+})^\trans.
	\end{align*}
	In particular, $R_\gamma$ is no longer Toeplitz. Nonetheless the non-Toeplitz behavior is essentially concentrated in the top-left corner of size $O(\log(T))$ since the remainder of the matrix behaves essentially as Toeplitz (for $i,j\geq C\log(T)$ for some large enough constant $C$). This modification may alter the behavior of the associated train and test MSE, especially if $r$ and $\hat{r}$ concentrate their energy in their first entries.
\end{remark}

\section{Proof of Theorem~\ref{th:deteq2}}
\label{app:Th2}

The first part of Theorem~\ref{th:deteq2} is directly obtained from \eqref{eq:ZQ} along with $Q_\gamma\leftrightarrow \bar{Q}_\gamma$. Indeed, from these relations, we have
\begin{align*}
	Q_\gamma \frac1{\sqrt{T}} X = Q_\gamma Z + Q_\gamma A &\leftrightarrow -\eta^2 \bar{Q}_\gamma A R_\gamma (I_T+\eta^2 R_\gamma)^{-1} + \bar{Q}_\gamma A \\
	&= \bar{Q}_\gamma A (I_T+\eta^2 R_\gamma)^{-1}. 
\end{align*}

The proof of the second part of Theorem~\ref{th:deteq2} is not as straightforward as it involves twice the matrix $Q_\gamma$ and thus results from Theorem~\ref{th:deteq1} cannot be immediately applied. To handle this term, first write
\begin{align}
	\label{eq:I-IV,2}
	\frac1TX^\trans Q_\gamma B Q_\gamma X &= \underbrace{Z^\trans Q_\gamma B Q_\gamma Z}_{(I)} + \underbrace{Z^\trans Q_\gamma B Q_\gamma A}_{(II)} + \underbrace{A^\trans Q_\gamma B Q_\gamma Z}_{(III)} + \underbrace{A^\trans Q_\gamma B Q_\gamma A}_{(IV)}.
\end{align}
Since $B$ is assumed symmetric, $(III)$ is the transposed version of $(II)$, so that only one of the two needs be studied.

Similar to Appendix~\ref{app:Th1}, we shall from now on simply write $Q_\gamma$ as $Q$, $\tilde{Q}_\gamma$ as $\tilde{Q}$, etc.

We start by addressing term $(I)$. We use again the Gaussian tools centered around the Gaussian integration by parts formula. We shall also benefit from the results of Theorem~\ref{th:deteq1}. Since $B$ is deterministic, it needs not be included early in calculations so we merely start by evaluating, for given indices $i,j,k,l$,
\begin{align*}
	\EE \left[ [Z^\trans Q]_{ij} [QZ]_{kl} \right] &= \sum_{m,m',p,p'=1}^n \sum_{q,q'\geq 0} \eta^2 \EE \left[ \varepsilon_{p,i-q} \varepsilon_{p',l-q'} Q_{mj} Q_{km'} \right] [W^q]_{mp} [W^{q'}]_{m'p'} \\
	&=  \sum_{m,m',p,p'=1}^n \sum_{q,q'\geq 0} \eta^2 \EE \left[ \frac{\partial (\varepsilon_{p,i-q} Q_{mj}Q_{km'})}{\partial \varepsilon_{p',l-q'}} \right] [W^q]_{mp} [W^{q'}]_{m'p'}
\end{align*}
where the second line follows from the Gaussian integration-by-parts formula. Developing the derivative based on \eqref{eq:partial_Q} and on the fact that $\partial \varepsilon_{ab}/\partial \varepsilon_{cd}={\bm\delta}_{ac}{\bm\delta}_{bd}$, we get after simplification
\begin{align}
	\label{eq:ZQQZ}
	\EE \left[ [Z^\trans Q]_{ij} [QZ]_{kl} \right] &= \eta^2 \sum_{q,q'\geq 0} \EE\left[ \frac1T [QW^q(W^{q'})^\trans Q]_{jk} {\bm\delta}_{i-q,l-q'}\right] \nonumber \\ 
	&- \eta^3 \sum_{q,q'\geq 0} \sum_{s=l-q'}^T \EE \left[ [ \frac1{\sqrt{T}}\varepsilon^\trans (W^q)^\trans Q(Z+A)]_{i-q,s}\frac1T [QW^{q'}(W^{q'+s-l})^\trans Q]_{kj} \right] \nonumber \\
	&- \eta^3 \sum_{q,q'\geq 0} \sum_{s=l-q'}^T \EE\left[ [ \frac1{\sqrt{T}}\varepsilon^\trans (W^q)^\trans Q]_{i-q,j} [Q(Z+A)]_{k,s} \frac1T \tr (W^{q'} (W^{q'+s-l})^\trans Q)\right] \nonumber \\
	&+ \EE[\zeta^{[1]}_{ijkl}]
\end{align}
for some $\zeta^{[1]}_{ijkl}\leftrightarrow 0$ (arising from terms consistent with the remark following \eqref{eq:partial_Q} in Appendix~\ref{app:Th1}) and where $\varepsilon=\{\varepsilon_{ij}\}_{ij=1}^{n,T}$. Inserting $B_{jk}$, summing over $j$ and $k$, we obtain after simplifications
\begin{align*}
	\EE \left[ [Z^\trans QBQZ]_{il} \right] &= \eta^2 \bar{G}_{il} -\eta^2 \EE\left[ [Z^\trans Q(Z+A)\bar{G}]_{il} \right] - \eta^2 \EE\left[ [Z^\trans QBQ(Z+A)\bar{R}]_{il} \right] + o(1)
\end{align*}
where $\bar{R}$ was introduced in Appendix~\ref{app:Th1} and we defined $\bar{G}$ the matrix with
\begin{align*}
	\bar{G}_{ij} &= \sum_{k\geq 0} \EE \left[ \frac1T\tr \left( B Q W^{k+(j-i)^+} (W^{k+(i-j)^+})^\trans Q \right)\right].
\end{align*}
Gathering the terms in $Z^\trans QBQZ$ together along with concentration arguments, we finally obtain
\begin{align*}
	Z^\trans QBQZ &\leftrightarrow \eta^2 \bar{G}(I_T+\eta^2\bar{R})^{-1}-\eta^2 Z^\trans Q (Z+A) \bar{G}(I_T+\eta^2\bar{R})^{-1}-\eta^2 Z^\trans QBQA\bar{R}(I_T+\eta^2\bar{R})^{-1}.
\end{align*}
In the right-hand side formulation, the second term can be approximated from the results of Theorem~\ref{th:deteq1} as well as the first part of Theorem~\ref{th:deteq2}; indeed, note from $(Z+A)^\trans Q(Z+A)=\tilde{Q}(Z+A)^\trans (Z+A)=I_T-\gamma\tilde{Q}$ that $Z^\trans Q (Z+A)=I_T-\gamma \tilde{Q}-A^\trans Q(Z+A)$, so that
\begin{align}
	\label{eq:ZQBQZ}
	Z^\trans QBQZ &\leftrightarrow \eta^2\gamma \bar{\tilde Q}\bar{G}(I_T+\eta^2\bar{R})^{-1} + \eta^2 A^\trans \bar{Q} A (I_T+\eta^2\bar{R})^{-1}\bar{G}(I_T+\eta^2\bar{R})^{-1}\nonumber \\ 
	&-\eta^2 Z^\trans QBQA\bar{R}(I_T+\eta^2\bar{R})^{-1}.
\end{align}
In this expression, the last right-hand side term still involves $Z^\trans QBQA$, yet to be characterized. This is the objective of the next step, which coincides with the study of the term $(II)$ in \eqref{eq:I-IV,2}.

Following the derivation of term $(I)$, terms $(II)$ and $(III)$ are easily obtained (indeed, they somewhat boil down to \eqref{eq:ZQQZ} without the first right-hand side term and without the components $\varepsilon^\trans (W^q)^\trans$ in the subsequent terms). Precisely, all calculus made, we find that
\begin{align*}
	QBQZ &\leftrightarrow -\eta^2 Q(Z+A)\bar{G} - \eta^2 QBQ(Z+A)\bar{R}
\end{align*}
from which
\begin{align*}
	QBQZ &\leftrightarrow -\eta^2 Q(Z+A)\bar{G} (I_T+\eta^2 \bar{R})^{-1} - \eta^2 QBQA\bar{R} (I_T+\eta^2 \bar{R})^{-1}.
\end{align*}
Again, the first right-hand side term is easily expressed by Theorem~\ref{th:deteq1} and the first result of Theorem~\ref{th:deteq2}, from which
\begin{align}
	\label{eq:QBQZ}
	QBQZ &\leftrightarrow -\eta^2 \bar{Q}A(I_T+\eta^2 \bar{R})^{-1} \bar{G} (I_T+\eta^2 \bar{R})^{-1} - \eta^2 QBQA\bar{R} (I_T+\eta^2 \bar{R})^{-1}.
\end{align}
but the second term now involves the quantity $QBQ$ which is our next target. Since studying $QBQ$ entails studying $A^\trans QBQA$, this shall provide us with the term $(IV)$ in \eqref{eq:I-IV,2}. To address $QBQ$, it suffices to estimate $\EE[ Q_{ij}Q_{kl} ]$; from the resolvent identity $Q_{ij} =\frac1\gamma {\bm\delta}_{ij} - \frac1\gamma [\frac1TXX^\trans Q]_{ij}$, this is developed as
\begin{align*}
	\EE \left[ Q_{ij} Q_{kl} \right] &= - \frac1\gamma \left( \EE [ [Z^\trans Z Q]_{ij}Q_{kl} ] + \EE [ [ZA^\trans Q]_{ij}Q_{kl} ] + \EE[ [AZ^\trans Q]_{ij}Q_{kl} ] + \EE [ [AA^\trans Q]_{ij}Q_{kl} ] \right) \nonumber \\
	&+ \frac1\gamma {\bm\delta}_{ij} \EE[ Q_{kl} ].
\end{align*}
The deterministic equivalent for $\EE[ Q_{kl} ]$ is already known, and we are then left to evaluate the first four terms, some of which can be retrieved from previous calculus. Developing each term, integrating the previously developed equivalents, while introducing the matrix $B$ and summing, after some tedious calculus, we finally obtain
\begin{align*}
	QBQ &\leftrightarrow \frac1\gamma B\bar{Q} + \frac{\eta^2}{\gamma} A(I_T+\eta^2 \bar{R})^{-1}\bar{G}(I_T+\eta^2\bar{R})^{-1}A^\trans \bar{Q} \nonumber \\
	&- \eta^2\bar{\tilde R} QBQ + \frac1{\gamma} \bar{\tilde G} \bar{Q} - \frac1\gamma A(I_T+\eta^2\bar{R})^{-1}A^\trans QBQ
\end{align*}
where we introduced the notation
\begin{align*}
	\bar{\tilde G} &= \sum_{q=-\infty}^\infty \eta^2 \EE \left[ \frac1T\tr \left( J^q (A+Z)^\trans QBQ (Z+A) \right) \right] \sum_{k\geq 0} W^{k+(-q)^+} (W^{k+q^+})^\trans.
\end{align*}
Gathering all terms proportional to $QBQ$, we finally obtain
\begin{align}
	\label{eq:QBQ}
	QBQ &\leftrightarrow \bar{Q}(B+\bar{\tilde G})\bar{Q} + \eta^2 \bar{Q}A(I_T+\eta^2 \bar{R})^{-1} \bar{G} (I_T+\eta^2\bar{R})^{-1}A^\trans \bar{Q}.
\end{align}
Substituting \eqref{eq:QBQ} in \eqref{eq:QBQZ}, then substituting the result in \eqref{eq:ZQBQZ}, we may now completely characterize $\frac1TX^\trans QBQX$ (after simplification) as
\begin{align*}
	\frac1TX^\trans QBQX &\leftrightarrow \eta^2\gamma^2 \bar{\tilde Q}\bar{G} \bar{\tilde Q} + (I_T+\eta^2\bar{R})^{-1}A^\trans \bar{Q}[B+\bar{\tilde G}]\bar{Q}A(I_T+\eta^2\bar{R})^{-1}.
\end{align*}

It remains to evaluate $\EE[\frac1T\tr ( J^q (A+Z)^\trans QBQ (Z+A) )]$ in the expression of $\bar{\tilde G}$. For this, we shall exploit the fact that $A=MU$ which, since $M$ has columns of exponentially decreasing norm, can be considered as a matrix of rank ``essentially of order $O(\log(T))$''; that is, while being full rank, $A$ can be well approximated in spectral norm by the product $\check M\check U$ of the first $O(\log(T))$ columns $\check M$ of $M$ and first $O(\log(T))$ rows $\check U$ of $U$. This entails that, in the deterministic approximation for $(A+Z)^\trans QBQ (Z+A)$, only the terms not involving a product with $A$ or $A^\trans$ will remain after taking the normalized trace. And thus we get, after development and simplification
\begin{align*}
\left\| \bar{\tilde G} - \sum_{q=-\infty}^\infty \gamma^2\eta^4 \frac1T\tr \left( J^q \bar{\tilde Q}\bar{G} \bar{\tilde Q} \right) \sum_{k\geq 0} W^{k+(-q)^+} (W^{k+q^+})^\trans \right\| \to 0.
\end{align*}
It then suffices to use concentration identities and the results of Appendix~\ref{app:Th1} to finally substitute $\bar R$ with $R$, $\bar{\tilde{R}}$ with $\tilde R$, and $\bar{G}$, $\bar{\tilde G}$ with $G$ and $\tilde G$, respectively. This concludes the proof of Theorem~\ref{th:deteq2}.


\begin{remark}[On the speed of convergence]
	\label{rem:speed}
	To better appreciate the interplay between $\eta^2$ and $n,T$, note that all convergences discussed in Appendices~\ref{app:Th1}--\ref{app:Th2} involve either quadratic forms of the type $a^\trans Qa$ for $Q\in\RR^{n\times n}$ a random matrix based on some $\varepsilon\in\RR^{n\times T}$, matrix with independent entries, or normalized traces $\frac1n\tr Q$. It is a standard central limit result in random matrix theory that the former quadratic form $a^\trans Qa$ fluctuates at speed $O(n^{-\frac12})$, that is, ${\rm var}(a^\trans Qa)=O(n^{-1})$, and that normalized traces fluctuate at the faster speed $O(n^{-1})$. As such, the results of Theorems~\ref{th:deteq1}--\ref{th:deteq2} and Proposition~\ref{prop:fundeq}--\ref{prop:fundeq2} can be trusted with high probability within a $O(n^{-\frac12})$ error bound.

	With respect to $\eta^2$, the bounds between random quantities and deterministic equivalents, say $Q$ and $\bar{Q}$, are proportional to $1/\eta^2$. This is why $\eta^2$ is assumed fixed and not decaying in our results. Nonetheless, as both bounds in $n$ and $\eta^2$ multiply, it is expected that convergence is maintained in general so long that $n^{-\frac12}/\eta^2\to 0$, i.e., when $\eta^2 \gg n^{-\frac12}$.
\end{remark}


\section*{Acknowledgments}
The work of Couillet and Tiomoko Ali is supported by the ANR RMT4GRAPH Project (ANR-14-CE28-0006).



\bibliographystyle{apalike}
\bibliography{/home/romano/Documents/PhD/phd-group/papers/rcouillet/tutorial_RMT/book_final/IEEEabrv.bib,/home/romano/Documents/PhD/phd-group/papers/rcouillet/tutorial_RMT/book_final/IEEEconf.bib,/home/romano/Documents/PhD/phd-group/papers/rcouillet/tutorial_RMT/book_final/tutorial_RMT.bib}

\end{document}